\documentclass[natbib]{pretty-preprint}

\usepackage{amsmath,amsfonts,bm}

\def\eqref#1{equation~\ref{#1}}
\def\1{\bm{1}}

\DeclareMathAlphabet{\mathsfit}{\encodingdefault}{\sfdefault}{m}{sl}
\SetMathAlphabet{\mathsfit}{bold}{\encodingdefault}{\sfdefault}{bx}{n}

\usepackage{multirow,float}
\usepackage{longtable}
\usepackage{colortbl}
\definecolor{timebraidrow}{gray}{0.95}
\usepackage{placeins} % \FloatBarrier: keep the ablation figures out of the references
\usepackage{wrapfig}
\usepackage{titletoc}
\hypersetup{linktoc=all}
\setcitestyle{authoryear,round,citesep={;},aysep={,},yysep={;}}
\tcbuselibrary{skins,raster,breakable}
\tcbset{
  promptcard/.style={
    enhanced, arc=2.5pt, boxrule=0.5pt,
    colframe=black!40, colback=blue!4, colbacklower=black!5,
    colbacktitle=blue!12, coltitle=black, fonttitle=\bfseries\small,
    left=6pt, right=6pt, top=3pt, bottom=3pt,
    toptitle=1pt, bottomtitle=1pt,
    fontupper=\small, fontlower=\small,
    before upper={\raggedright}, before lower={\raggedright}
  }
}
\newfloat{algorithm}{tbp}{loa}
\floatname{algorithm}{Algorithm}
\usepackage{listings}
\newcommand{\best}[1]{\textbf{#1}}
\newcommand{\second}[1]{\underline{#1}}
\newcommand{\third}[1]{#1$^{\dagger}$}

\definecolor{tsinclr}{RGB}{20,130,120}
\definecolor{tsoutclr}{RGB}{230,140,40}
\newcommand{\patchcell}[1]{\textcolor{#1}{\rule[-1pt]{2.6pt}{6pt}}\hspace{0.8pt}}
\newcommand{\tsinpatches}{\patchcell{tsinclr}\patchcell{tsinclr}\patchcell{tsinclr}}
\newcommand{\tsoutpatches}{\patchcell{tsoutclr}\patchcell{tsoutclr}\patchcell{tsoutclr}}
\title{\texorpdfstring{\raisebox{-0.18em}{\includegraphics[height=1.12em]{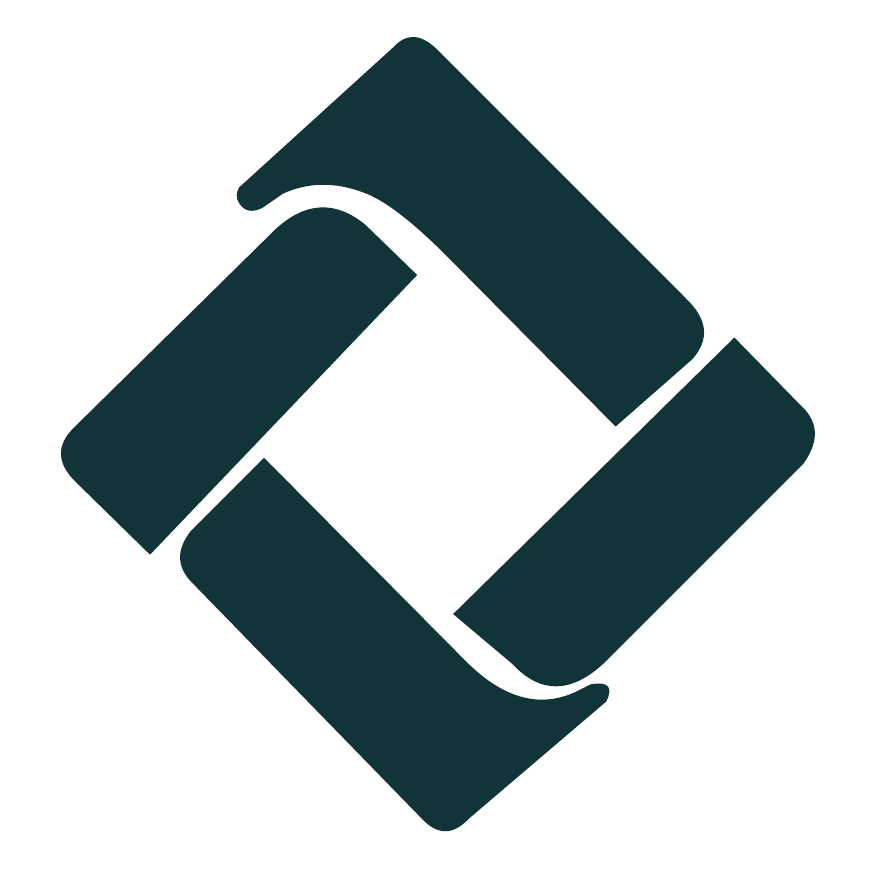}}\hspace{0.25em}}{}TimeBraid: Unifying Time Series and Language for Understanding and Forecasting}
\runningtitle{TimeBraid: Unifying Time Series and Language}

\author[1,3,\textdagger]{Xinyue Wang}
\author[2]{Jiacheng Pang}
\author[3]{Kun Zhou}
\author[2]{Kexin Zhang}
\author[2]{Defu Cao}
\author[1]{Fan Feng}
\author[2]{Faisal}
\author[1,3,\textdagger]{Songyao Jin}
\author[2]{Yan Liu}
\author[1,3]{Biwei Huang}
\affiliation[1]{University of California San Diego}
\affiliation[2]{University of Southern California}
\affiliation[3]{Aether AI}
\contribution[\textdagger]{Part of the work was completed during an internship at Aether AI}

\metadata[Correspondence][\faEnvelope]{Xinyue Wang (\email{xiw159@ucsd.edu})}
\resourcelinks{%
\resourceitem{}{\faGlobe}{\href{https://xinyuewangg.com/projects/timebraid/}{\textbf{Project}}}\hspace{2em}
\resourceitem{}{\faGithub}{\href{https://github.com/CharonWangg/TimeBraid}{\textbf{Code}}}\hspace{2em}
\resourceitem{}{\huggingfaceicon}{\href{https://huggingface.co/XinyueWangg/TimeBraid-2.5B}{\textbf{Model}}}\hspace{2em}
\resourceitem{Data}{\huggingfaceicon}{\href{https://huggingface.co/datasets/XinyueWangg/TimeBraid-Alignment}{Alignment}\enspace\textperiodcentered\enspace\href{https://huggingface.co/datasets/XinyueWangg/TimeBraid-SFT}{SFT}}
}

\abstract{%
  We present TimeBraid, a series of unified time-series and language models that align pretrained language models and pretrained time-series foundation models through interleaved global residual attention layers. Each model inherits knowledge, instruction following, and reasoning from one side, continuous-signal perception and zero-shot forecasting from the other, and fuses the two in a shared representation space where both modalities are understood and generated. We study the design choices that make such unified modeling work: where to align the two representation spaces, how to ground language in temporal structure, how to balance understanding with generation, and how to keep joint optimization stable. The resulting recipe combines a unified prompting scheme for diverse time-series and text tasks, stabilized joint training, and supervision from 2.2M curated series--text pairs and 4.9M instruction-tuning samples.
  Across benchmarks spanning time-series perception, understanding, reasoning, and both context-aided and unimodal forecasting, TimeBraid remains competitive with far larger general-purpose models and task-specific counterparts.

}

\keywords{Unified Time-Series and Language Models, Time-Series Analysis, Multi-Modal Forecasting}

\graphicspath{{figures/}}

\begin{document}

\maketitle

% Figure 1 is deliberately NOT a float: a [t] float cannot share page 1 with the
% title card, so it lands on page 2. A bare \centering block is used instead of a
% center environment because center's \addvspace swallows the negative skip that
% pulls the figure up against the card. The 1860 x 900 Figma candidate K uses
% the full text width while keeping the figure and caption together on page 1.
% An overflow here produces NO LaTeX warning, and the caption
% silently moves to the top of page 2. After any edit
% to the abstract, the metadata rows, the caption, or this width, run
%   python3 agent_notes/check-figure1-caption-on-page1.py
\begingroup
\centering
\vspace{-14pt}% leave about 8pt of visible separation below the title card
% Original figure retained: overview-homepage-full-width.pdf (full linewidth).
% Previous candidate retained: overview-figma-733-93-20260920.pdf (0.90 linewidth).
% Data-card alternative H: overview-figma-798-2-20260920.pdf (0.82 linewidth).
% Previous accepted version K: overview-figma-825-2-20260920.pdf.
% Continuation candidate L retained: overview-figma-836-2-20260923.pdf
% (TSAQA test row 37004; B is the gold answer, not a verified model output).
% Previous candidate M: TSAQA test row 30012 (Fourier magnitude) and
% TimeSeriesExam test row 499 / id 500 (Granger direction). Both displayed
% answers are dataset gold labels, not verified TimeBraid predictions.
% Current candidate N: Figma node 903:2, exported on 2026-09-24.
\includegraphics[width=\linewidth,keepaspectratio]{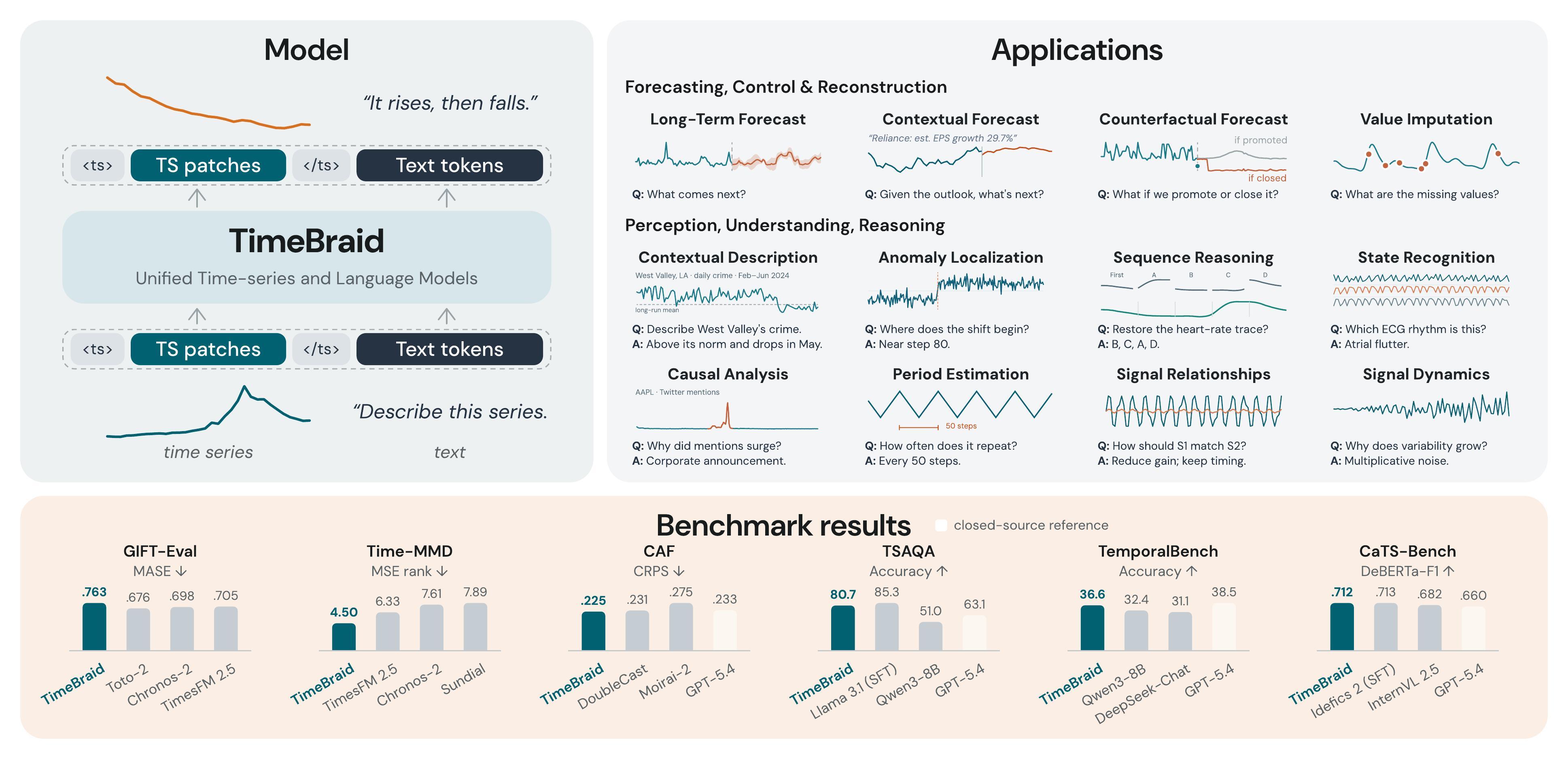}\par
\captionof{figure}{\textbf{Overview of TimeBraid.} The unified time-series and language interface, application examples, and results on time-series understanding (TSAQA, TemporalBench, CaTS-Bench) and forecasting (GIFT-Eval, Time-MMD, CAF) benchmarks.}%\textcolor{red}{Kun: Font in the figure is too small to read clearly.}
\label{fig:overview}
\endgroup

\section{Introduction}

% The motivation of unified models for multiple modalities
% \textcolor{red}{Kun: make sense but a bit overclaim, the first sentence is more like a motto before the whole paper. Also, key phrase``time series'' should appear here. Revised version:}
From early development, human intelligence is shaped by continuous interaction with a rich stream of heterogeneous signals. Vision, audition, language, and tactile sensation arrive as synchronized time series, allowing information from one modality to contextualize and complement information from others \citep{smithgasser2005embodied,smith2018curriculum,orhan2020childeyes}. Existing efforts to build multimodal systems have largely relied on composed workflows or architectures that merge separately developed, modality-specific modules. Unified multimodal models explore a complementary direction: learning shared representations for understanding and generating heterogeneous signals within a single set of model weights \citep{chameleonteam2024chameleon,zhou2024transfusion,xie2024showo,wang2024emu3,chen2025januspro,deng2025bagel,qwen3omni2025,nativemm2026roadmap}. Emerging evidence suggests that joint pretraining across modalities may foster capabilities that are difficult to acquire from any individual modality in isolation \citep{beyondlm2026}.

% The motivation of considering time-series as a necessary modality in unified models
However, time-series data remains largely outside this unification. Numerical records of dynamical systems (e.g., physiology, power grids, markets, and climate) lack the semantic context for identification and interpretation~\citep{goldberger2000physionet,hong2016probabilistic,tsay2010financial,lam2023graphcast,shallue2018identifying}. Language grounding can supply this context and connect time series with broader world knowledge, motivating their inclusion as an essential modality in unified models.

% Previous attempts towards unified time-series and language models, why not enough?
Existing approaches typically connect time series and language by adding specialized components, ranging from projection layers to signal tokenizers, to pretrained language or time-series models~\citep{xie2025chatts,timeomni2025,jin2024timellm,wang2025chattime}. However, such components introduce representational gaps, causing semantic misalignment and degrading pretrained capabilities, and each resulting system covers only a fragment of the full bidirectional interface (Section~\ref{sec:related_work}). Aligning the native representations of both pretrained hosts while largely preserving their capabilities remains underexplored.

% Towards the ambitious goal, what we have done?    
Such alignment requires large-scale paired pretraining, yet the inherent semantic gap between continuous signals and discrete language makes it challenging. To bridge this gap, we present \textbf{TimeBraid}, a mixture-of-Transformers architecture that combines modality-specific processing with cross-modal context sharing. Inspired by how specialized sensory organs perceive different signals while the brain integrates them for decision-making, TimeBraid employs separate pretrained experts for language reasoning, time-series perception, and forecasting, while connecting them through \emph{global residual attention}. This mechanism organizes their representations into a shared chronological context, enabling cross-modal interaction while substantially preserving pretrained language and forecasting capabilities. Building on this architecture, we introduce a two-stage training recipe that first aligns time series and language on large-scale paired data and then adapts the unified model to diverse downstream tasks.

Across twelve understanding and forecasting benchmarks \citep{tsaqa2026,cai2024timeseriesexam,temporalbench2026,catsbench2025,liu2024timemmd,wang2025chattime,zheng2026caf7m,williams2025cik,aksu2024gifteval}, TimeBraid at 1.2B--6.7B parameters serves both directions of the series--text interface in one model, as summarized in Figure~\ref{fig:overview}. For understanding, TimeBraid-6.7B reaches 80.65\% on TSAQA, compared with 85.26\% for TSAQA-specific LLaMA3.1-8B SFT and 63.10\% for GPT-5.4. On the held-out TB-MCQ suite it performs on par with GPT-5.4 (36.58\% vs.\ 38.50\%). Its CaTS-Bench captions match the strongest frontier VLMs and the models finetuned on the benchmark. For forecasting, TimeBraid attains the best average MSE rank on the nine Time-MMD domains and the best CRPS on CAF. On GIFT-Eval it largely maintains comparable zero-shot forecasting to state-of-the-art time-series foundation models. It is also the only non-LLM model whose Ctrl-F forecasts follow the stated condition, and attains the best Macro MSE and Macro MAE on CGTSF. Further analysis systematically ablates the design space of this new model class and singles out the shared attention space, stabilized joint optimization, and the alignment stage as the choices that matter most. Together, these results establish TimeBraid as a series of powerful and versatile unified models of time and language.
% \textcolor{red}{Kun: the three listed contributions can not shock me. Why not just share the experimental results with more details and highligh the improvement. Look, you have Figure1 to support you.} \textcolor{blue}{Experiments across [NUMBER] benchmarks and [NUMBER] domains demonstrate that \textbf{TimeBraid} delivers strong performance on both temporal understanding and generation tasks (Figure~\ref{fig:overview}). For temporal understanding, TimeBraid achieves [RESULT] on question answering, [RESULT] on anomaly explanation, and [RESULT] on imputation, outperforming [BASELINE/PRIOR STATE OF THE ART] by [MARGIN]. For temporal generation, it improves [METRIC] by [MARGIN] on [SHORT-/LONG-HORIZON] forecasting and remains effective under contextual and counterfactual conditions. Further analysis shows that [KEY ABLATION OR SCALING RESULT], highlighting the importance of [GLOBAL RESIDUAL ATTENTION/TWO-STAGE TRAINING/NATIVE ALIGNMENT]. Together, these results establish TimeBraid as [CAUTIOUS HIGH-LEVEL CONCLUSION].}

\section{Preliminaries}
\label{sec:preliminaries}

Every time-series and language task can be described as an interleaving of the two modalities. Formally, an example is a segment sequence $S=(s_1,\dots,s_M)$, where each segment is either a \emph{text segment} (a token sequence) or a \emph{time-series segment} (a univariate value sequence $(x_1,\dots,x_{n})\in\mathbb{R}^{n}$; a multivariate series enters as one segment per variable), and each segment is designated as \emph{context} or \emph{target}; the number and ordering of segments are unconstrained. Context segments are fully observed: context text $s_{\mathrm{text}}$ carries instructions, questions, and background, and a context time-series segment $s_{\mathrm{ts},C}$ carries observed measurements. Target segments are the model's outputs: a target text segment is a language response to be generated, and a target time-series segment $s_{\mathrm{ts},T}=(x_1,\dots,x_h,x_{h+1},\dots,x_{h+f})$ restates the observed history $(x_1,\dots,x_h)$ of its context segment and continues it with $f$ future values to be generated.

% Every task thereby becomes a single conditional generation problem: the model generates the content of each target segment conditioned on everything preceding it, and the example likelihood factorizes as $\prod_{i\in\mathcal{I}_{\mathrm{tgt}}} p_\theta(s_i\mid s_{<i})$, where $\mathcal{I}_{\mathrm{tgt}}$ indexes the target segments and, within a target time-series segment, only the $f$ future values are generated and scored. Canonical tasks are instances of this template: univariate forecasting is the pair $(s_{\mathrm{ts},C},\,s_{\mathrm{ts},T})$; contextual forecasting conditions the same pair on additional text; captioning and question answering emit a text target from time-series and text context; multivariate relational understanding composes several context series in one sequence.

\section{Method}

TimeBraid is a unified time-series and language model built on top of native language and time-series representations: a pretrained language model and a pretrained time-series foundation model. Its design pursues two properties: (i) the framework should be an omni learner and practitioner, not bound by a specific task format, accepting any time-series and text segment composition in input and output; and (ii) it should leverage the pretrained representations in existing language models and time-series foundation models rather than relearning them from scratch. We first describe the data recipe that develops these capabilities, then the unified prompting and model design that realize (i) and (ii), and finally the training objectives, optimization setup, and inference procedure. 

\subsection{Data Recipe}

\begin{figure}[!htbp]
    \centering
    % Original figure retained: data_recipe.pdf.
    \includegraphics[width=\linewidth]{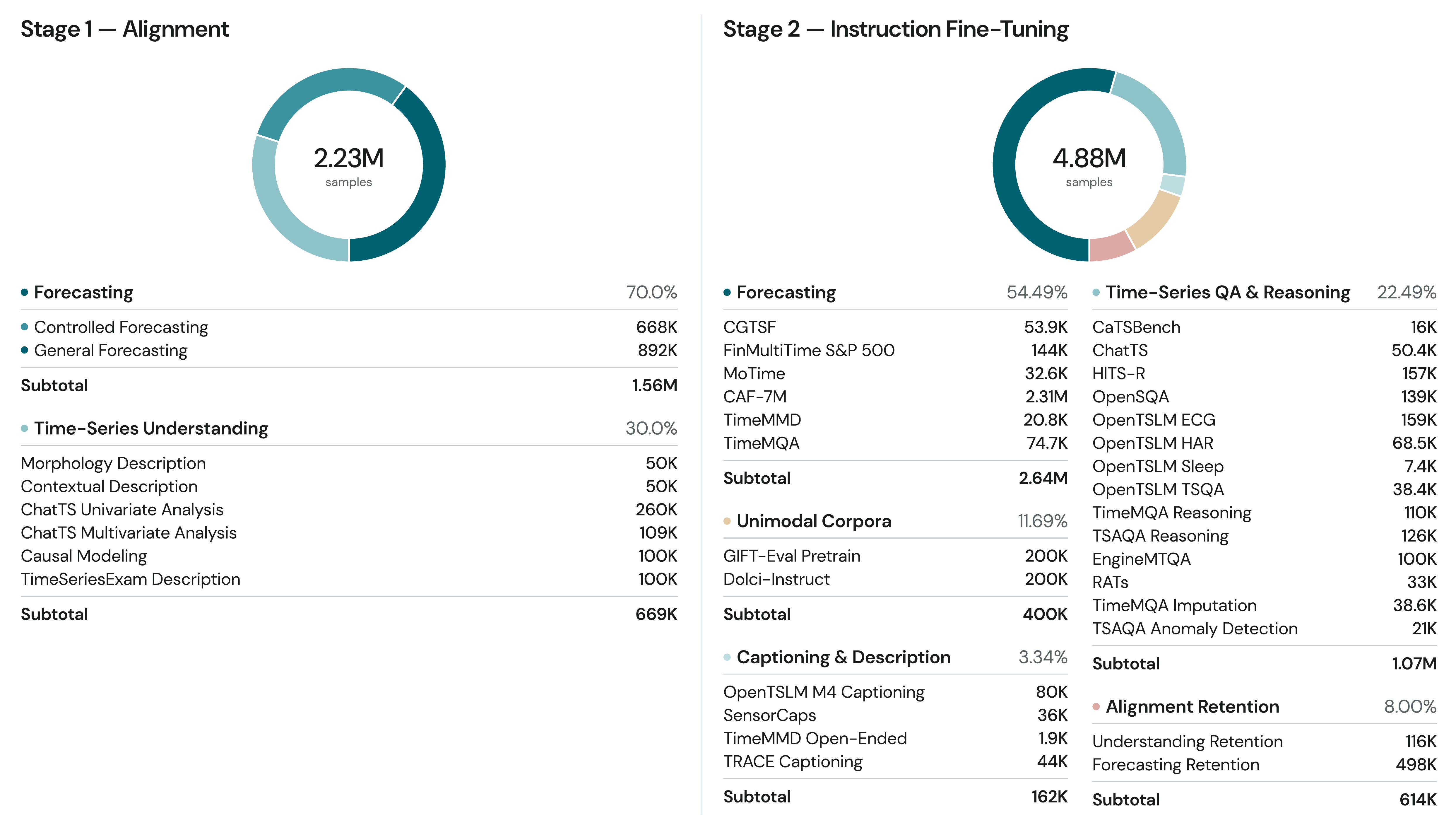}
    \caption{Two-stage training recipe. Stage 2 counts show the final one-pass sample inventory, while percentages show training-time sampling shares.}
    \label{fig:pipeline}
\end{figure}

Our data is arranged as a two-stage curriculum (Figure~\ref{fig:pipeline}). The first stage teaches basic time-series understanding and controllable forecasting on constructed paired data. The second stage builds on these basics and targets instruction following, task-specific performance, and real-world domain knowledge with a heterogeneous instruction mixture.

\paragraph{Stage 1: Alignment.}

The alignment stage uses 2.23M paired examples, split 3:7 between understanding and forecasting. The understanding data covers both univariate and multivariate cases: morphology captions describe the basic structures of a single series (trends, seasonality, regimes, salient features, inter-series relationships); context-rich captions describe real series together with their real-world context; programmable captions in the style of ChatTS \citep{xie2025chatts} are generated attribute-first, so each description matches its series precisely by construction; multivariate examples are generated from structural causal models with known interaction structure; and templated programmable tasks \citep{cai2024timeseriesexam} cover the same content from diverse task perspectives. The forecasting data focuses on controllability and is built with future-aware annotation: annotations are produced with access to the future, which is never included in the model input. The forecasting control set synthesizes several distinct futures from one shared history using programmable operators and captions the condition that leads to each future, so the model learns to follow stated conditions when forecasting. The general forecasting set annotates realized futures of real series, aligning context-conditioned forecasts with natural real-world evolution. We rewrite some forecasting examples in different tones and genres to make their presentation more diverse and realistic. Appendix~\ref{app:data_curation} details the construction of each family.

\paragraph{Stage 2: Supervised Fine-Tuning.}

The second stage uses a heterogeneous mixture spanning time-series question answering and reasoning, captioning and description, contextual and unimodal long-horizon forecasting, and domain-specific applications in energy, climate, healthcare, finance, sensing, and industrial monitoring. Its final one-pass inventory contains 4{,}881{,}583 samples. The training-time sampling shares allocate 54.49\% to forecasting, 22.49\% to time-series QA and reasoning, 3.34\% to captioning and description, 11.69\% to unimodal data, and 8.00\% to alignment retention. The retained alignment quota contains 614{,}113 samples: 116{,}294 from understanding and 497{,}819 from forecasting.

% Keep the data overview with its recipe, before introducing the model.
\FloatBarrier
\subsection{Unified Time-Series and Language Modeling}

\paragraph{Unified Time-Series and Language Prompting.} The prompting scheme defines the serialization $S \mapsto u_{1:T}$, mapping a segment sequence (Section~\ref{sec:preliminaries}) to the mixed sequence of $T$ text tokens and time-series patches that the model consumes. Each example is rendered as a chat transcript in the native chat template, and time-series segments are wrapped with special tokens \texttt{<ts></ts>} (as illustrated in the examples below). Each context time-series segment is prepended with a statistics block indicating the length, mean and standard deviation of the segment, and each target segment is a bare \texttt{<ts></ts>}. The text segments are tokenized into text tokens and the time-series segments are normalized and tokenized into patches of fixed length $P$, according to the pretrained language model and time-series foundation model respectively. For the context time-series segments, we use full-segment normalization. For a target time-series segment, we normalize the full segment with the statistics of its history prefix and then apply a rolling RevIN normalization \citep{kim2022revin} to better handle nonstationarity and respect the causal order in the prediction process. The global sequence ordering is determined by the chronological order of the original segment sequence. The examples below show one multivariate understanding and one univariate forecasting transcript.

\begin{center}
\begin{tcbraster}[raster columns=2, raster column skip=8pt]
\begin{tcolorbox}[promptcard, title=Example: multivariate understanding]
\textbf{User:} Equity: \texttt{<stats>len=94, mean=103, std=4</stats><ts>}\tsinpatches\texttt{</ts>}. Stress: \texttt{<stats>len=90, mean=19, std=7</stats><ts>}\tsinpatches\texttt{</ts>}. How do they relate?
\tcblower
\textbf{Assistant:} Stress leads each equity drawdown.
\end{tcolorbox}
\begin{tcolorbox}[promptcard, title=Example: univariate forecasting]
\textbf{User:} Here is the last 96 hours of server CPU load: \texttt{<stats>len=96, mean=42.1, std=8.3</stats><ts>}\tsinpatches\texttt{</ts>}. A deployment just shipped, forecast the next 96 hours.
\tcblower
\textbf{Assistant:} \texttt{<ts>}\tsinpatches\tsoutpatches\texttt{</ts>}
\end{tcolorbox}
\end{tcbraster}
\end{center}

\paragraph{Tri-Expert Architecture.}

% Place the architecture at the next page top, after the prompting examples.
\suppressfloats[t]
\begin{figure}[t]
    \centering
    % Original figure retained: architecture_without_masks.pdf.
    % Previous palette retained: architecture-figma-1204-2-20260920.pdf.
    \includegraphics[width=\linewidth]{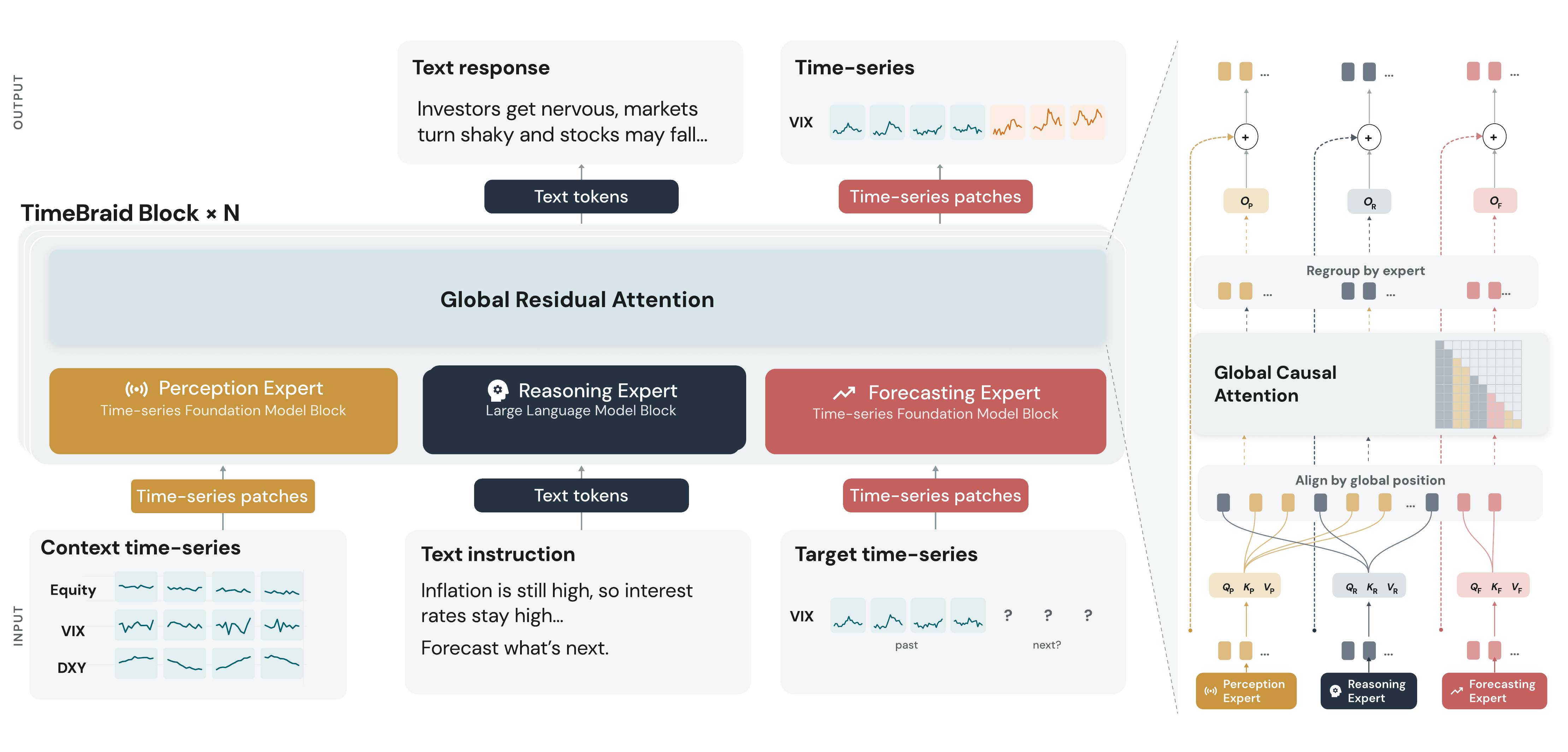}
    \caption{TimeBraid architecture: the model components and the flow of information between the time-series and language modalities.}
    \label{fig:architecture}
\end{figure}

The core part of TimeBraid is a Mixture-of-Transformers (MoT) architecture that unifies time-series perception, reasoning and forecasting within a single framework. The perception expert encodes context time-series segments, the reasoning expert handles text input and output such as instructions and language responses, and the forecasting expert generates target time-series segments based on the time-series and text contexts. We initialize the experts with their corresponding pretrained models, e.g., large language models and time-series foundation models. Note that while the perception and forecasting experts receive different input with different normalization, they can share the same underlying model (dual-tower architecture). One may also consider using different backbones (tri-tower architecture) to further disentangle the representations used for understanding and forecasting, analogous to unified vision models that decouple understanding from generation representations \citep{chen2025januspro}. However, we do not observe significant performance improvement by using extra backbones in ablation studies (Figure~\ref{fig:ablation_loss}E). The choice of backbone for each expert tower is not limited under this framework, and we leave the study of model composition as future work.
% \textcolor{red}{Kun: should avoid using separate paragraphs under a paratitle, it looks fragmented.}

% \textcolor{red}{Kun: why not add the Paratitle named Global Residual Attention.}
\paragraph{Global Residual Attention.} The time-series tower has fewer layers than the language tower, so each of its layers is paired with one language layer, evenly interleaved across depth; the remaining language layers are unpaired and process only their own tokens. The towers interact through global residual attention layers at the paired positions. At every paired layer, each tower $r\in\{\mathrm{L},\mathrm{T}\}$ first applies its native block to its own tokens, producing hidden states $H_r\in\mathbb{R}^{n_r\times d_r}$, where $n_r$ is the number of tokens in stream $r$ and $d_r$ the native width of tower $r$. The global residual attention layers project them into a shared attention space of width $d$ for self-attention on the global mixed sequence,

\begin{equation}
    Q_r = \psi^{Q}_r\!\big(\phi_r(H_r)\,W^{Q}_r\big),\qquad
    K_r = \psi^{K}_r\!\big(\phi_r(H_r)\,W^{K}_r\big),\qquad
    V_r = \phi_r(H_r)\,W^{V}_r,
    \label{eq:fusion_proj}
    \end{equation}
where $W^{Q}_r, W^{K}_r, W^{V}_r\in\mathbb{R}^{d_r\times d}$ map each stream to the shared width, $\phi_r$ is a stream-specific RMSNorm on hidden states, and $\psi^{Q}_r, \psi^{K}_r$ are per-head RMSNorms on queries and keys. The projected tokens of both streams are reordered into single global sequences $\tilde{Q}, \tilde{K}, \tilde{V}$ by their timeline positions $\tau$ and attended jointly under a causal mask,
    \begin{equation}
    O = \mathrm{Attention}\big(\mathrm{RoPE}_{\tau}(\tilde{Q}),\, \mathrm{RoPE}_{\tau}(\tilde{K}),\, \tilde{V}\big),\qquad
    H_r \leftarrow H_r + O_r\,W^{O}_r,
    \label{eq:fusion_attn}
    \end{equation}
where $O_r$ denotes the rows of $O$ belonging to stream $r$, routed back to their original positions, and the output projections $W^{O}_r\in\mathbb{R}^{d\times d_r}$ are zero-initialized so that each global layer starts as an identity map. A text token thus attends to all earlier tokens of both modalities, and a patch token does the same. The cross-modal information is exchanged token-to-token in both directions, enabling flexible and rich interactions between time-series and text.

% \textcolor{red}{Kun: why not add the Paratitle named Disentangled Positional Embeddings.}
\paragraph{Disentangled Positional Embeddings.} We use three kinds of positional embeddings: (1) the language tower applies rotary embeddings (RoPE) inside its native blocks, on the patch-expanded positions, (2) the time-series tower applies its native positions locally within each segment, restarting at every segment, (3) the residual attention applies RoPE on the shared global ordered positions $\tau$ (Eq.~\ref{eq:fusion_attn}), where both modalities are placed on one sequential axis. Algorithm~\ref{alg:mot_forward} in Appendix~\ref{app:implementation} gives pseudocode for the full forward pass.

% \textcolor{red}{Kun: The Training objective and Stabilizing the Joint Optimization paragraphs are not Architecture, right? Why not put them into a new subsection named Learning Objective.}

\subsection{Training and Inference}

\paragraph{Joint Training Loss.}

Time-series and language modeling are jointly optimized with separate losses. 
Text targets are trained with the standard next-token cross-entropy loss $\mathcal{L}_{\mathrm{CE}}$. Time-series targets are supervised autoregressively via Multi-Patch Prediction (MPP): every patch of a target segment is predicted from its preceding context, and each predicted patch incurs a point term and a quantile term,
\begin{equation}
\mathcal{L}_{\mathrm{TS}}
= \frac{1}{|\mathcal{T}|} \sum_{t \in \mathcal{T}} \bigg[
\big\lVert \hat{y}_t - y_t \big\rVert_2^2
+ \frac{1}{|\mathcal{Q}|} \sum_{q \in \mathcal{Q}}
\max\!\big(q\,(y_t - \hat{y}_t^{(q)}),\; (q-1)\,(y_t - \hat{y}_t^{(q)})\big)\bigg],
\label{eq:ts_loss}
\end{equation}
where $\mathcal{T}$ indexes the target patches of the example, $y_t$ is the $t$-th standardized target patch, $\hat{y}_t$ its point prediction, and $\hat{y}_t^{(q)}$ its prediction at quantile $q\in\mathcal{Q}=\{0.1,0.2,\dots,0.9\}$; the second term is the pinball loss. We use the point regression to describe the central trajectory and the quantile term to capture the predictive distribution and uncertainty. The total objective is $\mathcal{L} = \mathcal{L}_{\mathrm{CE}} + \alpha\,\mathcal{L}_{\mathrm{TS}}$.

\paragraph{Stabilizing the Joint Optimization.}

The language and time-series losses have different optimization characteristics, and naive joint training leads to ineffective learning. We observe that on strongly non-stationary segments, where the future departs from the history statistics, the regression loss $\mathcal{L}_{\mathrm{TS}}$ can spike by orders of magnitude, and the spikes disrupt language-side learning through the shared backbone updates. We address this with two techniques. First, \emph{robust normalization}: during the causal RevIN normalization, each raw target value $x$ is transformed as $y = \operatorname{asinh}\!\big((x - \mu)/\sigma\big)$, where $\mu$ and $\sigma$ are the rolling history statistics of the RevIN step, producing the standardized targets in Eq.~\ref{eq:ts_loss}. The map is linear near zero and logarithmic in the tails, so extreme targets are compressed into a bounded range. Second, \emph{loss capping}: the time-series loss of each response is rescaled by the detached factor $s = \min\!\big(1,\, c/\operatorname{stop\_gradient}(\ell)\big)$, where $\ell$ is the response's time-series loss value and $c$ is a fixed cap threshold (Appendix~\ref{app:implementation}). This caps the value and gradient contribution of outlier responses and leaves in-range responses unchanged.

\paragraph{Training Setup.}
The alignment and supervised fine-tuning stages train all parameters (the language tower, the time-series tower, and the residual attention layers) with AdamW, using a 500-step warmup and a consistent learning rate of $2\times10^{-5}$ for both modalities, a loss weight of $\alpha = 0.5$ for the time-series loss to reflect the optimal learning rate ratio (Section~\ref{sec:ablation}), a global batch size of 128, and BF16 precision for training efficiency. The alignment stage uses context length 2{,}560 and is trained for one epoch; the supervised fine-tuning stage uses context length 4{,}096 and trains for 30k steps.

\paragraph{Text Strength Modulation at Inference Time.}

Text ranges from decisive (a verified maintenance schedule for server load) to barely informative (noisy commentary on a financial series), and the history itself ranges from regular series that largely predict themselves to volatile ones whose next window hinges on stated events. For a modality this heterogeneous, we expose the conditioning strength as an inference-time modulation. Since $s_{\mathrm{ts},T}$ can be predicted either from $s_{\mathrm{ts},C}$ alone or jointly from $s_{\mathrm{ts},C}$ and $s_{\mathrm{text}}$, querying the model under the two conditioning modes yields two point predictions, $\hat{y}_{\mathrm{ts}}$ (time-series context only) and $\hat{y}_{\mathrm{text+ts}}$ (time-series and text contexts), which we interpolate with a guidance weight $\lambda\in[0,1]$,
\begin{equation}
\hat{y}_\lambda = (1-\lambda)\,\hat{y}_{\mathrm{ts}} + \lambda\,\hat{y}_{\mathrm{text+ts}}.
\label{eq:text_control}
\end{equation}
A small $\lambda$ leans on the global temporal regularities of the series, while a large $\lambda$ keeps the forecast sensitive to the text within the horizon it describes. It provides more flexibility for the model to adapt across domains and inputs of certainty.
%\textcolor{red}{Kun: We can also highlight few interesting application using such interleaved pattern.}

\section{Experiments}
\label{sec:experiments}

We evaluate TimeBraid on time-series understanding and forecasting, covering perception, question answering and reasoning, contextual description, and contextual, controlled, and unimodal forecasting. Comparisons include general-purpose language and vision--language models, specialized time-series models, and unified models. Evaluation protocols, data sources, relationships to the training mixture, and complete results are detailed in Appendices~\ref{app:data_overlap}, \ref{app:eval_protocols}, and~\ref{app:detailed_results}. In main-text tables, bold and underlining mark the best and second-best values in the primary comparison. Detailed appendix tables also mark third-best values with daggers. Dashes indicate unavailable or non-comparable results.

\subsection{Time-Series Understanding}

\begin{table}[!htbp]
    \centering
    \caption{Time-series understanding accuracy (\%) on TSAQA (per-category and overall), TimeSeriesExam (TSExam), and the 16 TemporalBench multiple-choice tasks (TB-MCQ, unweighted macro-average). The Closed-source Reference Models group provides references excluded from ranking. A.D.: anomaly detection; CLS: classification; Char.: characterization; Comp.: comparison; D.T.: data transformation; T.R.: temporal relation; PZ: puzzling/ordering format. Char./Comp./D.T./T.R.\ report the multiple-choice format. SFT denotes TSAQA-specific LoRA fine-tuning \citep{tsaqa2026}. Per-format and per-category results are in Tables~\ref{tab:tsaqa_main_results}, \ref{tab:timeseriesexam}, and \ref{tab:temporalbench_all_models}.}
    \label{tab:understanding_summary}
    \small
    \setlength{\tabcolsep}{3pt}
    \resizebox{\textwidth}{!}{%
\begin{tabular}{lcccccccccc}
    \toprule
\multirow{2}{*}{Model} & \multicolumn{8}{c}{TSAQA} & TSExam & TB-MCQ \\
     \cmidrule(lr){2-9}
     & A.D.\,$\uparrow$ & CLS\,$\uparrow$ & Char.\,$\uparrow$ & Comp.\,$\uparrow$ & D.T.\,$\uparrow$ & T.R.\,$\uparrow$ & PZ\,$\uparrow$ & Overall\,$\uparrow$ & Overall\,$\uparrow$ & Avg.\,$\uparrow$ \\
    \midrule
    \multicolumn{11}{l}{\textit{Closed-source Reference Models}} \\
    GPT-5.4 & 53.32 & 49.57 & 81.98 & 74.47 & 54.94 & 82.96 & 51.02 & 63.10 & 67.83 & 38.50 \\
    GPT-4.1 & 55.85 & 50.38 & 89.36 & 76.99 & 51.13 & 79.09 & 45.77 & 62.82 & 67.89 & 36.91 \\
    GPT-4o & 54.32 & 47.20 & 84.15 & 69.07 & 53.24 & 75.58 & 45.61 & 60.73 & 55.96 & 32.30 \\
    Gemini-2.5-Flash & 52.08 & 49.07 & 81.08 & 72.21 & 60.17 & 84.49 & 60.84 & 65.08 & 53.89 & 34.61 \\
    \midrule
    \multicolumn{11}{l}{\textit{Open-source Large Language Models}} \\
    Qwen3-8B & 50.60 & 50.52 & 66.87 & 63.21 & 34.46 & 67.14 & 21.93 & 51.04 & 46.66 & 32.43 \\
    LLaMA3.1-8B & 54.92 & 50.20 & 62.26 & 49.98 & 36.56 & 40.95 & 6.80 & 44.93 & 35.52 & 25.88 \\
    % LLaMA3.1-8B (SFT) & \best{91.02} & \best{91.27} & \second{83.68} & \best{79.31} & \best{86.62} & \second{97.41} & \best{67.68} & \best{85.26} & -- & -- \\
    % Qwen3-8B (SFT) & \second{87.70} & \second{90.05} & \best{85.42} & \second{79.08} & 84.99 & \best{97.56} & \second{66.21} & \second{84.29} & -- & -- \\
    \midrule
    \multicolumn{11}{l}{\textit{Time-Series Language Models}} \\
    ChatTS & 49.42 & 52.08 & 60.43 & 58.27 & 30.29 & 50.61 & 33.59 & 49.83 & 50.72 & 24.90 \\
    TimeOmni-1-7B & 50.98 & 56.47 & 59.06 & 44.55 & 28.39 & 40.79 & 30.93 & 44.40 & 35.25 & 25.67 \\
    \midrule
    \multicolumn{11}{l}{\textit{Unified Models}} \\
    ChatTime-7B & 50.20 & 43.47 & 27.29 & 25.14 & 25.15 & 23.80 & 27.87 & 38.38 & 41.94 & 28.45 \\
    TimeOmni-VL & 50.82 & 51.15 & 44.18 & 28.87 & 23.55 & 23.63 & 15.52 & 39.27 & 49.06 & 32.32 \\
    \rowcolor{timebraidrow}
    TimeBraid-1.2B (Ours) & 82.58 & 80.78 & 77.68 & 69.20 & 77.92 & 89.43 & 45.04 & 76.61 & 50.13 & 34.17 \\
    \rowcolor{timebraidrow}
    TimeBraid-2.5B (Ours) & 83.23 & 79.25 & 80.05 & 72.86 & 82.92 & 91.71 & 48.36 & 78.31 & \second{60.05} & \second{35.74} \\
    \rowcolor{timebraidrow}
    TimeBraid-6.7B (Ours) & 83.43 & 85.78 & 81.78 & 75.12 & \second{86.39} & 92.04 & 49.39 & 80.65 & \best{63.14} & \best{36.58} \\
    \bottomrule
    \end{tabular}%
    }
\end{table}

\paragraph{Perception, Question Answering, and Reasoning.}
The time-series understanding results are summarized in Table~\ref{tab:understanding_summary}. TSAQA poses multi-format questions about a presented series, from perception subtasks such as anomaly detection and classification to reasoning subtasks such as temporal relations and ordering. TimeSeriesExam and the TemporalBench multiple-choice suite (TB-MCQ) are exams of time-series concepts and temporal reasoning. TimeBraid-6.7B achieves 80.65\% overall TSAQA accuracy, compared with 85.26\% and 84.29\% for the benchmark-specific fine-tuned LLaMA3.1-8B and Qwen3-8B ~\citep{tsaqa2026}, and 63.10\% for GPT-5.4. TimeBraid-2.5B and TimeBraid-1.2B achieve 78.31\% and 76.61\%. All three variants remain competitive on TB-MCQ, with TimeBraid-6.7B landing within two points of GPT-5.4. TimeOmni-VL obtains 39.27\% on TSAQA and 49.06\% on TimeSeriesExam, where TimeBraid-6.7B reaches 63.14\%. Per-category breakdowns are provided in Tables~\ref{tab:timeseriesexam}, \ref{tab:tsaqa_main_results}, and \ref{tab:temporalbench_all_models}.

\begin{table}[!htbp]
\centering
\caption{Contextual description on the CaTS-Bench human-rewritten split: five metrics of alignment with reference captions and the released Numeric Fidelity score, all in $[0,1]$. The Closed-source Reference Models group provides references excluded from ranking. Rows marked finetuned are trained on the CaTS-Bench training split; the full comparison is in Table~\ref{tab:catsbench_hr}.}
\label{tab:catsbench_summary}
\scriptsize
\setlength{\tabcolsep}{4pt}
\resizebox{\textwidth}{!}{%
\begin{tabular}{lcccccc}
\toprule
\multirow{2}{*}{Model} & \multicolumn{5}{c}{Lexical and Semantic Alignment} & \multirow{2}{*}{Numeric\,$\uparrow$} \\
\cmidrule(lr){2-6}
 & DeBERTa-F1\,$\uparrow$ & SimCSE\,$\uparrow$ & BLEU\,$\uparrow$ & ROUGE-L\,$\uparrow$ & METEOR\,$\uparrow$ & \\
\midrule
\multicolumn{7}{l}{\textit{Closed-source Reference Models}} \\
GPT-5.4 & 0.660 & 0.843 & 0.064 & 0.239 & 0.264 & 0.778 \\
Gemini 2.0 Flash & 0.694 & 0.884 & 0.113 & 0.283 & 0.304 & 0.757 \\
GPT-4o & 0.685 & 0.886 & 0.090 & 0.259 & 0.314 & 0.739 \\
\midrule
\multicolumn{7}{l}{\textit{Open-source Large Language Models}} \\
InternVL 2.5 38B & 0.682 & 0.867 & 0.085 & 0.262 & 0.294 & 0.740 \\
Gemma 3 27B & 0.680 & 0.883 & 0.089 & 0.255 & \best{0.307} & 0.745 \\
Idefics 2 (finetuned) & \best{0.713} & \second{0.885} & \best{0.131} & \best{0.325} & \second{0.303} & \second{0.748} \\
Llama 3.2 Vision (finetuned) & 0.672 & 0.851 & 0.087 & 0.266 & 0.295 & 0.740 \\
Phi-4 Multimodal (finetuned) & 0.664 & 0.862 & 0.078 & 0.243 & 0.290 & 0.703 \\
\midrule
\multicolumn{7}{l}{\textit{Time-Series Language Models}} \\
ChatTS & 0.614 & 0.665 & 0.040 & 0.191 & 0.198 & 0.648 \\
TimeOmni-1-7B & 0.634 & 0.796 & 0.046 & 0.205 & 0.231 & \best{0.769} \\
\midrule
\multicolumn{7}{l}{\textit{Unified Models}} \\
ChatTime-7B & 0.371 & 0.234 & 0.004 & 0.032 & 0.030 & 0.077 \\
TimeOmni-VL & 0.598 & 0.568 & 0.027 & 0.179 & 0.165 & 0.370 \\
\rowcolor{timebraidrow}
TimeBraid-1.2B (Ours) & 0.706 & 0.880 & 0.120 & 0.305 & 0.288 & 0.666 \\
\rowcolor{timebraidrow}
TimeBraid-2.5B (Ours) & 0.708 & \second{0.885} & 0.121 & 0.307 & 0.289 & 0.670 \\
\rowcolor{timebraidrow}
TimeBraid-6.7B (Ours) & \second{0.712} & \best{0.886} & \second{0.123} & \second{0.313} & 0.292 & 0.684 \\
\bottomrule
\end{tabular}%
}
\end{table}

\paragraph{Contextual Description.}
Table~\ref{tab:catsbench_summary} reports contextual description results on the CaTS-Bench human-rewritten split, where the model writes a caption for a series given its background context, scored for alignment with human references and for the fidelity of the numbers it cites. In the primary comparison, TimeBraid-6.7B ranks second on DeBERTa-F1, BLEU, and ROUGE-L and achieves the best SimCSE, while TimeBraid-2.5B remains second or third on the same alignment metrics. All three variants lead every other series--text model by a wide margin on every alignment metric. Numeric fidelity is their weakest axis, below the general-purpose models. The gap is expected: the time-series backbone compresses each patch of 32 raw values into a single embedding, and this reduction preserves shape and dynamics but loses fine-grained local values. Results for all evaluated models are reported in Table~\ref{tab:catsbench_hr}.

% Keep the understanding tables with their discussion before forecasting.
\FloatBarrier
\subsection{Time-Series Forecasting}

\paragraph{Contextual Forecasting.}
Table~\ref{tab:forecasting_realworld_summary} reports contextual forecasting on real-world benchmarks, where TimeMMD and CGTSF pair series from domains such as health, energy, and traffic with aligned text. Among the models shown in this summary table, a TimeBraid variant ranks among the top three on eight of the nine TimeMMD domains. TimeBraid attains the best Macro MSE and Macro MAE on CGTSF. Detailed results are provided in Tables~\ref{tab:timemmd_text_official} and~\ref{tab:cgtsf}.

\begin{table}[!htbp]
\centering
\caption{Real-world contextual forecasting. Every text-conditioned model receives the paired text of each window, and a starred name (PatchTST$^{*}$) marks a time-series backbone extended with a text branch. TimeMMD reports per-domain MSE averaged over four horizons; TimeBraid, the foundation models, and MIGAS-1.5 use a fixed 128-step history, while ChatTime, Aurora, and the text-conditioned supervised baselines use the official 8/36/96 lookbacks. CGTSF reports training-split-standardized MSE on MSPG, LEU, and PTF, averaged over four history-length settings. The remaining baselines, detailed results, and full model names are in Tables~\ref{tab:timemmd_text_official} and~\ref{tab:cgtsf}.}
\label{tab:forecasting_realworld_summary}
\small
\setlength{\tabcolsep}{3.6pt}
\resizebox{\textwidth}{!}{%
\begin{tabular}{lcccccccccccc}
\toprule
\multirow{2}{*}{Model} & \multicolumn{9}{c}{TimeMMD (MSE\,$\downarrow$)} & \multicolumn{3}{c}{CGTSF (MSE\,$\downarrow$)} \\
\cmidrule(lr){2-10}\cmidrule(lr){11-13}
 & Agri. & Clim. & Econ. & Ener. & Env. & Health & Sec. & Soc. & Traf. & MSPG & LEU & PTF \\
\midrule
\multicolumn{13}{l}{\textit{Time-Series Foundation Models}} \\
TimesFM$_{2.5}$ & 0.127 & \best{0.859} & \second{0.017} & 0.239 & 0.569 & \best{0.665} & 114.8 & 0.845 & 0.152 & 0.585 & 0.777 & 0.243 \\
Chronos-2 & \best{0.085} & 0.988 & \best{0.015} & 0.262 & 0.566 & 1.217 & 110.3 & 1.110 & 0.201 & 0.649 & 0.715 & 0.230 \\
Moirai & 0.102 & 0.949 & 0.020 & 0.277 & 0.581 & 1.280 & 109.8 & 0.859 & 0.163 & 1.000 & 0.681 & 0.284 \\
Sundial & 0.102 & \second{0.865} & 0.022 & 0.266 & 0.561 & 1.240 & 111.7 & 0.856 & 0.160 & 0.763 & 0.683 & 0.242 \\
TimeMoE & 0.107 & 0.935 & 0.021 & 0.277 & 0.579 & 1.229 & 110.4 & \second{0.801} & 0.167 & 1.138 & 0.698 & 0.239 \\
\midrule
\multicolumn{13}{l}{\textit{Multimodal Forecasting Models}} \\
Aurora & 0.109 & 1.428 & 0.040 & 0.367 & 0.597 & 2.000 & 120.7 & 1.155 & 0.322 & 0.789 & 0.757 & 0.336 \\
MIGAS-1.5 & \second{0.086} & 0.983 & \best{0.015} & 0.242 & 0.671 & 1.302 & 109.9 & 1.110 & 0.200 & 0.757 & 0.887 & 0.386 \\
Time-LLM & 0.098 & 1.320 & 0.031 & 0.279 & 0.519 & 1.488 & 114.9 & 1.023 & 0.223 & 0.583 & 0.699 & 0.163 \\
GPT4MTS & 0.093 & 1.262 & 0.027 & 0.266 & 0.518 & 1.519 & 111.2 & 1.014 & 0.249 & 0.567 & 0.694 & 0.180 \\
Time-VLM & 0.093 & 1.282 & 0.020 & 0.253 & \second{0.494} & 1.482 & 113.0 & 1.068 & 0.229 & 0.818 & 0.725 & 0.175 \\
PatchTST$^{*}$ & 0.093 & 1.267 & 0.019 & 0.269 & \best{0.492} & 1.556 & 111.7 & 1.110 & 0.216 & 0.602 & 0.706 & 0.179 \\
\midrule
\multicolumn{13}{l}{\textit{Unified Models}} \\
ChatTime & 0.130 & 2.279 & 0.071 & 0.365 & 1.099 & 2.516 & 144.3 & 1.227 & 0.505 & 2.727 & 1.015 & 0.359 \\
\rowcolor{timebraidrow}
TimeBraid-1.2B (Ours) & 0.116 & 0.899 & 0.025 & 0.256 & 0.516 & 1.002 & 109.0 & 0.808 & \second{0.148} & 0.472 & \second{0.642} & \second{0.136} \\
\rowcolor{timebraidrow}
TimeBraid-2.5B (Ours) & 0.137 & 0.871 & 0.025 & \second{0.235} & 0.507 & 0.956 & \second{108.9} & 0.839 & \best{0.145} & \second{0.467} & 0.644 & \best{0.133} \\
\rowcolor{timebraidrow}
TimeBraid-6.7B (Ours) & 0.144 & 0.938 & 0.018 & \best{0.231} & 0.507 & \second{0.925} & \best{108.6} & \best{0.799} & 0.153 & \best{0.459} & \best{0.641} & 0.138 \\
\bottomrule
\end{tabular}%
}
\end{table}

\paragraph{Controlled Forecasting.}
Controlled forecasting examines whether language can direct a model's numerical predictions (Table~\ref{tab:forecasting_controlled_summary}). We introduce Ctrl-F, a controlled-forecasting evaluation set built from histories in the GIFT-Eval test set~\citep{aksu2024gifteval}, with multiple deterministic, text-conditioned synthetic continuations for each history. TimeBraid-6.7B achieves 43.33\% Top-1 accuracy against a 33.33\% chance level, showing that textual conditions can guide its numerical forecasts. CAF and CiK extend the evaluation to forecasting with informative scenarios and contextual knowledge. TimeBraid-6.7B achieves the lowest pooled CRPS on CAF, while TimeBraid-2.5B performs comparably to the strongest listed time-series foundation model on CiK. We see that the strongest LLMs remain ahead on Ctrl-F and CiK because they can easily convert the control signal to output and have the advantages of deep reasoning. 

\begin{table}[!htbp]
\centering
\caption{Controlled forecasting with verifiable targets. The Closed-source Reference Models group provides references excluded from ranking. Ctrl-F: history-$z$-normalized MSE and Top-1 correct-sibling retrieval accuracy. CAF: normalized CRPS on all 904 correct-context test cases. CiK: RCRPS under official task weights. Details in Tables~\ref{tab:ctrlf},~\ref{tab:caf}, and~\ref{tab:cik_context_stratified_paper_style}.}
\label{tab:forecasting_controlled_summary}
\small
\setlength{\tabcolsep}{5pt}
\begin{tabular}{lcccc}
\toprule
\multirow{2}{*}{Model} & \multicolumn{2}{c}{Ctrl-F} & CAF & CiK \\
\cmidrule(lr){2-3}\cmidrule(lr){4-4}\cmidrule(lr){5-5}
 & MSE\,$\downarrow$ & Top-1\,(\%)\,$\uparrow$ & CRPS\,$\downarrow$ & RCRPS\,$\downarrow$ \\
\midrule
\multicolumn{5}{l}{\textit{Closed-source Reference Models}} \\
GPT-5.4 & 5.188 & 51.33 & 0.233 & 0.145 \\
GPT-4o & 5.447 & 60.00 & 0.234 & 0.257 \\
Gemini-2.5-Flash & 6.624 & 56.00 & 0.247 & 0.110 \\
\midrule
\multicolumn{5}{l}{\textit{Open-source Large Language Models}} \\
DeepSeek-Chat & \second{6.405} & \best{59.33} & 0.267 & \best{0.225} \\
\midrule
\multicolumn{5}{l}{\textit{Time-Series Foundation Models}} \\
Moirai & 8.425 & 33.33 & 0.275 & \second{0.288} \\
Chronos-2 & 8.597 & 33.33 & 0.277 & 0.295 \\
TimesFM$_{2.5}$ & 8.784 & 33.33 & 0.281 & 0.295 \\
Sundial & 8.829 & 33.33 & 0.331 & 0.314 \\
TimeMoE & 9.077 & 33.33 & 0.414 & 0.327 \\
\midrule
\multicolumn{5}{l}{\textit{Multimodal Forecasting Models}} \\
Aurora & 9.091 & 33.33 & 0.460 & 0.313 \\
MIGAS-1.5 & 8.604 & 33.33 & 0.346 & 0.351 \\
\midrule
\multicolumn{5}{l}{\textit{Unified Models}} \\
ChatTime & 9.165 & 33.33 & 0.345 & 0.344 \\
TimeOmni-VL & 10.387 & 32.00 & 0.489 & 0.655 \\
\rowcolor{timebraidrow}
TimeBraid-1.2B (Ours) & 6.826 & 40.67 & \second{0.235} & 0.301 \\
\rowcolor{timebraidrow}
TimeBraid-2.5B (Ours) & 6.972 & 40.67 & \second{0.235} & 0.289 \\
\rowcolor{timebraidrow}
TimeBraid-6.7B (Ours) & \best{6.232} & \second{43.33} & \best{0.225} & 0.294 \\
\bottomrule
\end{tabular}
\end{table}

% Keep the controlled-forecasting table with its discussion before the next topic begins.
\FloatBarrier

\paragraph{Unimodal Forecasting.}
TimeBraid also supports competitive short-term and long-term forecasting from numerical histories alone (Tables~\ref{tab:gift_eval_mase_crps} and~\ref{tab:unimodal_forecasting_summary}). On GIFT-Eval \citep{aksu2024gifteval}, which spans diverse series and forecast lengths, it outperforms the listed statistical and task-specific supervised baselines, while the strongest specialized forecasting foundation models remain ahead. On the long-horizon ETT and Weather benchmarks, TimeBraid-2.5B achieves the second-best average MSE rank and third-best average MAE rank. We see that it largely preserves the unimodal forecasting capability from TimesFM2.5, and the gap is expected to be reduced by a more balanced data recipe. 

\begin{table*}[!htbp]
  \centering
  \caption{Aggregate GIFT-Eval forecasting performance. Lower is better.}
  \label{tab:gift_eval_mase_crps}
  \begingroup
  \setlength{\tabcolsep}{4.2pt}
  \renewcommand{\arraystretch}{1.12}
  \resizebox{\textwidth}{!}{%
    % Shade only our model column; numerical emphasis still denotes rank.
    \begin{tabular}{l >{\columncolor{timebraidrow}}c *{15}{c}}
      \toprule
      \multirow{2}{*}{\textbf{Metric}}
      &
      & \multicolumn{3}{c}{\textbf{Statistical Methods}}
      & \multicolumn{4}{c}{\textbf{Task-Specific Models (Supervised)}}
      & \multicolumn{8}{c}{\textbf{Time Series Foundation Models}} \\
      \cmidrule(lr){3-5}
      \cmidrule(lr){6-9}
      \cmidrule(lr){10-17}
      % Anchor the multirow label on the last shaded row so its text stays visible.
      & \multirow{-2}{*}{\shortstack{TimeBraid\\(Ours)}}
      & Naive
      & \shortstack{Seasonal\\Naive}
      & \shortstack{Auto\\ARIMA}
      & DeepAR
      & TiDE
      & N-BEATS
      & PatchTST
      & \shortstack{TimesFM\\2.5}
      & TabPFN-TS
      & \shortstack{Chronos\\2}
      & Moirai2
      & \shortstack{Sundial\\Base}
      & TiRex
      & \shortstack{Toto-2.0\\FnF}
      & Chronicle \\
      \midrule
      \textbf{MASE}
      & 0.763
      & 1.270
      & 1.000
      & 1.074
      & 1.343
      & 1.091
      & 0.938
      & 0.849
      & 0.705
      & 0.771
      & \second{0.698}
      & 0.728
      & 0.750
      & 0.716
      & \best{0.676}
      & 1.053 \\
      \textbf{CRPS}
      & 0.546
      & 1.591
      & 1.000
      & 0.912
      & 0.853
      & 0.772
      & 0.816
      & 0.587
      & 0.490
      & 0.544
      & \second{0.485}
      & 0.516
      & 0.559
      & 0.488
      & \best{0.463}
      & 0.754 \\
      \bottomrule
    \end{tabular}%
  }
  \endgroup
\end{table*}

\begin{table*}[!htbp]
\centering
\caption{Unimodal forecasting on ETT and Weather. Each dataset entry averages MSE or MAE over horizons $\{96,192,336,720\}$. Avg.\ rank aggregates the 20 dataset--horizon settings; lower is better. Per-horizon results are in Table~\ref{tab:unimodal_forecasting}.}
\label{tab:unimodal_forecasting_summary}
\scriptsize
\setlength{\tabcolsep}{2pt}
\resizebox{\textwidth}{!}{%
% TimeBraid occupies the first MSE/MAE pair; do not shade other model columns.
\begin{tabular}{l >{\columncolor{timebraidrow}}c >{\columncolor{timebraidrow}}c *{18}{c}}
\toprule
\multirow{3}{*}{Dataset} & \multicolumn{12}{c}{\textit{Zero-shot}} & \multicolumn{8}{c}{\textit{Full-shot}} \\
\cmidrule(lr){2-13}\cmidrule(lr){14-21}
 & \multicolumn{2}{>{\columncolor{timebraidrow}}c}{TimeBraid-2.5B} & \multicolumn{2}{c}{Sundial$_{\mathrm{L}}$} & \multicolumn{2}{c}{TimeMoE$_{\mathrm{U}}$} & \multicolumn{2}{c}{Moirai$_{\mathrm{L}}$} & \multicolumn{2}{c}{TimesFM$_{2.5}$} & \multicolumn{2}{c}{Chronos-2} & \multicolumn{2}{c}{iTransformer} & \multicolumn{2}{c}{TimeMixer} & \multicolumn{2}{c}{PatchTST} & \multicolumn{2}{c}{DLinear} \\
 & MSE & MAE & MSE & MAE & MSE & MAE & MSE & MAE & MSE & MAE & MSE & MAE & MSE & MAE & MSE & MAE & MSE & MAE & MSE & MAE \\
\midrule
ETTm1 & 0.373 & 0.375 & \best{0.331} & \second{0.369} & \second{0.356} & 0.391 & 0.422 & 0.391 & 0.375 & 0.370 & 0.373 & \best{0.356} & 0.407 & 0.409 & 0.381 & 0.395 & 0.387 & 0.400 & 0.403 & 0.406 \\
ETTm2 & \second{0.256} & 0.311 & \best{0.254} & 0.315 & 0.288 & 0.344 & 0.329 & 0.343 & 0.271 & \second{0.305} & \best{0.254} & \best{0.291} & 0.288 & 0.332 & 0.275 & 0.323 & 0.280 & 0.326 & 0.350 & 0.400 \\
ETTh1 & \best{0.393} & \second{0.407} & \second{0.395} & 0.420 & 0.412 & 0.426 & 0.480 & 0.439 & 0.396 & \best{0.405} & 0.420 & \best{0.405} & 0.454 & 0.447 & 0.448 & 0.442 & 0.468 & 0.454 & 0.455 & 0.451 \\
ETTh2 & \second{0.338} & 0.377 & \best{0.334} & 0.387 & 0.371 & 0.399 & 0.367 & 0.377 & 0.343 & \second{0.369} & 0.342 & \best{0.366} & 0.383 & 0.406 & 0.364 & 0.395 & 0.386 & 0.406 & 0.558 & 0.515 \\
Weather & \second{0.228} & 0.260 & 0.238 & 0.275 & 0.256 & 0.288 & 0.264 & 0.273 & 0.240 & \second{0.255} & \best{0.219} & \best{0.241} & 0.257 & 0.278 & 0.240 & 0.271 & 0.258 & 0.280 & 0.265 & 0.316 \\
\midrule
Avg. rank & \second{2.48} & 3.45 & \best{2.23} & 4.15 & 5.20 & 6.72 & 8.05 & 6.00 & 3.83 & \second{2.25} & 2.80 & \best{1.20} & 8.10 & 8.07 & 5.67 & 5.80 & 7.72 & 7.88 & 8.93 & 9.47 \\
\bottomrule
\end{tabular}%
}
\end{table*}

% Finish benchmark tables before starting the ablation discussion.
\FloatBarrier
\subsection{Ablation Studies}
\label{sec:ablation}

\begin{figure}[htbp]
    \centering
    \includegraphics[width=\linewidth]{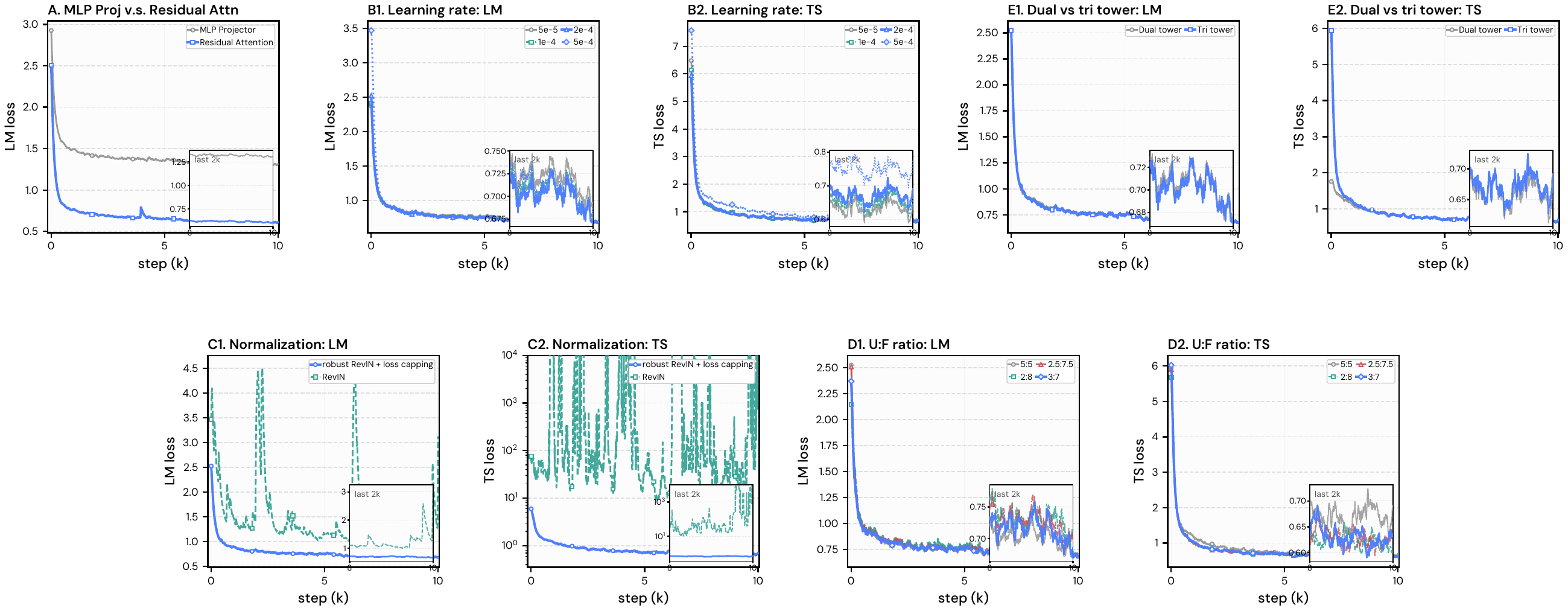}
    \caption{Training-loss trajectories for the design ablations; each panel compares 10k-step runs that differ only in the ablated factor: \textbf{(A)}~cross-modal interface, \textbf{(B)}~shared learning rate, \textbf{(C)}~plain vs.\ robust RevIN, \textbf{(D)}~understanding-to-forecasting data ratio, \textbf{(E)}~dual- vs.\ tri-tower. Insets zoom into the last 2k steps; lower is better.}
    \label{fig:ablation_loss}
\end{figure}

\begin{figure}[htbp]
    \centering
    \includegraphics[width=\linewidth]{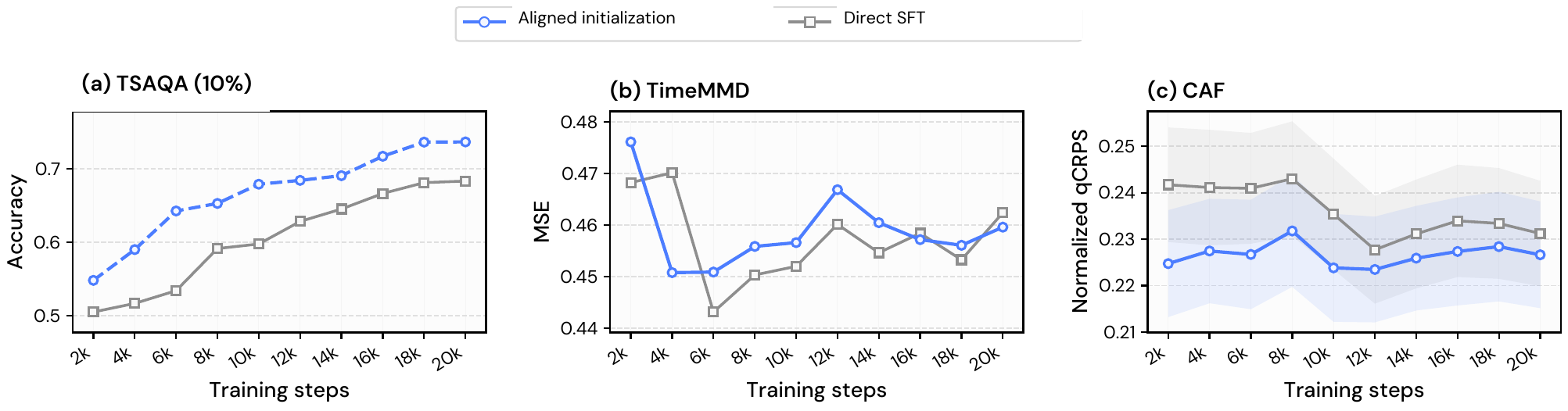}
    \caption{Effect of the alignment stage over the course of SFT. We compare aligned initialization with direct SFT from the pretrained towers on \textbf{(a)}~TSAQA accuracy (10\% test subset, higher is better), \textbf{(b)}~TimeMMD MSE (lower is better), and \textbf{(c)}~CAF normalized qCRPS (lower is better).}
    \label{fig:ablation_align}
\end{figure}

\begin{figure}[htbp]
    \centering
    \includegraphics[width=\linewidth]{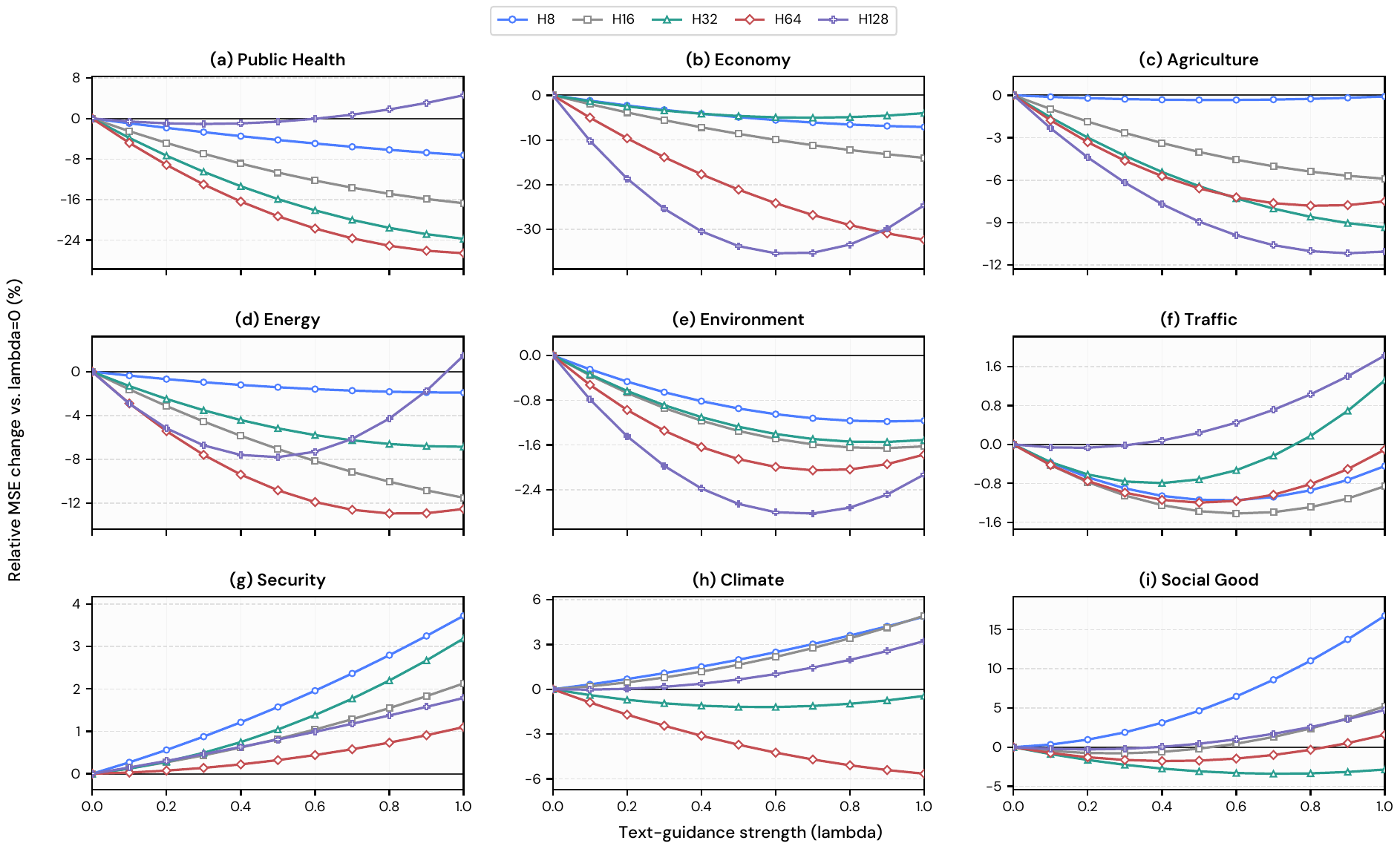}
    \caption{Effect of text guidance across domains and history lengths. Each panel sweeps the text-mixing weight $\lambda$ on one TimeMMD domain, reporting the relative MSE change against the text-free forecast ($\lambda=0$) at input histories 8--128; lower is better. At least one nonzero weight helps at every history length in Public Health, Economy, Agriculture, Energy, Environment, and Traffic; Security prefers $\lambda=0$; Climate and Social Good change sign with history. We report this sweep as a sensitivity analysis because the released text is not point-in-time certified; it is separate from model selection, for which a single $\lambda=0.3$ is selected on validation data and fixed across all domains.}
    \label{fig:ablation_text_effects}
\end{figure}

We organize the ablations around six research questions that arise when transplanting the unified-multimodal recipe to time series: where the cross-modal interaction should live (\textbf{RQ1}), whether perception and forecasting need separate towers (\textbf{RQ2}), whether the two training objectives can coexist stably (\textbf{RQ3}), how learning rate and data should be split between them (\textbf{RQ4}), whether the alignment stage is necessary at all (\textbf{RQ5}), and how strongly the forecast should draw on text at inference (\textbf{RQ6}). For RQ1--RQ4, each panel of Figure~\ref{fig:ablation_loss} compares runs that are trained for 10k steps on the alignment dataset and differ only in the ablated factor, and we analyze the loss curves of language modeling loss and time-series loss. For RQ5, we compare full recipes on downstream benchmarks (Figure~\ref{fig:ablation_align}); for RQ6, we sweep the inference-time text weight of the final model (Figure~\ref{fig:ablation_text_effects}).

\paragraph{RQ1: Interaction Space.}
We pivot the investigation of the proper interaction space on understanding tasks. The mainstream multimodal information fusion for understanding in vision--language models is an MLP connector that projects the visual modality into the language representation space \citep{liu2023llava}, and some time-series language models inherit this recipe \citep{xie2025chatts}. Whether the language space is an equally good host for temporal representations has not been fully examined. Panel~A compares the VLM-like MLP projector with the global residual attention we used in the final architecture. With the projector, the language loss plateaus early at roughly twice the level that residual attention reaches, and the gap never closes. We observe that the language space is a poor projection target for temporal representations, and the time-series patches are often projected to be global statistics such as mean and standard deviation. The two pretrained models align much more easily in a new shared space that favors neither geometry, which also aligns with the previous findings in the limits of alignment in vision, language, and time-series representations \citep{yashwante2026time}.

\paragraph{RQ2: Expert Separation.}
Unified vision models find that understanding and generation favor different representations and decouple them accordingly \citep{deng2025bagel, chen2025januspro}, which suggests giving perception and forecasting separate time-series towers. Panel~E tests this hypothesis: the tri-tower variant initializes an independent perception expert and forecasting expert from the pretrained time-series models, while the dual-tower model uses the same time-series model to process both perception and forecasting streams. The two trajectories are nearly identical on both losses throughout training. Unlike vision understanding and generation, a shared temporal representation space is capable of processing information for perception and forecasting. We leave the study of using different time-series foundation models for different experts for future work.

\paragraph{RQ3: Optimization Stability.}
The regression loss suffers from its scale-sensitive property. On strongly non-stationary segments, where future values depart from the history statistics, the time-series loss can grow arbitrarily large. Panel~C shows that this failure mode is real. With plain causal RevIN \citep{kim2022revin}, the time-series loss spikes recurrently over several orders of magnitude (C2), and every spike disrupts the language loss through the shared updates (C1). We find that these large spikes often cause the pretrained language representation to be corrupted. Robust normalization together with loss capping removes both symptoms and keeps the time-series loss stably and quietly optimized.

\paragraph{RQ4: Balancing the Two Modalities.}
Time-series and language have different internal properties and learning dynamics. It is critical to harmonize their joint training to prevent malignant competition. Panel~B sweeps a learning rate shared by both towers during joint training. The language loss is stable across the whole grid (B1), whereas the time-series loss clearly degrades at the largest rate (B2); the joint training therefore receives an $\alpha=0.5$ time-series loss weight in the final recipe. Panel~D varies the understanding-to-forecasting data ratio. Every forecasting-heavy mixture (3:7 to 2:8) reaches the same lower time-series loss, while the balanced 5:5 mixture is clearly worse (D2), and the language loss is unchanged across all ratios (D1). We therefore adopt 3:7, which keeps the largest understanding share without giving up any forecasting optimization, in effect the Pareto point of the tested range. All three optimization ablations lead to the same conclusion: the time-series objective is far more sensitive than the language objective, and protecting it with bounded targets, a smaller learning rate, and a larger data share does not make language learning better.

\paragraph{RQ5: Necessity of the Alignment Stage.}
Unified multimodal models conventionally align the modalities on paired data before instruction tuning \citep{liu2023llava,deng2025bagel}, but the alignment stage is costly, and its necessity is rarely tested. Figure~\ref{fig:ablation_align} compares runs that share the same SFT recipe and differ in initialization, using either the aligned model or the pretrained towers directly. On understanding (a), the aligned initialization stays ahead at every checkpoint, with the largest margins in early training. On TimeMMD (b), the two runs are hard to separate: the curves cross repeatedly and end at nearly the same MSE. Much of TimeMMD's text only weakly constrains the future (RQ6), so the numeric-history skills that SFT alone teaches already cover this benchmark setting. On CAF (c), where the text verifiably specifies the future, the benefit is persistent: the aligned run keeps a lower qCRPS at every checkpoint, and SFT alone never closes the gap. We keep the stage in the final recipe because alignment therefore pays where the forecast must draw on language and adds a lasting head start on understanding.

\paragraph{RQ6: Text Effects on Forecasting across Domains and History Lengths.}
Figure~\ref{fig:ablation_text_effects} shows that no single text weight works across all domains. Larger $\lambda$ usually helps Economy, Public Health, Agriculture, and Energy; Traffic gains little at smaller weights, Security does not improve, and Climate and Social Good change with history length. The first group receives reports on trade, petroleum, broiler markets, or influenza that address the forecast target or a direct driver. The weaker pairs are less direct. Traffic pairs national monthly vehicle miles traveled with local road counts, air travel, and long-term plans. Security pairs monthly FEMA grant amounts, which contain rare event-driven spikes, with past disasters and grant rules that do not specify the next spike's timing or size. Environment pairs volatile daily New York AQI with broad or out-of-state reports; Climate and Social Good sometimes use reports from another period or region \citep{liu2024timemmd}. Text can therefore add little because it does not match the target, or because it does not specify an abrupt future change. In domains with weak or noisy context, a longer history can be more reliable than text-heavy mixing: Security is best without text, while Traffic and Public Health retain only small text weights at $H=128$. In contrast, Economy, Agriculture, and Environment continue to benefit from larger text weights.

\FloatBarrier

\section{Related Work}
\label{sec:related_work}

\paragraph{Unified Multimodal Models.}
Unified multimodal models differ chiefly in how a discrete language backbone hosts a continuous modality \citep{nativemm2026roadmap,beyondlm2026}.
One line quantizes every input into a shared token vocabulary and trains a single next-token predictor \citep{chameleonteam2024chameleon,anygpt2024,wang2024emu3}; a second mixes autoregressive text prediction with diffusion-style generation inside one transformer, increasingly over continuous latents \citep{zhou2024transfusion,xie2024showo,deng2025bagel,xie2025showo2}; a third decouples the representations serving understanding from those serving generation \citep{chen2025januspro}.
Omni-modal systems extend these recipes to speech, audio, and video \citep{qwen3omni2025,minicpmo2026,dyninomni2026}.
Two lessons recur across this progression: continuous signals lose fidelity when forced through a discrete vocabulary \citep{fan2025fluid}, and understanding and generation favor different treatments of the same modality.
Yet time series, for all their ubiquity, remain largely absent from these systems.
TimeBraid carries both lessons to this missing modality: it aligns time series with language in continuous space, avoiding quantization at both input and output, and gives perception and forecasting different input processing while sharing one time-series backbone.

\paragraph{Time-Series Foundation Models.}
Pretraining on large signal corpora yields foundation models that forecast zero-shot across domains \citep{garza2023timegpt,rasul2023lagllama,das2024timesfm,ansari2024chronos,woo2024moirai}.
The family has since diversified along axes familiar from language modeling: sparse experts \citep{shi2025timemoe}, long-context and serial scaling \citep{liu2025timerxl,timers12026}, generative decoding \citep{liu2025sundial}, compact multi-task backbones \citep{goswami2024moment,ekambaram2024ttm,gao2024units,wang2025timefound}, and domain specialization \citep{cohen2024toto}.
Yet the interface stays numeric: domain knowledge and instructions have no way in and explanations no way out, even though textual context can materially improve forecasts \citep{williams2025cik,liu2024timemmd}.
TimeBraid builds on this line, initializing its perception and forecasting experts from pretrained time-series blocks to inherit their temporal competence, while adding the language interface and world knowledge these models lack.

\paragraph{Coupling Time Series with Language.}
Work coupling time series with language typically picks one pretrained host and moves the other modality into it.
Understanding models host series in an LLM through an attached encoder or projector \citep{xie2025chatts,timeomni2025,wang2025itformer,langer2026opentslm}, even when that encoder is itself a pretrained time-series foundation model \citep{tsreasoner}.
Forecasting models host series in a language or vision--language model via digit serialization \citep{gruver2023llmtime}, input reprogramming or adaptation \citep{jin2024timellm,llm4ts2024,liu2024unitime}, or rendered images \citep{zhong2025timevlm}; conversely, context-aided forecasters host text in a numeric model through an added text branch \citep{liu2024timemmd}.
All of these cover only one direction of the interface, and the move itself is lossy: alignment between time-series, vision, and language representations has clear limits \citep{yashwante2026time}, and on purely numeric benchmarks the language host adds little \citep{tan2024llmuseful}.
The few models covering both directions pay a different price: ChatTime hosts values in an LLM vocabulary as discrete tokens \citep{wang2025chattime}, TimeOmni-VL hosts them in pixel space as rendered plots \citep{timeomnivl2026}, and the concurrent Chronicle learns a joint space from scratch with 324M parameters, forgoing pretrained competence on both sides \citep{chronicle2026}.
TimeBraid instead keeps each modality in its own pretrained host and fuses the two in a shared representation space, covering both directions without forcing either modality into the other's geometry.

\section{Limitations}

TimeBraid is our early attempt toward unified time-series and language models. It is trained without a continued pretraining stage. Public interleaved series--text corpora at pretraining scale are not yet mature, so our recipe moves directly from alignment to supervised fine-tuning. Arguably, large-scale continued pretraining would instill broader and richer world knowledge and improve the model across the board.

Perception carries bounded numeric precision. The model sometimes hallucinates the value at a specific point and misreads complex waveforms (long, high-frequency, or low signal-to-noise). Training coverage contributes, but the main reason is the patch tokenization inherited from the time-series backbone, which compresses many raw values into each embedding and does not fully preserve grounded local detail.

Forecast control weakens when text contradicts history. The forecasting expert inherits the transition dynamics of its pretrained backbone, which bias it toward historically consistent continuations. Direct control signals that demand behavior at odds with the history (an abrupt level shift, a sudden regime break) are followed loosely. Furthermore, large-scale high-quality real-world text-conditioned forecasting data and multi-modal multi-variate data are not readily available in public access. When the input text guidance and content are highly complex and noisy, TimeBraid cannot completely translate it to effective forecasting signals. However, we see that using a frontier language model for reasoning and instruction conversion helps mitigate these issues, and it implies the need for combination with agents for a better forecasting system.

\section{Conclusion}

We presented TimeBraid, a unified time-series and language model that aligns the native representations of a pretrained language model and pretrained time-series foundation models through tri-expert global residual attention. Trained with a dual-stage recipe on curated alignment pairs and a broad instruction mixture, one set of weights understands and generates both modalities, forecasting at the level of dedicated time-series foundation models while remaining competitive with far larger general-purpose models on understanding and context-aided forecasting. Our ablations distill the design choices that make this model class work: where the cross-modal interaction should live, whether perception and forecasting need separate experts, and how strongly the forecast should draw on text at inference. We hope TimeBraid is a step toward unified models that treat time-series as an essential modality, understood and generated in its native representation.

% The venue-specific AI-use disclosure is omitted from the public preprint.
% Restore the input below in the ICLR submission version after author approval.
% \input{ai-use-statement}

\bibliographystyle{iclr2027_conference}

\bibliography{references}

\clearpage
\appendix
% Appendix figures are pinned in place with [H]; without \raggedbottom the
% style's \flushbottom spreads the resulting slack between headings and text.
\raggedbottom
\startcontents[appendices]
\section*{Appendix Contents}
\vspace{0.5em}
{
\setcounter{tocdepth}{2}
\titlecontents{section}
  [1.5em]
  {\addvspace{6pt}\bfseries}
  {\contentslabel{1.5em}}
  {\hspace*{-1.5em}}
  {\titlerule*[0.6em]{.}\contentspage}
\titlecontents{subsection}
  [3.8em]
  {\addvspace{2pt}}
  {\contentslabel{2.3em}}
  {\hspace*{-2.3em}}
  {\titlerule*[0.6em]{.}\contentspage}
\printcontents[appendices]{}{1}{}
}
\vspace{0.75em}
\noindent\rule{\textwidth}{0.4pt}
\clearpage

\section{Data Curation Pipeline}
\label{app:data_curation}

\begin{figure}[H]
    \centering
    \includegraphics[width=\linewidth]{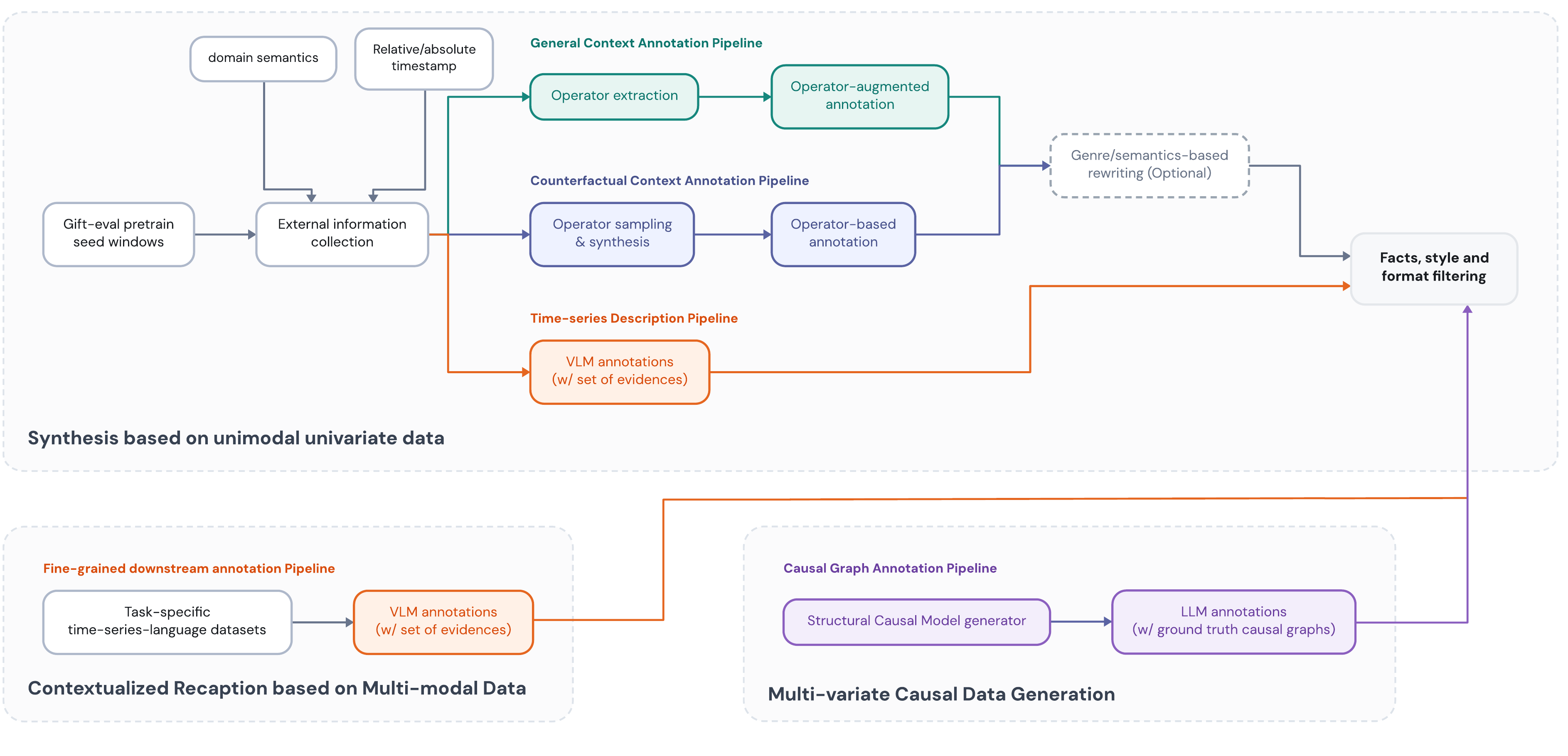}
    \caption{Alignment data curation for univariate understanding, multivariate understanding, and univariate forecasting.}
    \label{fig:data_pipeline}
\end{figure}

Figure~\ref{fig:data_pipeline} presents our data curation pipelines for univariate understanding, multivariate understanding, and univariate forecasting. Each pipeline pairs time series with language through a capability-specific annotation process, followed by shared filtering for factual consistency, style, and format.

\subsection{Univariate Understanding}
Univariate understanding data teaches the model to recognize temporal patterns and express them in language. We use vision-language models as the primary annotators because their exposure to charts and visual data equips them to perceive global morphology, such as trends, seasonality, and regime changes, together with local events such as peaks, troughs, and anomalies. Their annotations distill this temporal knowledge into textual supervision. For real series, each evidence pack combines the observed trace with deterministic statistics, extrema, trend segments, change points, periodic patterns, and temporal landmarks; the context-rich route additionally supplies domain semantics and calendar information. Separately, we adopt the attribute-first synthesis scheme of ChatTS \citep{xie2025chatts}, in which sampled temporal attributes determine both the generated series and its programmable description. We also rewrite procedurally generated exam prompts on pattern recognition, noise, similarity, anomalies, and Granger-causal structure into concise descriptive-answer supervision. All outputs pass factual, stylistic, and formatting filters.

\subsection{Multivariate Understanding}
Real multivariate time series rarely carry detailed, verifiable annotations of interactions across variables. We retain the cleaned multivariate descriptions from ChatTS \citep{xie2025chatts} for broad descriptive coverage and construct graph-grounded relational supervision with temporal structural equation models. The source generator samples sparse or random lagged graphs over 3 to 50 observed variables with maximum lags from 2 to 50 steps, then simulates their multivariate trajectories. Each evidence pack records the graph-authorized variable roles, lag offsets, intervention metadata when present, temporal anchors, and canonical evidence for direct parent-to-target relations, short directed chains, localized shock follow-ons, historical reference windows from the same SCM, and local parent/child motifs. A text-only LLM converts this evidence into natural descriptions under automatic grounding checks. The active mixture contains 100{,}000 SCM descriptions, including 74{,}210 positive relation examples and 25{,}790 comparison controls whose channels are not directly adjacent to the target.

\subsection{Univariate Forecasting}
Paired data that connects an observed history, a textual condition, and its corresponding future is scarce in natural corpora. We construct this supervision through the general-context and counterfactual-context pipelines in Figure~\ref{fig:data_pipeline}. The general branch annotates the realized suffix of a real window with its domain semantics and temporal information. The control branch produces three sibling continuations from one history: a baseline and two controlled variants, each composing up to three transformations drawn from level shifts, ramps, slope changes, pulses, seasonal-amplitude changes, saturations, shock decays, and analog replays. It renders the applied transformations as natural-language conditions, so the shared history can lead to distinct text-specified futures. The model input contains the observed history and the natural-language continuation condition, while time-series loss applies only to the held-out suffix. Optional genre and semantic rewrites diversify the conditions while preserving their connection to the target future. Both pipelines apply the shared factual, stylistic, and formatting filters.

\FloatBarrier

\section{Training Data}
\label{app:data_inventory}

Stage 1 contains 2{,}229{,}500 alignment examples, grouped into 12 families along the three curation routes of Appendix~\ref{app:data_curation}. Stage 2 uses a final inventory of 4{,}881{,}583 samples across five composition categories, with alignment retention restricted to its recipe-defined quota.

\subsection{Alignment Corpus Composition}

Table~\ref{tab:alignment_inventory} lists the 12 alignment families. The understanding and forecasting halves are fixed at 3:7 by construction (668{,}850 and 1{,}560{,}650 rows). Each forecasting branch appears in three variants: a base variant with horizon at most 512, a rewrite of the same rows into domain-grounded real-world context with horizon at most 32, and a partial genre rewrite under the same horizon limit. Control rows are organized into three-sibling groups that share one history: one neutral continuation and two operator-conditioned futures.

\begin{table}[H]
\centering
\small
\caption{Alignment corpus: 12 families, 2{,}229{,}500 examples. $H$ is the forecast horizon. Rewrite variants reuse the base rows and change only the user-visible text.}
\label{tab:alignment_inventory}
\begin{tabular}{llrr}
\toprule
Family & Curation route & $n$ & Share \\
\midrule
\multicolumn{4}{l}{\textit{Understanding}\quad 668{,}850 rows, 30.0\%} \\
\addlinespace[1pt]
\quad Morphology captions & morphology-grounded VLM annotation & 50{,}000 & 2.2\% \\
\quad Context-rich captions & context-grounded VLM annotation & 50{,}000 & 2.2\% \\
\quad ChatTS \citep{xie2025chatts} univariate & attribute-programmed synthesis & 260{,}000 & 11.7\% \\
\quad ChatTS \citep{xie2025chatts} multivariate & attribute-programmed synthesis & 108{,}850 & 4.9\% \\
\quad Multivariate SCM & SCM-grounded relation synthesis & 100{,}000 & 4.5\% \\
\quad TimeSeriesExam descriptions & procedural task synthesis & 100{,}000 & 4.5\% \\
\midrule
\multicolumn{4}{l}{\textit{Forecasting, control branch}\quad 668{,}291 rows, 30.0\%} \\
\addlinespace[1pt]
\quad Control, $H\!\le\!512$ & operator-controlled future synthesis & 338{,}953 & 15.2\% \\
\quad Control, $H\!\le\!32$ & domain-grounded rewriting & 164{,}918 & 7.4\% \\
\quad Control, $H\!\le\!32$ & genre-conditioned rewriting & 164{,}420 & 7.4\% \\
\midrule
\multicolumn{4}{l}{\textit{Forecasting, general branch}\quad 892{,}359 rows, 40.0\%} \\
\addlinespace[1pt]
\quad General, $H\!\le\!512$ & realized-future annotation & 528{,}507 & 23.7\% \\
\quad General, $H\!\le\!32$ & domain-grounded rewriting & 255{,}291 & 11.5\% \\
\quad General, $H\!\le\!32$ & genre-conditioned rewriting & 108{,}561 & 4.9\% \\
\bottomrule
\end{tabular}
\end{table}

\subsection{Supervised Fine-Tuning Sources}

Table~\ref{tab:sft_inventory} reports the Stage-2 data distribution.

\begingroup
\scriptsize
\setlength{\tabcolsep}{4pt}
\setlength{\LTleft}{\fill}
\setlength{\LTright}{\fill}
\begin{longtable}{lrr}
\caption{Final one-pass Stage-2 sample inventory and training-time sampling shares. References identify the source datasets; counts refer to our selected and reformatted examples.}
\label{tab:sft_inventory}\\
\toprule
Source & Final samples & Training Share \\
\midrule
\endfirsthead
\multicolumn{3}{l}{\footnotesize Table~\thetable\ continued from the previous page.}\\
\toprule
Source & Final samples & Training Share \\
\midrule
\endhead
\midrule
\multicolumn{3}{r}{\footnotesize Continued on the next page.}\\
\endfoot
\bottomrule
\endlastfoot
\multicolumn{3}{l}{\textit{Forecasting}\quad 2{,}640{,}566 samples, 54.49\%} \\
\addlinespace[1pt]
\quad CGTSF \citep{wang2025chattime} & 53,926 & 2.04\% \\
\quad FinMultiTime S\&P 500 \citep{xu2025finmultitime} & 144,387 & 5.46\% \\
\quad MoTime (News/Wiki) \citep{zhou2025motime} & 32,589 & 1.23\% \\
\quad CAF-7M \citep{zheng2026caf7m} & 2,314,216 & 42.15\% \\
\quad Time-MMD \citep{liu2024timemmd} & 20,768 & 0.79\% \\
\quad Time-MQA Forecasting \citep{kong2025timemqa} & 74,680 & 2.82\% \\
\midrule
\multicolumn{3}{l}{\textit{Time-Series QA \& Reasoning}\quad 1{,}065{,}021 samples, 22.49\%} \\
\addlinespace[1pt]
\quad CaTS-Bench \citep{catsbench2025} & 15,995 & 0.33\% \\
\quad ChatTS \citep{xie2025chatts} & 50,377 & 1.05\% \\
\quad HiTSR \citep{ding2026llatisa} & 157,147 & 3.24\% \\
\quad OpenSQA \citep{imran2024llasa} & 138,754 & 2.72\% \\
\quad OpenTSLM ECG \citep{langer2026opentslm} & 159,313 & 3.28\% \\
\quad OpenTSLM HAR \citep{langer2026opentslm} & 68,542 & 1.41\% \\
\quad OpenTSLM Sleep \citep{langer2026opentslm} & 7,434 & 0.15\% \\
\quad OpenTSLM TSQA \citep{langer2026opentslm} & 38,400 & 0.79\% \\
\quad Time-MQA Reasoning \citep{kong2025timemqa} & 110,212 & 2.27\% \\
\quad TSAQA Reasoning \citep{tsaqa2026} & 125,957 & 2.60\% \\
\quad EngineMT-QA \citep{wang2025itformer} & 100,296 & 2.07\% \\
\quad RATs40K \citep{yang2026timera} & 32,994 & 0.68\% \\
\quad Time-MQA Imputation \citep{kong2025timemqa} & 38,607 & 1.46\% \\
\quad TSAQA Anomaly Detection \citep{tsaqa2026} & 20,993 & 0.43\% \\
\midrule
\multicolumn{3}{l}{\textit{Captioning \& Description}\quad 161{,}883 samples, 3.34\%} \\
\addlinespace[1pt]
\quad OpenTSLM M4 Captioning \citep{langer2026opentslm} & 80,000 & 1.65\% \\
\quad SensorCaps \citep{imran2024llasa} & 35,960 & 0.74\% \\
\quad Time-MMD Open-Ended \citep{liu2024timemmd} & 1,925 & 0.04\% \\
\quad TRACE Captioning \citep{chen2025trace} & 43,998 & 0.91\% \\
\midrule
\multicolumn{3}{l}{\textit{Unimodal Corpora}\quad 400{,}000 samples, 11.69\%} \\
\addlinespace[1pt]
\quad GIFT-Eval Pretrain \citep{aksu2024gifteval} & 200,000 & 7.56\% \\
\quad Dolci-Instruct (No-Tools) \citep{olmo2025olmo3} & 200,000 & 4.12\% \\
\midrule
\multicolumn{3}{l}{\textit{Alignment Retention}\quad 614{,}113 samples, 8.00\%} \\
\addlinespace[1pt]
\quad Understanding Retention & 116,294 & 1.51\% \\
\quad Forecasting Retention & 497,819 & 6.48\% \\
\midrule
\textbf{Final inventory} & \textbf{4,881,583} & \textbf{100.00\%} \\
\end{longtable}
\endgroup

\subsection{Overlap with Evaluation Benchmarks}
\label{app:data_overlap}

To ensure a valid evaluation, we strictly enforce no overlap with the training mixture at the split level, keeping evaluation examples held out even when related training data originate from the same benchmark family or underlying data-generating process. Table~\ref{tab:eval_overlap} summarizes this relationship for each benchmark.

\begin{table}[H]
\centering
\small
\caption{Training status of the evaluation suites. Status is determined by split membership in the mixture.}
\label{tab:eval_overlap}
\begin{tabular}{p{3.6cm}p{8.8cm}}
\toprule
Benchmark & Status \\
\midrule
\textit{Understanding} & \\
\addlinespace[1pt]
TemporalBench (TB-MCQ) & Out-of-domain. \\
TSExam & Independently generated and rewritten samples from the same data-generating process are included in the alignment mixture. \\
TSAQA & Training split is included in the SFT mixture. \\
CaTS-Bench & Training split is included in the SFT mixture. \\
\midrule
\textit{Forecasting} & \\
\addlinespace[1pt]
ETT and Weather & Out-of-domain. \\
CiK & Out-of-domain. \\
Time-MMD & Training split is included in the SFT mixture. \\
CGTSF & Training split is included in the SFT mixture. \\
CAF & Part of the training split is included in the SFT mixture. \\
Ctrl-F & Independently generated samples from the same data-generating process are included in the alignment mixture. \\
GIFT-Eval & Part of the training split is used as seed data for data generation. \\
\bottomrule
\end{tabular}
\end{table}

\FloatBarrier

\section{Implementation Details}
\label{app:implementation}

\subsection{Model Configuration}

Table~\ref{tab:param_inventory} gives the parameter allocation across Qwen3 language towers \citep{yang2025qwen3}, TimesFM-derived time-series towers \citep{das2024timesfm}, and fusion stacks. Variant names round the total parameter count, including frozen weights and excluding buffers; shared weights are counted once.

\begin{table}[H]
\centering
\caption{Parameter inventory across TimeBraid scales, in billions of parameters.}
\label{tab:param_inventory}
\small
\begin{tabular}{lrrrr}
\toprule
Variant & Language tower & Fusion & Time-series tower & Total \\
\midrule
TimeBraid-1.2B & 0.596 & 0.378 & 0.231 & 1.205 \\
TimeBraid-2.5B & 1.721 & 0.545 & 0.231 & 2.497 \\
TimeBraid-6.7B & 4.022 & 2.265 & 0.389 & 6.676 \\
\bottomrule
\end{tabular}
\end{table}

\paragraph{Geometry and Layer Pairing.}
TimeBraid-1.2B and TimeBraid-2.5B interleave 20 time-series and fusion blocks across 28 language layers, while TimeBraid-6.7B pairs 36 time-series and fusion blocks with its 36 language layers. Time-series segments use patches of $P=32$ values, and each paired layer exchanges information through global residual attention.

\begin{algorithm}[t]
\caption{Pseudocode of TimeBraid's forward pass. Every language layer runs its native block. At paired positions, the corresponding time-series block runs and both streams exchange information through global residual attention (Eqs.~\ref{eq:fusion_proj} and~\ref{eq:fusion_attn}); at any unpaired position, the language stream proceeds alone.}
\label{alg:mot_forward}
\begin{lstlisting}
# H_l: (n_l, d_l) language hidden states; H_t: (n_t, d_t) time-series hidden states
# tau_l, tau_t: timeline positions of each stream (global chronological order)
# pair_of[i]: time-series layer paired with language layer i, or None;
#   smaller variants have unpaired language layers, while 6.7B pairs all 36 layers
# Wq_r, Wk_r, Wv_r: (d_r, d) projections to the shared width; Wo_r: (d, d_r), zero-init
# phi_r: hidden-state RMSNorm; psi_r: per-head QK RMSNorm (r in {l, t})

def forward(H_l, H_t, tau_l, tau_t):
    for i in range(num_lm_layers):
        H_l = lm_block[i](H_l)          # every language layer runs its native block (RoPE inside)
        if pair_of[i] is None:
            continue                    # unpaired layer: the language stream continues alone
        j = pair_of[i]
        H_t = ts_block[j](H_t)          # paired time-series layer (segment-local positions)
        H_l, H_t = fuse[j](H_l, H_t, tau_l, tau_t)   # global residual attention
    return H_l, H_t

def fuse(H_l, H_t, tau_l, tau_t):
    # project both streams into the shared attention width d (values: no head norm)
    q_l, k_l, v_l = psi_l(phi_l(H_l) @ Wq_l), psi_l(phi_l(H_l) @ Wk_l), phi_l(H_l) @ Wv_l
    q_t, k_t, v_t = psi_t(phi_t(H_t) @ Wq_t), psi_t(phi_t(H_t) @ Wk_t), phi_t(H_t) @ Wv_t

    # reorder into one global sequence by timeline position tau
    q, k, v, tau = merge_by_position([q_l, q_t], [k_l, k_t], [v_l, v_t], [tau_l, tau_t])

    # joint causal attention with RoPE on the shared global axis
    o = attention(rope(q, tau), rope(k, tau), v, mask="causal")

    # route rows back to their streams; zero-init Wo_r makes fusion start as identity
    o_l, o_t = split_by_stream(o)
    return H_l + o_l @ Wo_l, H_t + o_t @ Wo_t
\end{lstlisting}
\end{algorithm}

\subsection{Training Hyperparameters}

Both stages update every parameter of the model (language tower, time-series tower, and the global residual attention blocks) under the settings in Table~\ref{tab:hyperparams}. The two stages share the optimizer, schedule, batch size, and objective constants; they differ in context length, step budget, and data mixture.

\begingroup
\setlength{\LTleft}{\fill}
\setlength{\LTright}{\fill}
\begin{longtable}{lcc}
\caption{Training hyperparameters for both stages. Rows spanning both columns are identical across stages.}
\label{tab:hyperparams}\\
\toprule
 & Stage 1 (alignment) & Stage 2 (supervised fine-tuning) \\
\midrule
\endfirsthead
\multicolumn{3}{l}{\footnotesize Table~\thetable\ continued from the previous page.}\\
\toprule
 & Stage 1 (alignment) & Stage 2 (supervised fine-tuning) \\
\midrule
\endhead
\midrule
\multicolumn{3}{r}{\footnotesize Continued on the next page.}\\
\endfoot
\bottomrule
\endlastfoot
Optimizer & \multicolumn{2}{c}{AdamW, 8-bit states} \\
Learning rate & \multicolumn{2}{c}{$2\times10^{-5}$, all parameters} \\
Schedule & \multicolumn{2}{c}{constant with warmup} \\
Warmup steps & \multicolumn{2}{c}{500} \\
Weight decay & \multicolumn{2}{c}{0} \\
Gradient clipping & \multicolumn{2}{c}{1.0} \\
Precision & \multicolumn{2}{c}{BF16 mixed precision} \\
Random seed & \multicolumn{2}{c}{3407} \\
\midrule
Global batch size & \multicolumn{2}{c}{128} \\
Sequence packing & \multicolumn{2}{c}{disabled} \\
\midrule
Patch length $P$ & \multicolumn{2}{c}{32} \\
Time-series loss weight $\alpha$ & \multicolumn{2}{c}{0.5} \\
Point loss & \multicolumn{2}{c}{squared error} \\
Quantile set $\mathcal{Q}$ & \multicolumn{2}{c}{$\{0.1,0.2,\dots,0.9\}$} \\
Loss cap threshold $c$ & \multicolumn{2}{c}{2.0} \\
Target transform & \multicolumn{2}{c}{$\operatorname{asinh}$ on causal RevIN residuals} \\
\midrule
Context length & 2{,}560 & 4{,}096 \\
Training data & 2.23M alignment pairs & 4.88M final samples \\
Training steps & 17k steps & 30k steps \\
\end{longtable}
\endgroup

\FloatBarrier

\section{Prompt Template Examples}
\label{app:prompt_templates}

Examples are wrapped in Qwen3 chat templates and a series of time-series payloads. The transcript itself contains no time-series numbers: each segment appears in the text as the empty pair \texttt{<ts></ts>}, and the $N$-th occurrence binds to the $N$-th payload. \texttt{<ts>} and \texttt{</ts>} are added to the vocabulary as ordinary tokens, \texttt{<stats>} and \texttt{</stats>} stay plain text, and only assistant content receives cross-entropy.

A context segment is preceded by \texttt{<stats>len=$L$, mean=$\mu$, std=$\sigma$</stats>}, computed on the raw values before normalization, so the text stream carries the physical scale while the patch stream carries only normalized shape. A target segment emits a bare \texttt{<ts></ts>} and never receives a statistics block, so the model is not told the scale of the sequence it must produce. Target segments span the full window $[\text{history}\mid\text{future}]$ and reuse the mean and standard deviation of their context segment, which keeps every future-aware statistic out of the input, and only the future suffix is scored. Exam-style questions are the one family whose payload keeps raw values, because their options refer to absolute magnitudes.

Each card below is one real alignment row, one per task family. Inside every \texttt{<ts></ts>} we plot the payload it binds to: \textcolor{tsinclr}{\textbf{teal}} for observed and context spans, \textcolor{tsoutclr}{\textbf{orange}} for the future the assistant must generate.

\begin{tcolorbox}[promptcard, title=Univariate description]
\textbf{User:} Treat this cloud operations series from Borg Cluster Data 2011 as sampled every 5 minutes. \texttt{<stats>len=79, mean=0.0671359, std=0.0062208</stats>} \texttt{<ts>}\raisebox{-1.05ex}{\includegraphics[height=3.4ex]{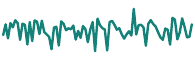}}\texttt{</ts>} Provide a brief written summary of the series.
\tcblower
\textbf{Assistant:} A sharp, isolated excursion dominates the middle of the window, contrasting with the otherwise stable baseline. The series holds a steady level near 0.065 until step 39, where it drops abruptly to a local minimum of 0.052. This is immediately followed by a steep recovery to 0.076 at step 40, creating a narrow V-shape. The signal then drifts upward, reaching the window's global maximum of 0.083 at step 55, before settling into a gradual decline toward the end. The amplitude of this central event is significantly larger than the ambient noise observed in the first and last thirds of the trace.
\end{tcolorbox}

\begin{tcolorbox}[promptcard, title=Context-rich description]
\textbf{User:} Describe the time series of fatalities due to a specific severity level in Porterville city. The data spans from 2003 to 2008 with a yearly frequency, covering a total population of 41,774. \texttt{<stats>len=6, mean=3.167, std=1.067</stats>} \texttt{<ts>}\raisebox{-1.05ex}{\includegraphics[height=3.4ex]{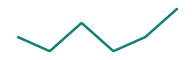}}\texttt{</ts>}

\textbf{Relevant context:} Time series of fatalities (severity: Killed) in Porterville city, CA. Data range: 2003-2008. Frequency: Yearly. Total population: 41,774.
\tcblower
\textbf{Assistant:} Porterville city recorded annual fatalities (severity: Killed) between 2003 and 2008 among a population of 41,774. The series began at 3.0 in 2003 and reached its minimum of 2.0 in 2004. The data then exhibited high volatility, spiking to 4.0 in 2005 before dropping back to 2.0 in 2006. Following this trough, the series trended upward, reaching a net increase of 2.0 over the period and closing at its maximum value of 5.0 in 2008.
\end{tcolorbox}

\begin{tcolorbox}[promptcard, title=Time-series question answering]
\textbf{User:} You are given a pair of synthetic time series generated by procedural time-series templates.

Time series 1: \texttt{<stats>len=296, mean=3.11567, std=2.01952</stats>} \texttt{<ts>}\raisebox{-1.05ex}{\includegraphics[height=3.4ex]{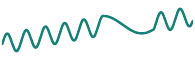}}\texttt{</ts>}

Time series 2: \texttt{<stats>len=236, mean=2.34241, std=1.70843</stats>} \texttt{<ts>}\raisebox{-1.05ex}{\includegraphics[height=3.4ex]{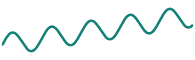}}\texttt{</ts>}

Instruction: Identify which time series contains the anomaly and describe the anomaly type based on the visible statistics and series patterns. Respond with one concise declarative sentence.
\tcblower
\textbf{Assistant:} Time series 1 has a speed up/down anomaly, meaning the period of its cyclic components differs from the rest of the series.
\end{tcolorbox}

\begin{tcolorbox}[promptcard, title=Multivariate relational understanding]
\textbf{User:} In a Web Servers system, there are 13 metrics:\\
Request Rates is of length 256: \texttt{<stats>len=256, mean=-0.4419, std=0.420386</stats>} \texttt{<ts>}\raisebox{-1.05ex}{\includegraphics[height=3.4ex]{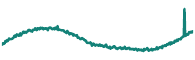}}\texttt{</ts>};\\
Backend Server Latency is of length 256: \texttt{<stats>len=256, mean=5.0224, std=1.20905</stats>} \texttt{<ts>}\raisebox{-1.05ex}{\includegraphics[height=3.4ex]{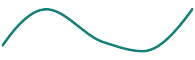}}\texttt{</ts>};\\
\emph{[6 metrics elided, same form]}\\
Security Event Logs is of length 256: \texttt{<stats>len=256, mean=-119.453, std=155.282</stats>} \texttt{<ts>}\raisebox{-1.05ex}{\includegraphics[height=3.4ex]{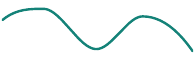}}\texttt{</ts>};\\
\emph{[4 metrics elided]}. Analyze the trend of Security Event Logs in detail.
\tcblower
\textbf{Assistant:} From point 11 to point 55, there is an increasing trend. From point 55 to point 126, there is a decreasing trend. From point 126 to point 188, there is an increasing trend. From point 188 to point 245, there is a decreasing trend.
\end{tcolorbox}

\begin{tcolorbox}[promptcard, title=Contextual forecasting]
\textbf{User:} We have been tracking our annual volume for the last five years: \texttt{<stats>len=5, mean=4165.4, std=175.205</stats>} \texttt{<ts>}\raisebox{-1.05ex}{\includegraphics[height=3.4ex]{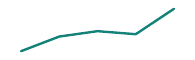}}\texttt{</ts>}. Historically, we've seen similar growth patterns during previous expansion phases where a slow start was followed by a rapid surge in demand. Based on current market projections and our scaling strategy, we expect a similar acceleration over the next five years, starting with a modest increase to around 4,701 in year 6 and climbing steeply to reach approximately 5,622 by year 10. Please provide the full 10-year sequence, including the observed history and the forecasted values.
\tcblower
\textbf{Assistant:} \texttt{<ts>}\raisebox{-1.05ex}{\includegraphics[height=3.4ex]{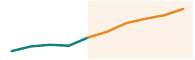}}\texttt{</ts>}
\end{tcolorbox}

% This long three-example card may span a page instead of leaving the preceding
% page half empty when the complete box no longer fits.
\begin{tcolorbox}[promptcard, breakable, title={Controllable forecasting, three sibling rows}]
\textbf{User:} Local tech monitors are tracking weekly visitor traffic for the site, with current data showing \texttt{<stats>len=6, mean=525.667, std=302.232</stats>} \texttt{<ts>}\raisebox{-1.05ex}{\includegraphics[height=3.4ex]{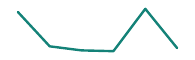}}\texttt{</ts>}. With a major marketing push and site relaunch peaking in about four weeks, officials expect a significant surge in traffic, potentially reaching around 869 visitors. To assist with server capacity planning, please provide the full sequence of weekly traffic, including the observed data and the forecast for the next 5 weeks.

\smallskip
\noindent\textbf{Assistant:} \texttt{<ts>}\raisebox{-1.05ex}{\includegraphics[height=3.4ex]{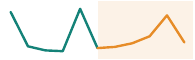}}\texttt{</ts>}
\tcbline
\textbf{User:} Team,

I'm reviewing our weekly page views for the current seasonal cycle to coordinate our upcoming resource allocation. We've seen some volatility lately, but typically this period involves a dip followed by a recovery as our annual event nears.

Based on the last six weeks \texttt{<stats>len=6, mean=525.667, std=302.232</stats>} \texttt{<ts>}\raisebox{-1.05ex}{\includegraphics[height=3.4ex]{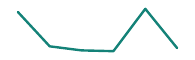}}\texttt{</ts>}, we expect a similar pattern. We anticipate traffic hitting a low of around 348 next week, climbing to a peak of about 661 by week four, and then dropping back to roughly 339 in week five.

Please provide the full sequence of 11 weekly values, including the observed data and the five-week forecast, so we can finalize the schedule.

\smallskip
\noindent\textbf{Assistant:} \texttt{<ts>}\raisebox{-1.05ex}{\includegraphics[height=3.4ex]{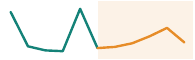}}\texttt{</ts>}
\tcbline
\textbf{User:} Weekly Page View Log -- Landing Page

Current Status: Tracking traffic trends for the landing page prior to decommissioning.

Recent Activity: \texttt{<stats>len=6, mean=525.667, std=302.232</stats>} \texttt{<ts>}\raisebox{-1.05ex}{\includegraphics[height=3.4ex]{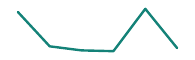}}\texttt{</ts>}

Note: This page is scheduled for decommissioning starting next week. Traffic is expected to drop immediately to nearly zero and remain flat once the URL redirect is active.

Please provide the completed 11-week sequence, including the observed data and the projected values for the final 5 weeks.

\smallskip
\noindent\textbf{Assistant:} \texttt{<ts>}\raisebox{-1.05ex}{\includegraphics[height=3.4ex]{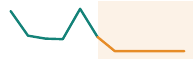}}\texttt{</ts>}
\end{tcolorbox}

\FloatBarrier

\section{Evaluation Protocols}
\label{app:eval_protocols}

For every public benchmark, we use its official split or released evaluation set and follow its official evaluation protocol and input-length setting unless stated otherwise. Ctrl-F uses our frozen evaluation set. Each paragraph below states whether baseline results are taken from an existing study or evaluated by us.

\paragraph{TemporalBench.}
TemporalBench is a multi-domain benchmark over real numerical series that separates historical-structure interpretation (T1), context-free prediction (T2), context-grounded reasoning (T3), and event-conditioned forecasting (T4), testing whether agents can align temporal patterns with external context and adapt when conditions change. Multiple-choice tasks use accuracy; forecasting uses domain-specific MAE or sMAPE variants, where sMAPE normalizes absolute error by the magnitudes of the target and prediction and MIMIC uses the benchmark's observation-weighted aggregation. We evaluate every model ourselves on the official TemporalBench evaluation set \citep{temporalbench2026} (Table~\ref{tab:temporalbench_all_models}).

\paragraph{TimeSeriesExam.}
TimeSeriesExam is a configurable, procedurally generated multiple-choice exam in which one or two controlled synthetic series, represented as plots or serialized values, are paired with answer options and an in-context example to test pattern and noise understanding, anomaly detection, comparative reasoning, and Granger-causality reasoning. We use the official evaluation set \citep{cai2024timeseriesexam} and take the shared baseline results from the TS-Reasoner paper \citep{tsreasoner}. GPT-5.4, Gemini-2.5-Flash, TimeOmni-1-7B, TimeOmni-VL, and TimeBraid are additional evaluations performed by us on the same released set with its scorer (Table~\ref{tab:timeseriesexam}).

\paragraph{TSAQA.}
TSAQA formulates each instance as a numerical series, natural-language context, and a true-or-false, multiple-choice, or puzzling question, spanning anomaly detection and classification together with characterization, comparison, data transformation, and temporal-relation reasoning. Puzzling questions ask the model to reorder four shuffled successor patches and are scored by the fraction placed in the correct position. We use the official test set \citep{tsaqa2026}; except for GPT-5.4, the LLM and finetuned baseline results are taken from the TSAQA paper, while GPT-5.4, the time-series language models, and the unified models are evaluated by us on the same 42{,}000 cases with the paper-defined scoring rule (Table~\ref{tab:tsaqa_main_results}).

\paragraph{CaTS-Bench.}
In its primary captioning task, CaTS-Bench pairs raw time-indexed values with contextual metadata, a line plot, and a standardized instruction, asking models to synthesize numerical, visual, and semantic evidence into a grounded description evaluated against human-rewritten references. Embedding and lexical metrics measure semantic and wording consistency with the reference; Numeric Fidelity extracts numbers from the reference and output and combines their matching precision and recall. We use the official human-rewritten evaluation set \citep{catsbench2025}, take the available baseline results from the CaTS-Bench paper, and evaluate GPT-5.4, the time-series language models, and the unified models ourselves on the same set with the six released-formula local metrics (Table~\ref{tab:catsbench_hr}); the separate Gemini-dependent Numeric Score 2.0 is not a column in that table.

\paragraph{CiK.}
CiK comprises 71 probabilistic forecasting tasks across seven domains, each paired with historical, future, covariate, causal, or intemporal natural-language context that is essential beyond the observed numerical history, and tests whether models integrate both modalities. RCRPS emphasizes context-relevant future regions, penalizes violations of textual constraints, and normalizes scores by task scale; we aggregate it using the official task weights. We evaluate every model ourselves on the official 355-instance CiK evaluation set \citep{williams2025cik} (Table~\ref{tab:cik_context_stratified_paper_style}).

\paragraph{TimeMMD.}
Time-MMD evaluates multimodal forecasting across nine real-world domains by pairing numerical histories with temporally aligned historical text at four frequency-dependent horizons, testing whether extra-numerical domain knowledge improves prediction. We report each domain's MSE and MAE macro-averaged over its four horizons. We use the official evaluation split \citep{liu2024timemmd}. Every row of Table~\ref{tab:timemmd_text_official} is evaluated by us, with each multimodal forecasting model receiving the window's paired text: GPT4MTS, TaTS, Time-VLM \citep{zhong2025timevlm}, the text-enhanced PatchTST$^{*}$ (VoT style \citep{wang2026unlocking}), iTransformer$^{*}$, and RaFT$^{*}$, and the additional DLinear, Reformer (MMTSFlib style \citep{liu2024timemmd}), and Time-LLM runs. TimeBraid, all time-series foundation models, and MIGAS-1.5 use a fixed 128-step lookback; the text-conditioned baselines, ChatTime, and Aurora use the official lookbacks of 8, 36, and 96. We select a single text-mixing weight of $\lambda=0.3$ on the validation split and then hold it fixed across all nine domains and history lengths, rather than tuning it separately for each dataset; this uniform setting simplifies evaluation and tests generality across domains.

\paragraph{CGTSF.}
CGTSF uses the chronological 6:2:2 splits of MSPG, LEU, and PTF, pairing histories with leakage-controlled background, weather-forecast, and calendar information to test the auxiliary role of text in forecasting at four history lengths per dataset. We standardize errors using training-split statistics and report Macro MSE and Macro MAE as the equal averages across the twelve dataset$\times$history-length settings. We evaluate every model ourselves on the released CGTSF datasets using the paper-defined splits \citep{wang2025chattime} (Table~\ref{tab:cgtsf}). We adopt the same implementation of the dataset-specific baselines as in the Time-MMD evaluation.

\paragraph{CAF.}
CAF-7M evaluates context-aided probabilistic forecasting on verified history--scenario--future windows from datasets reserved for testing, with contexts designed to provide information complementary to the numerical history. Normalized CRPS evaluates the full predictive distribution after adjusting for target scale. We use the official evaluation split \citep{zheng2026caf7m}; DoubleCast and TimeLLM are taken from the CAF-7M paper, and all remaining rows are evaluated by us under the same official protocol (Table~\ref{tab:caf}).

\paragraph{Ctrl-F.}
Ctrl-F is our controlled-future test set, built on numerical histories replayed from held-out GIFT-Eval test anchors. Each history is paired with three deliberately separated synthetic future siblings and a condition identifying the intended continuation, testing whether forecasts follow the condition rather than the shared history alone. The sibling futures are constructed counterfactual targets rather than official GIFT targets, generated with the same procedure used for the control rows of the alignment corpus. Top-1 is correct when a prediction has lower MAE to its intended sibling than to either alternative, with a chance level of $33.33\%$. We evaluate every model ourselves on the fixed Ctrl-F evaluation set using the same evaluation protocol (Table~\ref{tab:ctrlf}).

\begin{tcolorbox}[promptcard, breakable, title={Ctrl-F instance: one 22-step hospital history, three sibling conditions at horizon 12}]
\textbf{User:} \texttt{(history=22, horizon=12)}

Use this continuation condition: The series drops sharply to a local minimum at future step 3, followed by a volatile recovery that peaks at future steps 4 and 9 before ending with a slight upward trend.

History: \texttt{<stats>len=22, mean=53.1818, std=8.89656</stats>} \texttt{<ts>}\raisebox{-1.05ex}{\includegraphics[height=3.4ex]{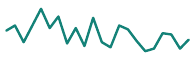}}\texttt{</ts>}

\smallskip
\noindent\textbf{Assistant:} \texttt{<ts>}\raisebox{-2.6ex}{\includegraphics[height=7ex]{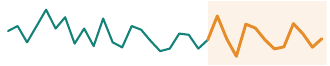}}\texttt{</ts>}
\tcbline
\textbf{User:} \emph{[identical, except]} Use this continuation condition: Maintain a strong downward drift across future steps 1 to 12.

\smallskip
\noindent\textbf{Assistant:} \texttt{<ts>}\raisebox{-2.6ex}{\includegraphics[height=7ex]{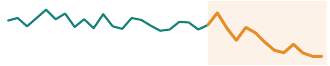}}\texttt{</ts>}
\tcbline
\textbf{User:} \emph{[identical, except]} Use this continuation condition: Create a local downward pulse in future steps 3 to 8. Cap upper peaks in future steps 1 to 12.

\smallskip
\noindent\textbf{Assistant:} \texttt{<ts>}\raisebox{-2.6ex}{\includegraphics[height=7ex]{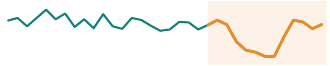}}\texttt{</ts>}
\end{tcolorbox}

\paragraph{GIFT-Eval.}
GIFT-Eval evaluates general zero-shot point and probabilistic forecasting across 97 dataset--frequency--horizon configurations spanning univariate and multivariate series, seven domains, ten sampling frequencies, and short-to-long horizons, targeting generalization across heterogeneous forecasting settings. The official leaderboard's MASE scales absolute point error by seasonal-naive error, while CRPS evaluates the predictive distribution; both are normalized against Seasonal Naive per configuration and aggregated by geometric mean, so lower is better and 1.0 matches Seasonal Naive. We take baseline results from the official leaderboard \citep{aksu2024gifteval} and evaluate TimeBraid using the last 2{,}880 history points (Table~\ref{tab:gift_eval_mase_crps}).

\paragraph{Long-Term Forecasting.}
Long-term forecasting follows the standard multivariate setting on ETTm1, ETTm2, ETTh1, ETTh2, and the 21-variable Jena Weather dataset at horizons $\{96,192,336,720\}$, testing numerical-history-only extrapolation of trends, periodic structure, and long-range dependencies \citep{zhou2021informer,wu2021autoformer}. TimeBraid-2.5B and the time-series foundation models are evaluated zero-shot, while the task-specific supervised baselines are trained on each dataset. Most baseline results are taken from the Sundial and Time-MoE papers \citep{liu2025sundial,shi2025timemoe}. We additionally evaluate TimesFM 2.5 and Chronos-2 ourselves, using the same 2{,}880-step lookback windows as TimeBraid (Table~\ref{tab:unimodal_forecasting}).

\section{Detailed Experimental Results}
\label{app:detailed_results}
\enlargethispage{\baselineskip}

This appendix reports the benchmark-level results summarized in the main text. Unless stated otherwise, higher is better for accuracy and similarity metrics, whereas lower is better for forecasting errors and ranks. Across tables, bold, underline, and a dagger mark the best, second-best, and third-best distinct values within the primary comparison, excluding the separately listed Closed-source Reference Models; ties share a rank. Dashes mark unavailable results. Shading identifies TimeBraid.

\subsection{Time-Series Understanding}

\begin{table}[H]
\centering
\caption{TimeSeriesExam accuracy (\%) by category: pattern recognition (PR), noise understanding (NU), anomaly detection (AD), similarity analysis (SA), causality analysis (CA), and overall (OA).}
\label{tab:timeseriesexam}
\footnotesize
\setlength{\tabcolsep}{6pt}
\begin{tabular}{lcccccc}
\toprule
Model & PR & NU & AD & SA & CA & OA \\
\midrule
\multicolumn{7}{l}{\textit{Closed-source Reference Models}} \\
GPT-5.4 & 69.34 & 64.29 & 70.37 & 74.17 & 50.00 & 67.83 \\
GPT-4o & 59.03 & 55.17 & 53.49 & 62.83 & 31.75 & 55.96 \\
Gemini-2.5-Flash & 54.97 & 54.76 & 47.22 & 63.33 & 41.67 & 53.89 \\
GPT-4o (vision) & 67.12 & 62.07 & 62.79 & 64.60 & 26.98 & 62.12 \\
GPT-4.1 (vision) & 69.81 & 68.97 & 68.22 & 75.22 & 41.27 & 67.89 \\
\midrule
\multicolumn{7}{l}{\textit{Open-source Large Language Models}} \\
DeepSeek-Chat & \third{65.23} & 55.17 & \best{52.71} & \third{63.71} & \third{42.86} & \third{59.89} \\
Llama-3.1-8B-Instruct & 37.73 & 37.93 & 30.23 & 36.28 & 28.57 & 35.52 \\
Qwen2.5-7B-Instruct & 47.17 & 47.13 & 41.86 & 53.10 & 41.27 & 46.66 \\
\midrule
\multicolumn{7}{l}{\textit{Open-source Vision-Language Models}} \\
Qwen2.5-VL-7B-Instruct & 25.34 & 32.18 & 19.38 & 42.48 & 12.70 & 26.61 \\
InternVL3-8B & 50.13 & 52.87 & 43.41 & 54.87 & 38.09 & 49.01 \\
\midrule
\multicolumn{7}{l}{\textit{Time-Series Language Models}} \\
ChatTS & 50.13 & 50.57 & \second{50.38} & 61.95 & 34.92 & 50.72 \\
TS-Reasoner-7B & 52.29 & \second{60.92} & \best{52.71} & \second{64.60} & 41.27 & 54.26 \\
TimeOmni-1-7B & 42.82 & 48.81 & 22.22 & 24.17 & 19.44 & 35.25 \\
\midrule
\multicolumn{7}{l}{\textit{Unified Models}} \\
ChatTime-7B & 42.85 & 49.42 & 35.65 & 44.24 & 34.92 & 41.94 \\
TimeOmni-VL & 50.83 & 57.14 & 36.11 & 54.17 & 41.67 & 49.06 \\
\rowcolor{timebraidrow}
TimeBraid-1.2B (Ours) & 55.25 & 48.81 & 37.04 & 54.17 & 38.89 & 50.13 \\
\rowcolor{timebraidrow}
TimeBraid-2.5B (Ours) & \second{67.13} & \third{59.52} & \third{46.30} & 58.33 & \best{48.61} & \second{60.05} \\
\rowcolor{timebraidrow}
TimeBraid-6.7B (Ours) & \best{69.89} & \best{65.48} & 42.59 & \best{69.17} & \second{47.22} & \best{63.14} \\
\bottomrule
\end{tabular}
\end{table}

\begin{table}[H]
% Keep the expanded TSAQA table separate so both result tables fit on the page.
\centering
\caption{TSAQA test-set accuracy (\%). A.D.: anomaly detection; CLS: classification; TF/MC/PZ: true-or-false, multiple-choice, and puzzling/ordering formats. SFT denotes TSAQA-specific LoRA fine-tuning \citep{tsaqa2026}. Closed-source Reference Models use zero-shot evaluation. Best, second, and third results are marked within the primary comparison, including the TSAQA-specific SFT rows and excluding Closed-source Reference Models; ties share a rank.}
\label{tab:tsaqa_main_results}
\scriptsize
\setlength{\tabcolsep}{3pt}
\resizebox{\textwidth}{!}{%
\begin{tabular}{lcccccccccccc}
\toprule
\multirow{2}{*}{Model} & \multicolumn{1}{c}{A.D.} & \multicolumn{1}{c}{CLS} & \multicolumn{2}{c}{Characterization} & \multicolumn{2}{c}{Comparison} & \multicolumn{2}{c}{Data Transform} & \multicolumn{3}{c}{Temporal Relation} & \multirow{2}{*}{Overall} \\
\cmidrule(lr){2-2}\cmidrule(lr){3-3}\cmidrule(lr){4-5}\cmidrule(lr){6-7}\cmidrule(lr){8-9}\cmidrule(lr){10-12}
 & TF & MC & TF & MC & TF & MC & TF & MC & TF & MC & PZ & \\
\midrule
\multicolumn{13}{l}{\textit{Closed-source Reference Models}} \\
GPT-5.4 & 53.32 & 49.57 & 82.08 & 81.98 & 77.59 & 74.47 & 65.79 & 54.94 & 74.12 & 82.96 & 51.02 & 63.10 \\
GPT-4.1 & 55.85 & 50.38 & 92.97 & 89.36 & 83.57 & 76.99 & 54.36 & 51.13 & 65.90 & 79.09 & 45.77 & 62.82 \\
GPT-4o & 54.32 & 47.20 & 88.15 & 84.15 & 78.61 & 69.07 & 60.66 & 53.24 & 62.25 & 75.58 & 45.61 & 60.73 \\
Claude-3.5-Sonnet & 51.27 & 41.23 & 74.39 & 78.45 & 66.59 & 74.14 & 65.79 & 57.07 & 82.05 & 82.15 & 54.56 & 61.19 \\
Gemini-2.5-Flash & 52.08 & 49.07 & 85.48 & 81.08 & 77.79 & 72.21 & 63.62 & 60.17 & 75.05 & 84.49 & 60.84 & 65.08 \\
\midrule
\multicolumn{13}{l}{\textit{Open-source Large Language Models}} \\
Qwen3-8B & 50.60 & 50.52 & 77.35 & 66.87 & 71.04 & 63.21 & 52.43 & 34.46 & 65.22 & 67.14 & 21.93 & 51.04 \\
LLaMA3.1-8B & 54.92 & 50.20 & 68.10 & 62.26 & 67.84 & 49.98 & 51.90 & 36.56 & 54.82 & 40.95 & 6.80 & 44.93 \\
Ministral-8B & 53.35 & 34.08 & 71.06 & 63.93 & 47.54 & 52.90 & 50.70 & 25.28 & 50.58 & 33.88 & 30.77 & 44.65 \\
Qwen3-0.6B & 50.40 & 35.83 & 62.00 & 48.78 & 58.03 & 37.51 & 49.03 & 23.62 & 51.99 & 37.33 & 13.38 & 39.06 \\
LLaMA3.2-1B & 49.47 & 39.48 & 63.74 & 52.55 & 61.02 & 36.82 & 48.87 & 4.20 & 48.97 & 5.44 & 6.76 & 35.70 \\
Gemma3-1B & 49.15 & 49.83 & 63.74 & 47.71 & 61.19 & 43.37 & 49.37 & 24.88 & 49.42 & 25.84 & 23.97 & 43.03 \\
LLaMA3.1-8B (SFT) & \best{91.02} & \best{91.27} & \best{92.44} & \second{83.68} & \best{86.72} & \best{79.31} & \best{90.17} & \best{86.62} & \best{96.94} & \second{97.41} & \best{67.68} & \best{85.26} \\
Qwen3-8B (SFT) & \second{87.70} & \second{90.05} & \second{92.37} & \best{85.42} & \second{86.55} & \second{79.08} & \second{89.84} & \third{84.99} & \second{96.84} & \best{97.56} & \second{66.21} & \second{84.29} \\
Ministral-8B (SFT) & 71.56 & 74.28 & 91.31 & 80.78 & 84.14 & 74.63 & 75.15 & 71.61 & \third{94.07} & \third{94.15} & 56.82 & 74.74 \\
Qwen3-0.6B (SFT) & \third{83.68} & \third{85.78} & 89.38 & 74.87 & 80.65 & 64.84 & 80.51 & 73.28 & 93.92 & 93.79 & \third{63.34} & 78.32 \\
LLaMA3.2-1B (SFT) & 83.08 & 83.83 & 87.71 & 74.37 & 78.61 & 60.88 & 68.09 & 51.67 & 91.39 & 88.81 & 57.53 & 73.48 \\
Gemma3-1B (SFT) & 83.10 & 84.05 & 87.88 & 72.54 & 78.61 & 59.31 & 64.06 & 45.23 & 91.00 & 88.05 & 42.92 & 69.70 \\
\midrule
\multicolumn{13}{l}{\textit{Time-Series Language Models}} \\
ChatTS & 49.42 & 52.08 & 69.20 & 60.43 & 67.13 & 58.27 & 49.97 & 30.29 & 54.04 & 50.61 & 33.59 & 49.83 \\
TimeOmni-1-7B & 50.98 & 56.47 & 39.93 & 59.06 & 40.34 & 44.55 & 50.70 & 28.39 & 50.58 & 40.79 & 30.93 & 44.40 \\
\midrule
\multicolumn{13}{l}{\textit{Unified Models}} \\
ChatTime-7B & 50.20 & 43.47 & 48.92 & 27.29 & 50.49 & 25.14 & 49.53 & 25.15 & 49.81 & 23.80 & 27.87 & 38.38 \\
TimeOmni-VL & 50.82 & 51.15 & 41.92 & 44.18 & 40.71 & 28.87 & 53.23 & 23.55 & 58.33 & 23.63 & 15.52 & 39.27 \\
\rowcolor{timebraidrow}
TimeBraid-1.2B (Ours) & 82.58 & 80.78 & 89.71 & 77.68 & 81.80 & 69.20 & 80.28 & 77.92 & 89.62 & 89.43 & 45.04 & 76.61 \\
\rowcolor{timebraidrow}
TimeBraid-2.5B (Ours) & 83.23 & 79.25 & 90.64 & 80.05 & 83.23 & 72.86 & 83.41 & 82.92 & 89.71 & 91.71 & 48.36 & 78.31 \\
\rowcolor{timebraidrow}
TimeBraid-6.7B (Ours) & 83.43 & \third{85.78} & \third{91.41} & \third{81.78} & \third{85.94} & \third{75.12} & \third{86.98} & \second{86.39} & 92.05 & 92.04 & 49.39 & \third{80.65} \\
\bottomrule
\end{tabular}%
}
\end{table}

\begin{table}[H]
\caption{Full results on the CaTS-Bench human-rewritten split.}
\label{tab:catsbench_hr}
\centering
\scriptsize
\setlength{\tabcolsep}{3pt}
\resizebox{\textwidth}{!}{%
\begin{tabular}{lcccccc}
\toprule
\multirow{2}{*}{Model} & \multicolumn{5}{c}{Lexical and Semantic Metrics} & Numeric \\
\cmidrule(lr){2-6}
 & DeBERTa F1 & SimCSE & BLEU & ROUGE-L & METEOR & \\
\midrule
\multicolumn{7}{l}{\textit{Closed-source Reference Models}} \\
GPT-5.4 & 0.660 & 0.843 & 0.064 & 0.239 & 0.264 & 0.778 \\
Gemini 2.0 Flash & 0.694 & 0.884 & 0.113 & 0.283 & 0.304 & 0.757 \\
GPT-4o & 0.685 & 0.886 & 0.090 & 0.259 & 0.314 & 0.739 \\
\midrule
\multicolumn{7}{l}{\textit{Open-source Vision-Language Models}} \\
InternVL 2.5 38B & 0.682 & 0.867 & 0.085 & 0.262 & 0.294 & 0.740 \\
Gemma 3 27B & 0.680 & \third{0.883} & 0.089 & 0.255 & \best{0.307} & \third{0.745} \\
LLaVA v1.6 (finetuned) & 0.644 & 0.802 & 0.064 & 0.234 & 0.250 & 0.589 \\
Idefics 2 (finetuned) & \best{0.713} & \second{0.885} & \best{0.131} & \best{0.325} & \second{0.303} & \second{0.748} \\
QwenVL (finetuned) & 0.678 & 0.863 & 0.090 & 0.257 & 0.272 & 0.669 \\
Llama 3.2 Vision (finetuned) & 0.672 & 0.851 & 0.087 & 0.266 & \third{0.295} & 0.740 \\
Phi-4 Multimodal Instruct (finetuned) & 0.664 & 0.862 & 0.078 & 0.243 & 0.290 & 0.703 \\
\midrule
\multicolumn{7}{l}{\textit{Time-Series Language Models}} \\
ChatTS & 0.614 & 0.665 & 0.040 & 0.191 & 0.198 & 0.648 \\
TimeOmni-1-7B & 0.634 & 0.796 & 0.046 & 0.205 & 0.231 & \best{0.769} \\
\midrule
\multicolumn{7}{l}{\textit{Unified Models}} \\
ChatTime-7B & 0.371 & 0.234 & 0.004 & 0.032 & 0.030 & 0.077 \\
TimeOmni-VL & 0.598 & 0.568 & 0.027 & 0.179 & 0.165 & 0.370 \\
\rowcolor{timebraidrow}
TimeBraid-1.2B (Ours) & 0.706 & 0.880 & 0.120 & 0.305 & 0.288 & 0.666 \\
\rowcolor{timebraidrow}
TimeBraid-2.5B (Ours) & \third{0.708} & \second{0.885} & \third{0.121} & \third{0.307} & 0.289 & 0.670 \\
\rowcolor{timebraidrow}
TimeBraid-6.7B (Ours) & \second{0.712} & \best{0.886} & \second{0.123} & \second{0.313} & 0.292 & 0.684 \\
\bottomrule
\end{tabular}%
}
\end{table}

\subsection{Time-Series Forecasting}

\subsubsection{Contextual Reasoning Forecasting}

\begin{table}[H]
\centering
\caption{Results on CiK: RCRPS weighted by the official task weights, reported with standard errors and stratified by context type. Asterisks mark models that do not use natural-language context.}
\label{tab:cik_context_stratified_paper_style}
\scriptsize
\setlength{\tabcolsep}{3pt}
\resizebox{\textwidth}{!}{%
\begin{tabular}{lcccccc}
\toprule
\textsc{Model} & \shortstack{\textsc{Average}\\\textsc{RCRPS}} & \shortstack{\textsc{Intemporal}\\\textsc{Information}} & \shortstack{\textsc{Historical}\\\textsc{Information}} & \shortstack{\textsc{Future}\\\textsc{Information}} & \shortstack{\textsc{Covariate}\\\textsc{Information}} & \shortstack{\textsc{Causal}\\\textsc{Information}} \\
\midrule
\multicolumn{7}{l}{\textit{Closed-source Reference Models}} \\
\quad Gemini-2.5-Flash & 0.110 $\pm$ 0.002 & 0.134 $\pm$ 0.003 & 0.142 $\pm$ 0.001 & 0.036 $\pm$ 0.001 & 0.116 $\pm$ 0.003 & 0.252 $\pm$ 0.012 \\
\quad GPT-5.4 & 0.145 $\pm$ 0.000 & 0.182 $\pm$ 0.001 & 0.116 $\pm$ 0.001 & 0.049 $\pm$ 0.000 & 0.160 $\pm$ 0.001 & 0.414 $\pm$ 0.001 \\
\quad GPT-4o & 0.257 $\pm$ 0.001 & 0.305 $\pm$ 0.002 & 0.135 $\pm$ 0.001 & 0.159 $\pm$ 0.000 & 0.238 $\pm$ 0.001 & 0.613 $\pm$ 0.006 \\
\quad GPT-5.4-mini & 0.277 $\pm$ 0.000 & 0.302 $\pm$ 0.001 & 0.173 $\pm$ 0.001 & 0.212 $\pm$ 0.000 & 0.261 $\pm$ 0.001 & 0.553 $\pm$ 0.000 \\
\midrule
\multicolumn{7}{l}{\textit{Open-source Large Language Models}} \\
\quad DeepSeek-Chat & \best{0.225 $\pm$ 0.004} & 0.286 $\pm$ 0.006 & 0.141 $\pm$ 0.001 & \best{0.089 $\pm$ 0.008} & \best{0.166 $\pm$ 0.001} & \best{0.431 $\pm$ 0.005} \\
\quad Qwen-2.5-7B-Inst & 0.344 $\pm$ 0.001 & 0.368 $\pm$ 0.002 & 0.195 $\pm$ 0.001 & \third{0.283 $\pm$ 0.002} & 0.333 $\pm$ 0.001 & 0.562 $\pm$ 0.001 \\
\quad Llama-3-8B-Inst & 0.476 $\pm$ 0.003 & 0.573 $\pm$ 0.005 & 0.245 $\pm$ 0.001 & 0.301 $\pm$ 0.000 & 0.486 $\pm$ 0.004 & 0.526 $\pm$ 0.001 \\
\midrule
\multicolumn{7}{l}{\textit{Time-Series Foundation Models}} \\
\quad Moirai-2$^{*}$ & \second{0.288 $\pm$ 0.002} & \best{0.252 $\pm$ 0.003} & \second{0.119 $\pm$ 0.002} & 0.366 $\pm$ 0.004 & 0.244 $\pm$ 0.003 & \third{0.435 $\pm$ 0.008} \\
\quad TimesFM$^{*}$ & 0.295 $\pm$ 0.002 & \second{0.265 $\pm$ 0.003} & \best{0.114 $\pm$ 0.001} & 0.373 $\pm$ 0.002 & 0.251 $\pm$ 0.002 & 0.474 $\pm$ 0.006 \\
\quad Sundial$^{*}$ & 0.314 $\pm$ 0.001 & 0.285 $\pm$ 0.002 & \second{0.119 $\pm$ 0.001} & 0.381 $\pm$ 0.001 & 0.263 $\pm$ 0.001 & 0.486 $\pm$ 0.004 \\
\quad TimeMoE$^{*}$ & 0.327 $\pm$ 0.001 & 0.303 $\pm$ 0.001 & 0.150 $\pm$ 0.000 & 0.387 $\pm$ 0.002 & 0.271 $\pm$ 0.001 & 0.488 $\pm$ 0.000 \\
\quad Chronos-2$^{*}$ & 0.295 $\pm$ 0.003 & 0.271 $\pm$ 0.005 & 0.183 $\pm$ 0.007 & 0.354 $\pm$ 0.004 & \third{0.243 $\pm$ 0.003} & 0.445 $\pm$ 0.012 \\
\midrule
\multicolumn{7}{l}{\textit{Traditional Time-Series Models}} \\
\quad DLinear$^{*}$ & 0.337 $\pm$ 0.001 & 0.337 $\pm$ 0.001 & 0.230 $\pm$ 0.000 & 0.345 $\pm$ 0.003 & 0.284 $\pm$ 0.001 & 0.548 $\pm$ 0.000 \\
\quad iTransformer$^{*}$ & 0.359 $\pm$ 0.001 & 0.347 $\pm$ 0.001 & 0.230 $\pm$ 0.000 & 0.394 $\pm$ 0.003 & 0.305 $\pm$ 0.001 & 0.565 $\pm$ 0.000 \\
\quad ARIMA$^{*}$ & 0.501 $\pm$ 0.003 & 0.620 $\pm$ 0.003 & 0.189 $\pm$ 0.004 & 0.319 $\pm$ 0.006 & 0.275 $\pm$ 0.002 & 0.455 $\pm$ 0.005 \\
\midrule
\multicolumn{7}{l}{\textit{Multimodal Forecasting Models}} \\
\quad UniTime & 0.368 $\pm$ 0.002 & 0.453 $\pm$ 0.002 & 0.156 $\pm$ 0.000 & \second{0.197 $\pm$ 0.004} & 0.392 $\pm$ 0.002 & \second{0.433 $\pm$ 0.001} \\
\midrule
\multicolumn{7}{l}{\textit{Unified Models}} \\
\quad ChatTime & 0.344 $\pm$ 0.000 & 0.332 $\pm$ 0.000 & 0.191 $\pm$ 0.000 & 0.334 $\pm$ 0.000 & 0.315 $\pm$ 0.000 & 0.436 $\pm$ 0.000 \\
\quad TimeOmni-VL & 0.655 $\pm$ 0.028 & 0.678 $\pm$ 0.043 & 0.260 $\pm$ 0.008 & 0.648 $\pm$ 0.023 & 0.586 $\pm$ 0.035 & 0.818 $\pm$ 0.132 \\
\rowcolor{timebraidrow}
\quad TimeBraid-1.2B (Ours) & 0.301 $\pm$ 0.002 & 0.278 $\pm$ 0.003 & 0.138 $\pm$ 0.002 & 0.359 $\pm$ 0.004 & 0.250 $\pm$ 0.002 & 0.472 $\pm$ 0.006 \\
\rowcolor{timebraidrow}
\quad TimeBraid-2.5B (Ours) & \third{0.289 $\pm$ 0.002} & 0.270 $\pm$ 0.002 & 0.142 $\pm$ 0.002 & 0.337 $\pm$ 0.003 & \second{0.239 $\pm$ 0.002} & 0.468 $\pm$ 0.005 \\
\rowcolor{timebraidrow}
\quad TimeBraid-6.7B (Ours) & 0.294 $\pm$ 0.002 & \third{0.267 $\pm$ 0.003} & \third{0.134 $\pm$ 0.003} & 0.359 $\pm$ 0.004 & \third{0.243 $\pm$ 0.003} & 0.437 $\pm$ 0.009 \\
\bottomrule
\end{tabular}%
}
\end{table}

\subsubsection{TemporalBench}

\begin{table}[H]
\centering
\caption{TemporalBench results under the official evaluation setup: (a) accuracy on the 16 multiple-choice tasks, with Average the unweighted macro-average, and (b) forecasting error, not averaged across datasets because the metrics differ.}
\label{tab:temporalbench_all_models}
\scriptsize
\setlength{\tabcolsep}{2.0pt}
\caption*{\textbf{(a) Multiple-choice accuracy.}}
\resizebox{\textwidth}{!}{%
\begin{tabular}{l|cccc|cccc|cccc|cccc|c}
\toprule
\multirow{2}{*}{Model} & \multicolumn{4}{c|}{FreshRetailNet} & \multicolumn{4}{c|}{PSML} & \multicolumn{4}{c|}{Causal Chambers} & \multicolumn{4}{c|}{MIMIC} & \multirow{2}{*}{Average} \\
\cmidrule(lr){2-5}\cmidrule(lr){6-9}\cmidrule(lr){10-13}\cmidrule(lr){14-17}
 & T1 & T2 & T3 & T4 & T1 & T2 & T3 & T4 & T1 & T2 & T3 & T4 & T1 & T2 & T3 & T4 & \\
\midrule
\multicolumn{18}{l}{\textit{Closed-source Reference Models}} \\
GPT-5.4 & 45.45\% & 28.03\% & 41.48\% & 42.42\% & 50.50\% & 31.33\% & 33.20\% & 57.33\% & 18.67\% & 53.33\% & 37.20\% & 45.33\% & 35.11\% & 29.08\% & 35.56\% & 31.91\% & 38.50\% \\
GPT-4.1 & 38.64\% & 31.82\% & 45.45\% & 38.64\% & 48.50\% & 27.33\% & 40.00\% & 53.33\% & 12.00\% & 46.00\% & 48.80\% & 41.33\% & 18.62\% & 29.08\% & 42.68\% & 28.37\% & 36.91\% \\
GPT-4o & 63.07\% & 16.67\% & 28.98\% & 39.39\% & 69.00\% & 23.33\% & 35.20\% & 36.67\% & 10.00\% & 22.67\% & 34.00\% & 42.00\% & 46.81\% & 19.86\% & 0.00\% & 29.08\% & 32.30\% \\
Gemini-2.5-Flash & 59.66\% & 22.73\% & 29.55\% & 38.64\% & 72.50\% & 23.33\% & 22.00\% & 46.67\% & 10.67\% & 45.33\% & 37.20\% & 44.00\% & 42.55\% & 29.08\% & 0.00\% & 29.79\% & 34.61\% \\
Claude-Sonnet-4 & 55.11\% & 29.55\% & 34.66\% & 43.94\% & 50.00\% & 28.67\% & 33.20\% & 40.00\% & 15.33\% & 60.67\% & 37.60\% & 58.67\% & 35.11\% & 21.99\% & 0.00\% & 32.62\% & 36.07\% \\
\midrule
\multicolumn{18}{l}{\textit{Open-source Large Language Models}} \\
DeepSeek-Chat & \best{63.64\%} & 15.91\% & 28.98\% & 34.09\% & \best{57.00\%} & 25.33\% & 32.00\% & \second{56.67\%} & 14.67\% & 28.00\% & 33.60\% & 32.00\% & 28.19\% & 24.11\% & 0.00\% & 23.40\% & 31.10\% \\
Qwen3-8B & \second{36.36\%} & \third{30.30\%} & \second{51.70\%} & \third{40.15\%} & 38.50\% & 29.33\% & \best{48.80\%} & \third{54.67\%} & 17.33\% & 15.33\% & 42.80\% & 15.33\% & 16.49\% & 21.28\% & \second{38.49\%} & 21.99\% & 32.43\% \\
Llama-3.1-8B & 29.55\% & 25.76\% & \third{46.59\%} & 22.73\% & 30.00\% & 23.33\% & \third{38.00\%} & 50.67\% & 15.33\% & 12.67\% & 27.60\% & 13.33\% & 13.30\% & 17.02\% & 30.54\% & 17.73\% & 25.88\% \\
\midrule
\multicolumn{18}{l}{\textit{Time-Series Language Models}} \\
ChatTS & 15.91\% & 18.18\% & 31.82\% & 34.09\% & \third{46.00\%} & 19.33\% & \second{38.40\%} & 54.00\% & \third{22.67\%} & 7.33\% & 35.20\% & 11.33\% & 30.85\% & 11.35\% & 0.00\% & 21.99\% & 24.90\% \\
TimeOmni-1-7B & 26.70\% & 21.97\% & 29.55\% & 26.52\% & 33.50\% & 24.00\% & 26.00\% & 40.00\% & 21.33\% & 23.33\% & 30.40\% & 19.33\% & \second{35.64\%} & 27.66\% & 0.00\% & 24.82\% & 25.67\% \\
\midrule
\multicolumn{18}{l}{\textit{Unified Models}} \\
ChatTime-7B & 31.25\% & 27.27\% & 31.82\% & 23.48\% & 37.50\% & 27.33\% & 30.00\% & 19.33\% & \second{23.33\%} & 20.67\% & 32.80\% & 26.00\% & \best{40.43\%} & 26.24\% & 27.20\% & \best{30.50\%} & 28.45\% \\
TimeOmni-VL & 19.89\% & \second{39.39\%} & 43.75\% & \second{43.18\%} & \second{52.00\%} & 29.33\% & 11.20\% & 44.67\% & 18.00\% & 28.67\% & \third{45.20\%} & 31.33\% & 18.09\% & \best{33.33\%} & 33.47\% & \second{29.79\%} & 32.32\% \\
\rowcolor{timebraidrow}
TimeBraid-1.2B (Ours) & 17.61\% & \best{42.42\%} & 41.48\% & \best{50.00\%} & 14.50\% & \third{32.67\%} & 19.20\% & 18.67\% & 21.33\% & \best{61.33\%} & 36.00\% & \best{61.33\%} & 34.57\% & \third{29.08\%} & \third{35.98\%} & \best{30.50\%} & \third{34.17\%} \\
\rowcolor{timebraidrow}
TimeBraid-2.5B (Ours) & \third{34.66\%} & 25.76\% & \best{52.27\%} & 39.39\% & 14.50\% & \second{36.00\%} & 16.80\% & 52.00\% & 16.00\% & \second{50.67\%} & \best{47.20\%} & \third{54.00\%} & \third{35.11\%} & \second{32.62\%} & 35.15\% & \second{29.79\%} & \second{35.74\%} \\
\rowcolor{timebraidrow}
TimeBraid-6.7B (Ours) & 30.68\% & 27.27\% & 44.32\% & 37.88\% & 18.00\% & \best{41.33\%} & 23.60\% & \best{58.00\%} & \best{24.00\%} & \third{49.33\%} & \second{46.40\%} & \second{59.33\%} & 33.51\% & 17.73\% & \best{46.86\%} & \third{26.95\%} & \best{36.58\%} \\
\bottomrule
\end{tabular}%
}
\end{table}

% Allow the two detailed panels to paginate without shrinking their contents.
\begin{table}[H]
\ContinuedFloat
\centering
\caption[]{TemporalBench results (continued): forecasting error on 852 current-leaderboard-dev cases.}
\scriptsize
\setlength{\tabcolsep}{3.6pt}
\caption*{\textbf{(b) Forecasting error (FreshRetailNet and Causal Chambers: MAE; PSML: sMAPE; MIMIC: OW-sMAPE).}}
\resizebox{\textwidth}{!}{%
\begin{tabular}{l|cc|cc|cc|cc}
\toprule
\multirow{2}{*}{Model} & \multicolumn{2}{c|}{FreshRetailNet} & \multicolumn{2}{c|}{PSML} & \multicolumn{2}{c|}{Causal Chambers} & \multicolumn{2}{c}{MIMIC} \\
\cmidrule(lr){2-3}\cmidrule(lr){4-5}\cmidrule(lr){6-7}\cmidrule(lr){8-9}
 & T2 & T4 & T2 & T4 & T2 & T4 & T2 & T4 \\
\midrule
\multicolumn{9}{l}{\textit{Closed-source Reference Models}} \\
GPT-5.4 & 0.132 & 0.129 & 0.236 & 0.243 & 3.287 & 4.279 & 9.853 & 10.005 \\
GPT-4o & 0.127 & 0.234 & 0.333 & 0.435 & 1.988 & 2.755 & 15.913 & 16.860 \\
Gemini-2.5-Flash & 0.106 & 0.118 & 0.304 & 0.332 & 2.244 & 2.370 & 9.902 & 12.557 \\
Claude-Sonnet-4 & 0.116 & 0.182 & 0.253 & 0.318 & 2.561 & 2.725 & 9.098 & 13.545 \\
\midrule
\multicolumn{9}{l}{\textit{Open-source Large Language Models}} \\
DeepSeek-Chat & \best{0.113} & 0.135 & 0.333 & 0.374 & \best{2.532} & \best{2.126} & 14.981 & 14.789 \\
\midrule
\multicolumn{9}{l}{\textit{Time-Series Foundation Models}} \\
TimesFM$_{2.5}$ & \best{0.113} & \best{0.115} & 0.229 & 0.231 & 4.850 & 4.846 & 9.215 & 9.223 \\
Chronos-2 & \third{0.116} & 0.119 & 0.207 & 0.209 & 4.812 & 4.811 & 9.643 & 9.647 \\
Moirai-2 & \second{0.115} & \second{0.116} & 0.213 & 0.210 & \third{4.766} & \third{4.767} & 9.051 & 9.045 \\
Sundial & 0.122 & 0.121 & 0.205 & \third{0.203} & 5.309 & 5.332 & 9.973 & 9.982 \\
TimeMoE & 0.119 & 0.121 & \best{0.198} & \second{0.198} & 5.331 & 5.347 & 9.600 & 9.632 \\
\midrule
\multicolumn{9}{l}{\textit{Multimodal Forecasting Models}} \\
MIGAS-1.5 & \second{0.115} & \best{0.115} & 0.222 & 0.221 & \second{4.717} & \second{4.753} & 9.553 & 9.602 \\
Aurora & 0.126 & 0.125 & \third{0.204} & 0.204 & 6.918 & 6.917 & 10.469 & 10.475 \\
\midrule
\multicolumn{9}{l}{\textit{Unified Models}} \\
ChatTime-7B & 0.118 & 0.143 & 0.225 & 0.223 & 5.397 & 5.220 & 10.256 & 10.379 \\
TimeOmni-VL & 1.537 & 1.587 & 0.726 & 0.787 & 41.261 & 54.678 & 77.746 & 85.100 \\
\rowcolor{timebraidrow}
TimeBraid-1.2B (Ours) & 0.117 & \third{0.118} & 0.205 & 0.204 & 5.317 & 5.316 & \third{8.946} & \second{8.996} \\
\rowcolor{timebraidrow}
TimeBraid-2.5B (Ours) & 0.120 & 0.119 & \second{0.201} & \third{0.203} & 5.288 & 5.307 & \second{8.945} & \third{9.002} \\
\rowcolor{timebraidrow}
TimeBraid-6.7B (Ours) & \third{0.116} & \second{0.116} & \best{0.198} & \best{0.197} & 5.304 & 5.316 & \best{8.931} & \best{8.972} \\
\bottomrule
\end{tabular}%
}
\vspace{2pt}
\end{table}

\subsubsection{TimeMMD}

% TimeMMD text re-eval (official 8/36/96 lookbacks, paired text): same
% replacement set as tab:timemmd_text. TimeBraid / MIGAS / TSFM remain at
% lookback 128; ChatTime / Aurora already use official lookbacks.
% best/second/third and Avg Rank are over the twenty listed models;
% ties share a mark. Avg Rank uses average ranks on ties.
\begin{table}[H]
\centering
\caption{TimeMMD forecasting with paired text, reporting per-domain MSE and MAE macro-averaged over four horizons. Every multimodal forecasting model receives each window's paired text and runs at the official 8/36/96 lookbacks, as do ChatTime and Aurora; TimeBraid, the time-series foundation models, and MIGAS-1.5 use a fixed 128-step history. Avg Rank averages each model's rank across the nine domains; lower is better.}
\label{tab:timemmd_text_official}
\scriptsize
\setlength{\tabcolsep}{0.8pt}
\resizebox{\textwidth}{!}{%
\begin{tabular}{c|>{\columncolor{timebraidrow}}c>{\columncolor{timebraidrow}}c|>{\columncolor{timebraidrow}}c>{\columncolor{timebraidrow}}c|>{\columncolor{timebraidrow}}c>{\columncolor{timebraidrow}}c|cc|cc|cc|cc|cc|cc|cc|cc|cc|cc|cc|cc|cc|cc|cc|cc|cc}
\toprule
\multicolumn{1}{c|}{} & \multicolumn{8}{c|}{\textit{Unified Models}} & \multicolumn{22}{c|}{\textit{Multimodal Forecasting Models}} & \multicolumn{10}{c}{\textit{Time-Series Foundation Models}} \\
Models & \multicolumn{2}{>{\columncolor{timebraidrow}}c|}{TimeBraid-1.2B} & \multicolumn{2}{>{\columncolor{timebraidrow}}c|}{TimeBraid-2.5B} & \multicolumn{2}{>{\columncolor{timebraidrow}}c|}{TimeBraid-6.7B} & \multicolumn{2}{c|}{ChatTime-7B} & \multicolumn{2}{c|}{Time-LLM} & \multicolumn{2}{c|}{GPT4MTS} & \multicolumn{2}{c|}{TaTS} & \multicolumn{2}{c|}{Time-VLM} & \multicolumn{2}{c|}{PatchTST$^{*}$} & \multicolumn{2}{c|}{iTransformer$^{*}$} & \multicolumn{2}{c|}{RaFT$^{*}$} & \multicolumn{2}{c|}{DLinear} & \multicolumn{2}{c|}{Reformer} & \multicolumn{2}{c|}{MIGAS-1.5} & \multicolumn{2}{c|}{Aurora} & \multicolumn{2}{c|}{TimesFM$_{2.5}$} & \multicolumn{2}{c|}{Chronos-2} & \multicolumn{2}{c|}{Moirai$_{\mathrm{Small}}$} & \multicolumn{2}{c|}{Sundial$_{\mathrm{Base}}$} & \multicolumn{2}{c}{Time-MoE-200M}  \\
\midrule
Metric & MSE & MAE & MSE & MAE & MSE & MAE & MSE & MAE & MSE & MAE & MSE & MAE & MSE & MAE & MSE & MAE & MSE & MAE & MSE & MAE & MSE & MAE & MSE & MAE & MSE & MAE & MSE & MAE & MSE & MAE & MSE & MAE & MSE & MAE & MSE & MAE & MSE & MAE & MSE & MAE \\
\midrule
Agriculture & 0.116 & 0.209 & 0.137 & 0.232 & 0.144 & 0.235 & 0.130 & 0.240 & 0.098 & 0.204 & 0.093 & \second{0.189} & \third{0.091} & 0.196 & 0.093 & \third{0.193} & 0.093 & 0.197 & 0.164 & 0.287 & 0.190 & 0.322 & 0.249 & 0.374 & 0.497 & 0.556 & \second{0.086} & \best{0.185} & 0.109 & 0.208 & 0.127 & 0.223 & \best{0.085} & \best{0.185} & 0.102 & 0.197 & 0.102 & 0.195 & 0.107 & 0.200  \\
Climate & 0.899 & \third{0.746} & \third{0.871} & \second{0.729} & 0.938 & 0.760 & 2.279 & 1.222 & 1.320 & 0.938 & 1.262 & 0.919 & 1.307 & 0.938 & 1.282 & 0.930 & 1.267 & 0.923 & 1.247 & 0.910 & 1.848 & 1.092 & 1.244 & 0.907 & 1.131 & 0.850 & 0.983 & 0.791 & 1.428 & 0.972 & \best{0.859} & \second{0.729} & 0.988 & 0.794 & 0.949 & 0.770 & \second{0.865} & \best{0.714} & 0.935 & 0.748  \\
Economy & 0.025 & 0.125 & 0.025 & 0.125 & 0.018 & 0.104 & 0.071 & 0.219 & 0.031 & 0.143 & 0.027 & 0.134 & \second{0.016} & \third{0.100} & 0.020 & 0.113 & 0.019 & 0.109 & 0.108 & 0.297 & 0.177 & 0.351 & 0.192 & 0.382 & 1.118 & 0.977 & \best{0.015} & \second{0.098} & 0.040 & 0.166 & \third{0.017} & 0.104 & \best{0.015} & \best{0.096} & 0.020 & 0.109 & 0.022 & 0.119 & 0.021 & 0.117  \\
Energy & 0.256 & 0.348 & \second{0.235} & \second{0.335} & \best{0.231} & \best{0.332} & 0.365 & 0.460 & 0.279 & 0.384 & 0.266 & 0.385 & 0.277 & 0.387 & 0.253 & 0.367 & 0.269 & 0.381 & 0.294 & 0.407 & 0.245 & 0.350 & 0.341 & 0.434 & 0.555 & 0.571 & 0.242 & 0.347 & 0.367 & 0.448 & \third{0.239} & \third{0.338} & 0.262 & 0.357 & 0.277 & 0.368 & 0.266 & 0.353 & 0.277 & 0.357  \\
Environment & 0.516 & \third{0.509} & 0.507 & \second{0.498} & 0.507 & 0.513 & 1.099 & 0.757 & 0.519 & 0.516 & 0.518 & 0.517 & \best{0.431} & \best{0.489} & \third{0.494} & 0.521 & \second{0.492} & 0.510 & 0.519 & 0.521 & 0.548 & 0.553 & 0.579 & 0.625 & 0.523 & 0.575 & 0.671 & 0.591 & 0.597 & 0.575 & 0.569 & 0.521 & 0.566 & 0.527 & 0.581 & 0.543 & 0.561 & 0.550 & 0.579 & 0.562  \\
Health(US) & 1.002 & 0.640 & \third{0.956} & \third{0.619} & \second{0.925} & \second{0.608} & 2.516 & 1.037 & 1.488 & 0.827 & 1.519 & 0.826 & 1.665 & 0.810 & 1.482 & 0.820 & 1.556 & 0.799 & 1.726 & 0.808 & 1.534 & 0.873 & 1.714 & 0.870 & 1.567 & 0.824 & 1.302 & 0.728 & 2.000 & 1.022 & \best{0.665} & \best{0.484} & 1.217 & 0.698 & 1.280 & 0.763 & 1.240 & 0.749 & 1.229 & 0.731  \\
Security & \third{108.957} & \second{4.695} & \second{108.864} & \best{4.691} & \best{108.552} & \third{4.715} & 144.336 & 5.700 & 114.884 & 5.486 & 111.210 & 5.288 & 120.945 & 5.656 & 112.981 & 5.320 & 111.667 & 5.265 & 119.281 & 5.588 & 160.934 & 6.405 & 109.860 & 5.264 & 119.267 & 5.682 & 109.911 & 4.784 & 120.737 & 6.028 & 114.802 & 5.150 & 110.309 & 4.819 & 109.829 & 4.811 & 111.686 & 5.010 & 110.392 & 4.927  \\
SocialGood & \third{0.808} & \best{0.351} & 0.839 & 0.368 & \best{0.799} & \second{0.352} & 1.227 & 0.576 & 1.023 & 0.468 & 1.014 & 0.451 & 1.148 & 0.464 & 1.068 & 0.498 & 1.110 & 0.484 & 1.187 & 0.468 & 0.994 & 0.477 & 0.979 & 0.500 & 1.008 & 0.638 & 1.110 & 0.396 & 1.155 & 0.576 & 0.845 & 0.365 & 1.110 & 0.396 & 0.859 & 0.364 & 0.856 & \third{0.361} & \second{0.801} & 0.364  \\
Traffic & \second{0.148} & 0.213 & \best{0.145} & \third{0.206} & 0.153 & 0.211 & 0.505 & 0.555 & 0.223 & 0.299 & 0.249 & 0.321 & 0.227 & 0.278 & 0.229 & 0.305 & 0.216 & 0.267 & 0.216 & 0.286 & 0.460 & 0.523 & 0.316 & 0.451 & 0.319 & 0.471 & 0.200 & 0.216 & 0.322 & 0.426 & \third{0.152} & \best{0.179} & 0.201 & 0.217 & 0.163 & \second{0.191} & 0.160 & 0.211 & 0.167 & 0.225  \\
\midrule
\multicolumn{1}{c|}{Avg Rank} & \third{5.83} & \third{5.39} & \second{5.11} & \second{5.11} & \best{4.50} & \best{4.78} & 18.56 & 18.61 & 12.28 & 12.78 & 10.50 & 11.33 & 11.11 & 10.17 & 9.50 & 11.33 & 9.72 & 10.00 & 14.89 & 13.50 & 14.67 & 16.22 & 14.17 & 16.22 & 15.22 & 17.28 & 8.06 & 6.56 & 17.00 & 16.78 & 6.33 & 5.44 & 7.61 & 6.50 & 8.67 & 7.17 & 7.89 & 6.61 & 8.39 & 8.22  \\
\bottomrule
\end{tabular}%
}
\end{table}

\subsubsection{CGTSF}

CGTSF is the context-guided forecasting benchmark of ChatTime \citep{wang2025chattime}, with three real-world datasets paired with aligned text: MSPG (Melbourne solar power generation), LEU (London electricity usage), and PTF (Paris traffic flow).

\begin{table}[H]
\centering
\caption{CGTSF context-guided forecasting \citep{wang2025chattime} on MSPG (Melbourne solar power generation), LEU (London electricity usage), and PTF (Paris traffic flow). Errors are standardized using training-split statistics; per-dataset entries average four history-length settings, and Macro MSE and Macro MAE weight the twelve settings equally. We replicate the official evaluation setting and evaluate every baseline ourselves.}
\label{tab:cgtsf}
\small
\setlength{\tabcolsep}{3.5pt}
\begin{tabular}{lccccc}
\toprule
\multirow{2}{*}{Model} & \multicolumn{3}{c}{Per-dataset (MSE\,/\,MAE\,$\downarrow$)} & \multicolumn{2}{c}{Macro} \\
\cmidrule(lr){2-4}\cmidrule(lr){5-6}
 & MSPG & LEU & PTF & MSE\,$\downarrow$ & MAE\,$\downarrow$ \\
\midrule
\multicolumn{6}{l}{\textit{Time-Series Foundation Models}} \\
Chronos-2 & 0.649\,/\,0.388 & 0.715\,/\,\best{0.414} & 0.230\,/\,0.285 & 0.531 & \third{0.363} \\
TimesFM-2.5 & 0.585\,/\,0.384 & 0.777\,/\,0.442 & 0.243\,/\,0.306 & 0.535 & 0.377 \\
Sundial & 0.763\,/\,0.507 & 0.683\,/\,0.453 & 0.242\,/\,0.306 & 0.563 & 0.422 \\
Moirai-2 & 1.000\,/\,0.590 & 0.681\,/\,\second{0.422} & 0.284\,/\,0.332 & 0.655 & 0.448 \\
Time-MoE-200M & 1.138\,/\,0.713 & 0.698\,/\,0.450 & 0.239\,/\,0.302 & 0.692 & 0.488 \\
\midrule
\multicolumn{6}{l}{\textit{Multimodal Forecasting Models}} \\
MIGAS-1.5 & 0.757\,/\,0.476 & 0.887\,/\,0.550 & 0.386\,/\,0.444 & 0.677 & 0.490 \\
Aurora & 0.789\,/\,0.562 & 0.757\,/\,0.503 & 0.336\,/\,0.394 & 0.627 & 0.486 \\
Time-LLM & 0.583\,/\,0.476 & 0.699\,/\,0.480 & 0.163\,/\,\third{0.273} & 0.482 & 0.409 \\
GPT4MTS & 0.567\,/\,0.438 & 0.694\,/\,0.481 & 0.180\,/\,\third{0.273} & 0.481 & 0.397 \\
Time-VLM & 0.818\,/\,0.644 & 0.725\,/\,0.498 & 0.175\,/\,0.278 & 0.573 & 0.473 \\
PatchTST$^{*}$ & 0.602\,/\,0.479 & 0.706\,/\,0.489 & 0.179\,/\,0.277 & 0.496 & 0.415 \\
iTransformer$^{*}$ & 0.737\,/\,0.507 & 0.698\,/\,0.490 & 0.193\,/\,0.285 & 0.543 & 0.428 \\
RaFT$^{*}$ & 1.398\,/\,0.608 & 0.701\,/\,0.503 & 0.238\,/\,0.322 & 0.779 & 0.478 \\
DLinear & 0.619\,/\,0.470 & 0.687\,/\,0.487 & 0.197\,/\,0.300 & 0.501 & 0.419 \\
Reformer & 0.859\,/\,0.584 & 0.736\,/\,0.514 & 0.269\,/\,0.355 & 0.621 & 0.484 \\
\midrule
\multicolumn{6}{l}{\textit{Unified Models}} \\
ChatTime & 2.727\,/\,0.945 & 1.015\,/\,0.507 & 0.359\,/\,0.373 & 1.367 & 0.609 \\
\rowcolor{timebraidrow}
TimeBraid-1.2B (Ours) & \third{0.472}\,/\,\third{0.371} & \second{0.642}\,/\,0.425 & \second{0.136}\,/\,\second{0.229} & \third{0.417} & \second{0.342} \\
\rowcolor{timebraidrow}
TimeBraid-2.5B (Ours) & \second{0.467}\,/\,\second{0.368} & \third{0.644}\,/\,\third{0.423} & \best{0.133}\,/\,\best{0.225} & \second{0.415} & \best{0.338} \\
\rowcolor{timebraidrow}
TimeBraid-6.7B (Ours) & \best{0.459}\,/\,\best{0.362} & \best{0.641}\,/\,0.424 & \third{0.138}\,/\,\second{0.229} & \best{0.413} & \best{0.338} \\
\bottomrule
\end{tabular}
\end{table}

We report history-$z$-normalized errors: per-dataset MSE/MAE average each dataset's four history-length settings, and Macro MSE and Macro MAE equally average the twelve dataset$\times$history-length settings.

\subsubsection{CAF (Context-Aided Forecasting)}

CAF evaluates whether models use scenario information that complements the observed history while forecasting a predictive distribution. We report normalized CRPS with standard errors on all 904 correct-context test cases and their HARD and EASY strata; the evaluation set is held out from our mixture as summarized in Appendix~\ref{app:data_overlap}.

\subsubsection{Ctrl-F}

Ctrl-F tests whether a forecast follows a natural-language future condition when three distinct continuations share the same observed history. The set contains 50 domain-balanced groups across commerce, compute, epidemic, health, and traffic, whose histories are replayed from held-out GIFT-Eval test anchors. Each group contributes three original-condition windows, three paraphrased-condition windows, and one shared blank-text control, yielding 350 instances in total. We report history-normalized forecasting error on the 150 original-condition windows and Top-1 sibling retrieval against the three candidate futures (chance $=33.33\%$).

\begin{table}[H]
\centering
\begin{minipage}[t]{0.49\textwidth}
\centering
\caption{CAF normalized CRPS on all 904 correct-context examples and the hard/easy strata (lower is better; mean $\pm$ s.e.).}
\label{tab:caf}
\small
\setlength{\tabcolsep}{2pt}
\resizebox{\linewidth}{!}{%
\begin{tabular}{lccc}
\toprule
\multirow{2}{*}{Model} & \multicolumn{3}{c}{Normalized CRPS\,$\downarrow$} \\
\cmidrule(lr){2-4}
 & ALL & HARD & EASY \\
\midrule
\multicolumn{4}{l}{\textit{Closed-source Reference Models}} \\
GPT-5.4 & 0.233\,{\scriptsize$\pm$0.011} & 0.225\,{\scriptsize$\pm$0.012} & 0.243\,{\scriptsize$\pm$0.019} \\
GPT-4o & 0.234\,{\scriptsize$\pm$0.011} & 0.228\,{\scriptsize$\pm$0.012} & 0.241\,{\scriptsize$\pm$0.018} \\
Gemini-2.5-Flash & 0.247\,{\scriptsize$\pm$0.012} & 0.247\,{\scriptsize$\pm$0.016} & 0.246\,{\scriptsize$\pm$0.018} \\
\midrule
\multicolumn{4}{l}{\textit{Open-source Large Language Models}} \\
DeepSeek-Chat & 0.267\,{\scriptsize$\pm$0.013} & 0.254\,{\scriptsize$\pm$0.015} & 0.283\,{\scriptsize$\pm$0.021} \\
\midrule
\multicolumn{4}{l}{\textit{Time-Series Foundation Models}} \\
Moirai-2 & 0.275\,{\scriptsize$\pm$0.014} & 0.299\,{\scriptsize$\pm$0.019} & 0.248\,{\scriptsize$\pm$0.020} \\
Chronos-2 & 0.277\,{\scriptsize$\pm$0.015} & 0.298\,{\scriptsize$\pm$0.019} & 0.252\,{\scriptsize$\pm$0.022} \\
TimesFM$_{2.5}$ & 0.281\,{\scriptsize$\pm$0.016} & 0.293\,{\scriptsize$\pm$0.020} & 0.267\,{\scriptsize$\pm$0.025} \\
Sundial & 0.331\,{\scriptsize$\pm$0.017} & 0.340\,{\scriptsize$\pm$0.021} & 0.321\,{\scriptsize$\pm$0.028} \\
Time-MoE-200M & 0.414\,{\scriptsize$\pm$0.023} & 0.388\,{\scriptsize$\pm$0.023} & 0.445\,{\scriptsize$\pm$0.042} \\
\midrule
\multicolumn{4}{l}{\textit{Multimodal Forecasting Models}} \\
MIGAS-1.5 & 0.346\,{\scriptsize$\pm$0.017} & 0.379\,{\scriptsize$\pm$0.023} & 0.307\,{\scriptsize$\pm$0.025} \\
Aurora & 0.460\,{\scriptsize$\pm$0.021} & 0.428\,{\scriptsize$\pm$0.021} & 0.497\,{\scriptsize$\pm$0.039} \\
DoubleCast & \second{0.231\,{\scriptsize$\pm$0.001}} & \third{0.251\,{\scriptsize$\pm$0.001}} & \best{0.204\,{\scriptsize$\pm$0.002}} \\
TimeLLM & 0.501\,{\scriptsize$\pm$0.001} & 0.428\,{\scriptsize$\pm$0.001} & 0.587\,{\scriptsize$\pm$0.001} \\
\midrule
\multicolumn{4}{l}{\textit{Unified Models}} \\
ChatTime-7B & 0.345\,{\scriptsize$\pm$0.015} & 0.375\,{\scriptsize$\pm$0.020} & 0.310\,{\scriptsize$\pm$0.022} \\
TimeOmni-VL & 0.489\,{\scriptsize$\pm$0.025} & 0.425\,{\scriptsize$\pm$0.027} & 0.565\,{\scriptsize$\pm$0.043} \\
\rowcolor{timebraidrow}
TimeBraid-1.2B (Ours) & \third{0.235\,{\scriptsize$\pm$0.012}} & \second{0.243\,{\scriptsize$\pm$0.016}} & \third{0.225\,{\scriptsize$\pm$0.018}} \\
\rowcolor{timebraidrow}
TimeBraid-2.5B (Ours) & \third{0.235\,{\scriptsize$\pm$0.012}} & \second{0.243\,{\scriptsize$\pm$0.014}} & 0.226\,{\scriptsize$\pm$0.019} \\
\rowcolor{timebraidrow}
TimeBraid-6.7B (Ours) & \best{0.225\,{\scriptsize$\pm$0.012}} & \best{0.231\,{\scriptsize$\pm$0.015}} & \second{0.219\,{\scriptsize$\pm$0.018}} \\
\bottomrule
\end{tabular}
}
\end{minipage}\hfill
\begin{minipage}[t]{0.49\textwidth}
\centering
\caption{Ctrl-F on 150 correct-arm windows, reporting history-normalized MSE and MAE with correct-sibling Top-1 accuracy (chance $=33.33\%$).}
\label{tab:ctrlf}
\small
\setlength{\tabcolsep}{3pt}
\resizebox{\linewidth}{!}{%
\begin{tabular}{lccc}
\toprule
Model & MSE\,$\downarrow$ & MAE\,$\downarrow$ & Top-1 (\%)\,$\uparrow$ \\
\midrule
\multicolumn{4}{l}{\textit{Closed-source Reference Models}} \\
GPT-5.4 & 5.188 & 1.563 & 51.33 \\
GPT-4o & 5.447 & 1.591 & 60.00 \\
Gemini-2.5-Flash & 6.624 & 1.771 & 56.00 \\
\midrule
\multicolumn{4}{l}{\textit{Open-source Large Language Models}} \\
DeepSeek-Chat & \second{6.405} & \second{1.692} & \best{59.33} \\
\midrule
\multicolumn{4}{l}{\textit{Time-Series Foundation Models}} \\
Moirai-2 & 8.425 & 1.987 & 33.33 \\
Chronos-2 & 8.597 & 2.004 & 33.33 \\
TimesFM-2.5 & 8.784 & 2.025 & 33.33 \\
Sundial & 8.829 & 2.074 & 33.33 \\
Time-MoE-200M & 9.077 & 2.114 & 33.33 \\
\midrule
\multicolumn{4}{l}{\textit{Multimodal Forecasting Models}} \\
MIGAS-1.5 & 8.604 & 2.009 & 33.33 \\
Aurora & 9.091 & 2.141 & 33.33 \\
\midrule
\multicolumn{4}{l}{\textit{Unified Models}} \\
ChatTime & 9.165 & 2.140 & 33.33 \\
TimeOmni-VL & 10.387 & 2.367 & 32.00 \\
\rowcolor{timebraidrow}
TimeBraid-1.2B (Ours) & \third{6.826} & \third{1.768} & \third{40.67} \\
\rowcolor{timebraidrow}
TimeBraid-2.5B (Ours) & 6.972 & 1.781 & \third{40.67} \\
\rowcolor{timebraidrow}
TimeBraid-6.7B (Ours) & \best{6.232} & \best{1.689} & \second{43.33} \\
\bottomrule
\end{tabular}
}
\end{minipage}
\end{table}
\subsubsection{Unimodal Forecasting}

\begin{figure}[H]
\centering
\includegraphics[width=0.94\textwidth]{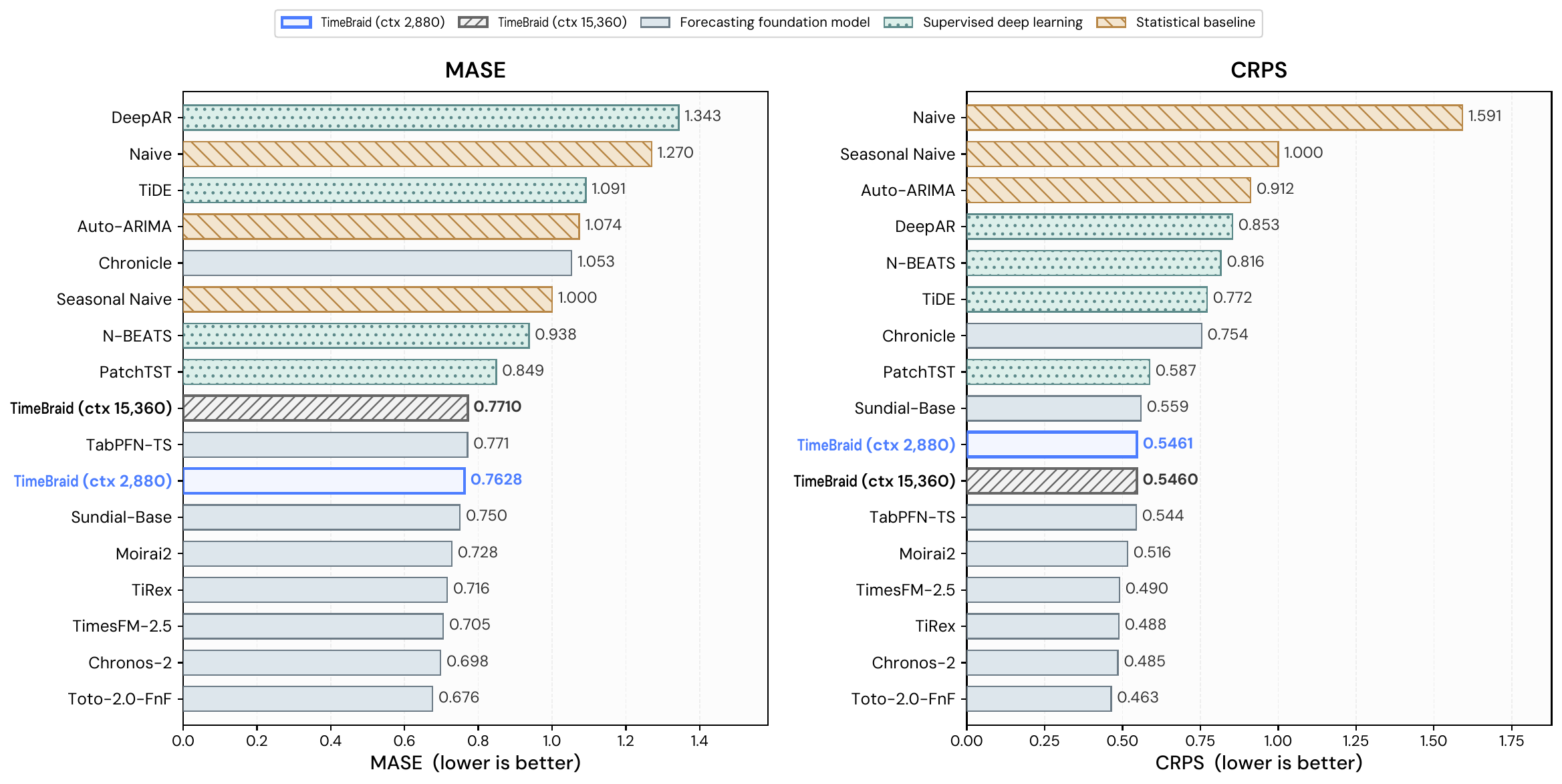}
\caption{Aggregate GIFT-Eval forecasting performance across models and context lengths. Lower is better.}
\label{fig:gift_eval}
\end{figure}

\begin{table}[H]
\centering
\caption{Per-horizon long-term forecasting on ETT and Weather, reporting MSE and MAE at horizons $\{96,192,336,720\}$ and their average. TimeBraid-2.5B uses a 2{,}880-step lookback; the baselines retain their official input lengths. Avg Rank averages each model's rank across the twenty dataset--horizon settings.}
\label{tab:unimodal_forecasting}
\scriptsize
\setlength{\tabcolsep}{1.6pt}
\resizebox{\textwidth}{!}{%
\begin{tabular}{c|c|>{\columncolor{timebraidrow}}c>{\columncolor{timebraidrow}}c|cc|cc|cc|cc|cc|cc|cc|cc|cc}
\toprule
\multicolumn{2}{c|}{} & \multicolumn{2}{>{\columncolor{timebraidrow}}c|}{\textit{TimeBraid}} & \multicolumn{10}{c|}{\textit{Time-Series Foundation Models}} & \multicolumn{8}{c}{\textit{Traditional Time-Series Models}} \\
\multicolumn{2}{c|}{Models} & \multicolumn{2}{>{\columncolor{timebraidrow}}c|}{TimeBraid-2.5B} & \multicolumn{2}{c|}{Sundial$_{\mathrm{Large}}$} & \multicolumn{2}{c|}{Time-MoE$_{\mathrm{Ultra}}$} & \multicolumn{2}{c|}{Moirai$_{\mathrm{Large}}$} & \multicolumn{2}{c|}{TimesFM$_{2.5}$} & \multicolumn{2}{c|}{Chronos-2} & \multicolumn{2}{c|}{iTransformer} & \multicolumn{2}{c|}{TimeMixer} & \multicolumn{2}{c|}{PatchTST} & \multicolumn{2}{c}{DLinear} \\
\multicolumn{2}{c|}{} & \multicolumn{2}{>{\columncolor{timebraidrow}}c|}{(Ours)} & \multicolumn{2}{c|}{(Zero-shot)} & \multicolumn{2}{c|}{(Zero-shot)} & \multicolumn{2}{c|}{(Zero-shot)} & \multicolumn{2}{c|}{(Zero-shot)} & \multicolumn{2}{c|}{(Zero-shot)} & \multicolumn{2}{c|}{(Full-shot)} & \multicolumn{2}{c|}{(Full-shot)} & \multicolumn{2}{c|}{(Full-shot)} & \multicolumn{2}{c}{(Full-shot)} \\
\midrule
\multicolumn{2}{c|}{Metric} & MSE & MAE & MSE & MAE & MSE & MAE & MSE & MAE & MSE & MAE & MSE & MAE & MSE & MAE & MSE & MAE & MSE & MAE & MSE & MAE \\
\midrule
\multirow{5}{*}{ETTm1} & 96 & 0.310 & 0.336 & \best{0.273} & \third{0.329} & \second{0.281} & 0.341 & 0.380 & 0.361 & 0.307 & \second{0.326} & \third{0.301} & \best{0.312} & 0.334 & 0.368 & 0.320 & 0.357 & 0.329 & 0.367 & 0.345 & 0.372 \\
 & 192 & 0.356 & 0.364 & \second{0.312} & \second{0.357} & \best{0.305} & \third{0.358} & 0.412 & 0.383 & 0.358 & \third{0.358} & \third{0.352} & \best{0.343} & 0.377 & 0.391 & 0.361 & 0.381 & 0.367 & 0.385 & 0.380 & 0.389 \\
 & 336 & \third{0.388} & 0.385 & \best{0.343} & \second{0.378} & \second{0.369} & 0.395 & 0.436 & 0.400 & 0.389 & \third{0.381} & \third{0.388} & \best{0.367} & 0.426 & 0.420 & 0.390 & 0.404 & 0.399 & 0.410 & 0.413 & 0.413 \\
 & 720 & \second{0.439} & 0.415 & \best{0.397} & \second{0.413} & 0.469 & 0.472 & 0.462 & 0.420 & \third{0.444} & \third{0.414} & 0.451 & \best{0.403} & 0.491 & 0.459 & 0.454 & 0.441 & 0.454 & 0.439 & 0.474 & 0.453 \\
\cmidrule{2-22}
 & Avg & \third{0.373} & 0.375 & \best{0.331} & \second{0.369} & \second{0.356} & 0.391 & 0.422 & 0.391 & 0.375 & \third{0.370} & \third{0.373} & \best{0.356} & 0.407 & 0.409 & 0.381 & 0.395 & 0.387 & 0.400 & 0.403 & 0.406 \\
\midrule
\multirow{5}{*}{ETTm2} & 96 & \third{0.171} & \third{0.250} & 0.172 & 0.255 & 0.198 & 0.288 & 0.211 & 0.274 & \second{0.170} & \second{0.239} & \best{0.161} & \best{0.225} & 0.180 & 0.264 & 0.175 & 0.258 & 0.175 & 0.259 & 0.193 & 0.292 \\
 & 192 & \best{0.225} & \third{0.290} & \third{0.227} & 0.296 & 0.235 & 0.312 & 0.281 & 0.318 & 0.234 & \second{0.283} & \second{0.226} & \best{0.271} & 0.250 & 0.309 & 0.237 & 0.299 & 0.241 & 0.302 & 0.284 & 0.362 \\
 & 336 & \best{0.273} & \third{0.324} & \second{0.275} & 0.331 & 0.293 & 0.348 & 0.341 & 0.355 & 0.293 & \second{0.321} & \third{0.278} & \best{0.307} & 0.311 & 0.348 & 0.298 & 0.340 & 0.305 & 0.343 & 0.369 & 0.427 \\
 & 720 & \third{0.356} & \second{0.378} & \best{0.343} & \second{0.378} & 0.427 & 0.428 & 0.485 & 0.428 & 0.389 & \third{0.379} & \second{0.352} & \best{0.359} & 0.412 & 0.407 & 0.391 & 0.396 & 0.402 & 0.400 & 0.554 & 0.522 \\
\cmidrule{2-22}
 & Avg & \second{0.256} & \third{0.311} & \best{0.254} & 0.315 & 0.288 & 0.344 & 0.329 & 0.343 & \third{0.271} & \second{0.305} & \best{0.254} & \best{0.291} & 0.288 & 0.332 & 0.275 & 0.323 & 0.280 & 0.326 & 0.350 & 0.400 \\
\midrule
\multirow{5}{*}{ETTh1} & 96 & \third{0.364} & 0.384 & \best{0.346} & 0.383 & \second{0.349} & \second{0.379} & 0.381 & 0.388 & 0.366 & \third{0.381} & 0.369 & \best{0.372} & 0.386 & 0.405 & 0.375 & 0.400 & 0.414 & 0.419 & 0.386 & 0.400 \\
 & 192 & \third{0.396} & \third{0.404} & \best{0.386} & 0.410 & \second{0.395} & 0.413 & 0.434 & 0.415 & 0.401 & \second{0.402} & 0.417 & \best{0.400} & 0.441 & 0.436 & 0.436 & 0.429 & 0.460 & 0.445 & 0.437 & 0.432 \\
 & 336 & \best{0.407} & \second{0.413} & \second{0.410} & 0.426 & 0.447 & 0.453 & 0.485 & 0.445 & \third{0.415} & \best{0.412} & 0.447 & \third{0.417} & 0.487 & 0.458 & 0.484 & 0.458 & 0.501 & 0.466 & 0.481 & 0.459 \\
 & 720 & \best{0.404} & \second{0.430} & \second{0.438} & 0.459 & 0.457 & 0.462 & 0.611 & 0.510 & \best{0.404} & \best{0.423} & \third{0.445} & \third{0.431} & 0.503 & 0.491 & 0.498 & 0.482 & 0.500 & 0.488 & 0.519 & 0.516 \\
\cmidrule{2-22}
 & Avg & \best{0.393} & \second{0.407} & \second{0.395} & \third{0.420} & 0.412 & 0.426 & 0.480 & 0.439 & \third{0.396} & \best{0.405} & 0.420 & \best{0.405} & 0.454 & 0.447 & 0.448 & 0.442 & 0.468 & 0.454 & 0.455 & 0.451 \\
\midrule
\multirow{5}{*}{ETTh2} & 96 & 0.284 & 0.334 & \best{0.269} & \third{0.330} & 0.292 & 0.352 & 0.296 & \third{0.330} & \second{0.275} & \second{0.318} & \third{0.280} & \best{0.316} & 0.297 & 0.349 & 0.289 & 0.341 & 0.302 & 0.348 & 0.333 & 0.387 \\
 & 192 & \second{0.341} & 0.373 & \best{0.325} & 0.373 & 0.347 & 0.379 & 0.361 & \third{0.371} & \third{0.344} & \second{0.365} & 0.346 & \best{0.361} & 0.380 & 0.400 & 0.372 & 0.392 & 0.388 & 0.400 & 0.477 & 0.476 \\
 & 336 & \second{0.360} & \third{0.391} & \best{0.354} & 0.400 & 0.406 & 0.419 & 0.390 & \second{0.390} & 0.374 & \second{0.390} & \third{0.370} & \best{0.386} & 0.428 & 0.432 & 0.386 & 0.414 & 0.426 & 0.433 & 0.594 & 0.541 \\
 & 720 & \best{0.368} & \third{0.410} & 0.389 & 0.443 & 0.439 & 0.447 & 0.423 & 0.418 & \third{0.377} & \second{0.402} & \second{0.370} & \best{0.400} & 0.427 & 0.445 & 0.412 & 0.434 & 0.431 & 0.446 & 0.831 & 0.657 \\
\cmidrule{2-22}
 & Avg & \second{0.338} & \third{0.377} & \best{0.334} & 0.387 & 0.371 & 0.399 & 0.367 & \third{0.377} & 0.343 & \second{0.369} & \third{0.342} & \best{0.366} & 0.383 & 0.406 & 0.364 & 0.395 & 0.386 & 0.406 & 0.558 & 0.515 \\
\midrule
\multirow{5}{*}{Weather} & 96 & \second{0.145} & \third{0.189} & \third{0.157} & 0.208 & \third{0.157} & 0.211 & 0.199 & 0.211 & \second{0.145} & \second{0.179} & \best{0.139} & \best{0.172} & 0.174 & 0.214 & 0.163 & 0.209 & 0.177 & 0.218 & 0.196 & 0.255 \\
 & 192 & \second{0.193} & \third{0.236} & 0.207 & 0.256 & 0.208 & 0.256 & 0.246 & 0.251 & \third{0.195} & \second{0.228} & \best{0.186} & \best{0.219} & 0.221 & 0.254 & 0.208 & 0.250 & 0.225 & 0.259 & 0.237 & 0.296 \\
 & 336 & \second{0.245} & \third{0.278} & 0.259 & 0.295 & 0.255 & 0.290 & 0.274 & 0.291 & 0.260 & \second{0.275} & \best{0.240} & \best{0.260} & 0.278 & 0.296 & \third{0.251} & 0.287 & 0.278 & 0.297 & 0.283 & 0.335 \\
 & 720 & \third{0.329} & \third{0.337} & \second{0.327} & 0.342 & 0.405 & 0.397 & 0.337 & 0.340 & 0.359 & \second{0.336} & \best{0.311} & \best{0.311} & 0.358 & 0.349 & 0.339 & 0.341 & 0.354 & 0.348 & 0.345 & 0.381 \\
\cmidrule{2-22}
 & Avg & \second{0.228} & \third{0.260} & \third{0.238} & 0.275 & 0.256 & 0.288 & 0.264 & 0.273 & 0.240 & \second{0.255} & \best{0.219} & \best{0.241} & 0.257 & 0.278 & 0.240 & 0.271 & 0.258 & 0.280 & 0.265 & 0.316 \\
\midrule
\multicolumn{2}{c|}{Avg Rank} & \second{2.48} & \third{3.45} & \best{2.23} & 4.15 & 5.20 & 6.72 & 8.05 & 6.00 & 3.83 & \second{2.25} & \third{2.80} & \best{1.20} & 8.10 & 8.07 & 5.67 & 5.80 & 7.72 & 7.88 & 8.93 & 9.47 \\
\bottomrule
\end{tabular}%
}
\end{table}

\FloatBarrier

% Start the retention discussion with its heading, paragraph, and wrapped table.
\clearpage
\subsection{Language Ability Retention}
\label{app:language_retention}

\begin{wraptable}[15]{r}{0.54\textwidth}
\vspace{-0.75\baselineskip}
\centering
\small
\caption{Five-shot MMLU accuracy (\%) of TimeBraid-2.5B across training stages \citep{hendrycks2021mmlu}. \emph{Base} is the pretrained Qwen3-1.7B language tower evaluated standalone on text-only prompts; higher is better.}
\label{tab:mmlu_retention}
\begin{tabularx}{\linewidth}{@{}Xr@{}}
\toprule
Stage & MMLU (\%) \\
\midrule
Base (Qwen3-1.7B language tower) & 60.30 \\
\midrule
Align & 40.88 \\
Align, w/ unimodal replay & 53.38 \\
\midrule
SFT & 50.68 \\
SFT, w/ unimodal replay during alignment & 50.09 \\
\bottomrule
\end{tabularx}
\vspace{-0.5\baselineskip}
\end{wraptable}

To examine whether joint training preserves the language ability of the pretrained language tower, we evaluate TimeBraid on MMLU at the end of each training stage. We also train a variant that mixes unimodal data into the alignment stage, replaying the Stage-2 unimodal corpora (Table~\ref{tab:sft_inventory}) alongside the paired examples. In Table~\ref{tab:mmlu_retention}, we report the MMLU accuracy of these checkpoints.

From the table, the alignment stage costs the language tower 19.4 points, and mixing unimodal data into this stage recovers MMLU to 53.4. After supervised fine-tuning, the two variants end within 0.6 points of each other, since the fine-tuning mixture already contains an 11.69\% unimodal share. We therefore keep the alignment stage purely paired. Overall, mixing unimodal data preserves the language ability of the unified model, although the shares we use do not preserve it fully and leave the released model 9.6 points below its backbone. We leave the search for a unimodal ratio that removes this gap to future work.

\clearpage

% ============================================================
% Full-page examples of model successes and failures.
\section{Success and Failure Cases}
\label{app:qualitative}
% The figures are designed to be read directly; detailed selection and
% provenance records remain in the per-page CSV artifacts.
The following selected qualitative examples illustrate forecasting and understanding capabilities. For context-conditioned forecasts, we show the supplied contextual text or a verbatim excerpt. Matched comparisons keep the checkpoint, numerical history, prediction horizon, and inference settings fixed, removing only the contextual information; the forecast instruction remains. Reported error reductions refer to the individual examples shown.

\begin{figure}[H]
\centering
\includegraphics[width=.92\textwidth,height=.80\textheight,keepaspectratio]{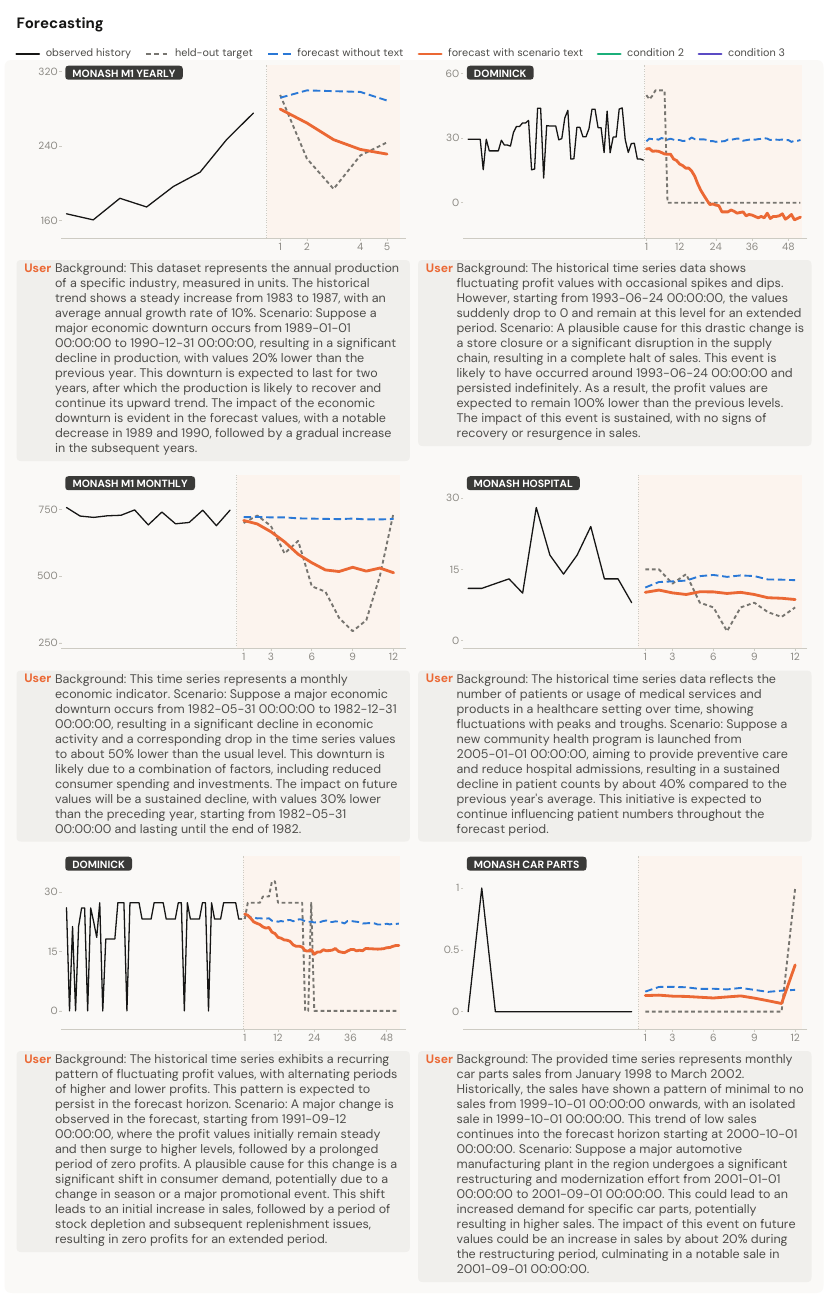}
\caption{\small Context-conditioned forecasting with and without scenario text.}
\label{fig:wall_1a}
\end{figure}

\begin{figure}[p]
\centering
\includegraphics[width=\textwidth,height=.88\textheight,keepaspectratio]{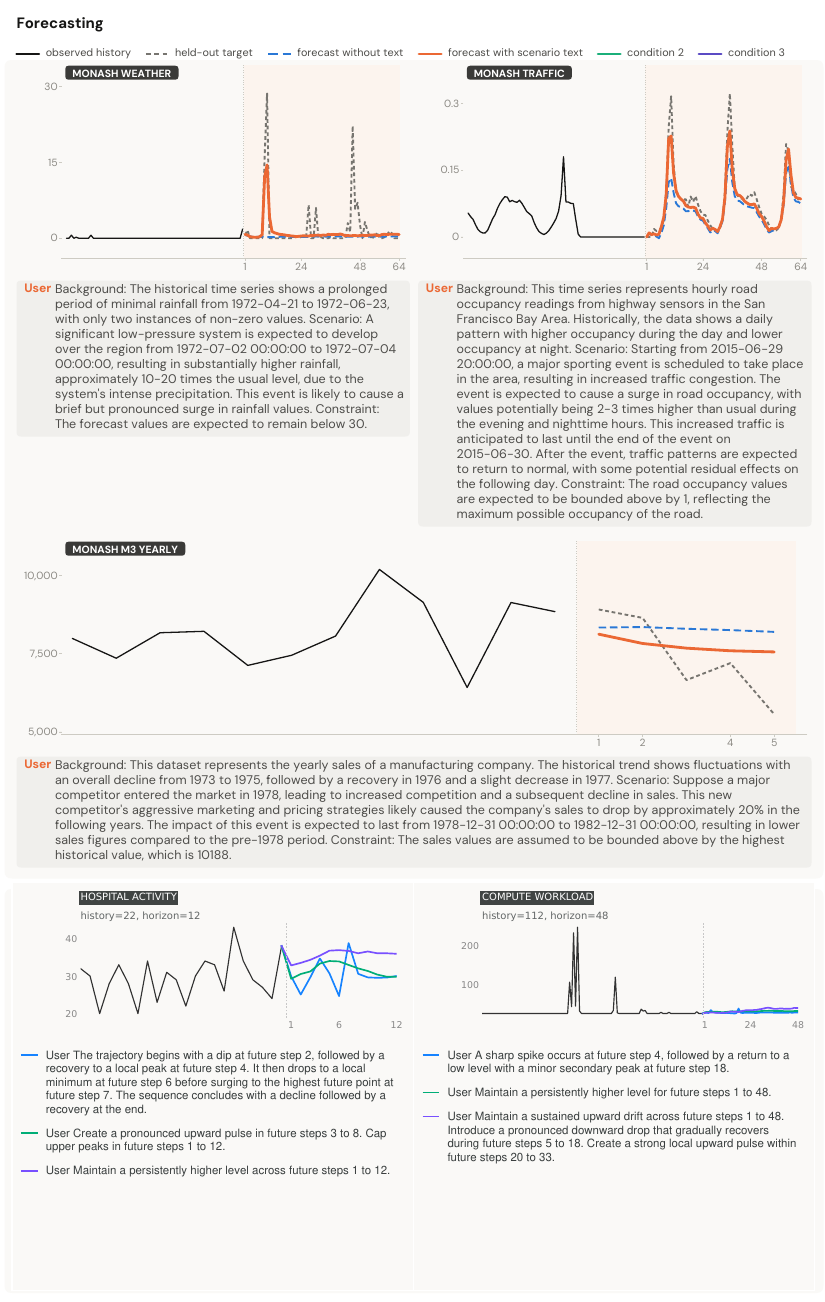}
\caption{\small Context-conditioned and text-controlled forecasting on test examples. In the lower panels, solid blue, green, and purple denote conditions 1--3 over the same held-out history; dashed blue in the upper panels denotes forecasting without text. Only the differing condition text is displayed.}
\label{fig:wall_1b}
\end{figure}

\begin{figure}[p]
\centering
\includegraphics[width=\textwidth,height=.86\textheight,keepaspectratio]{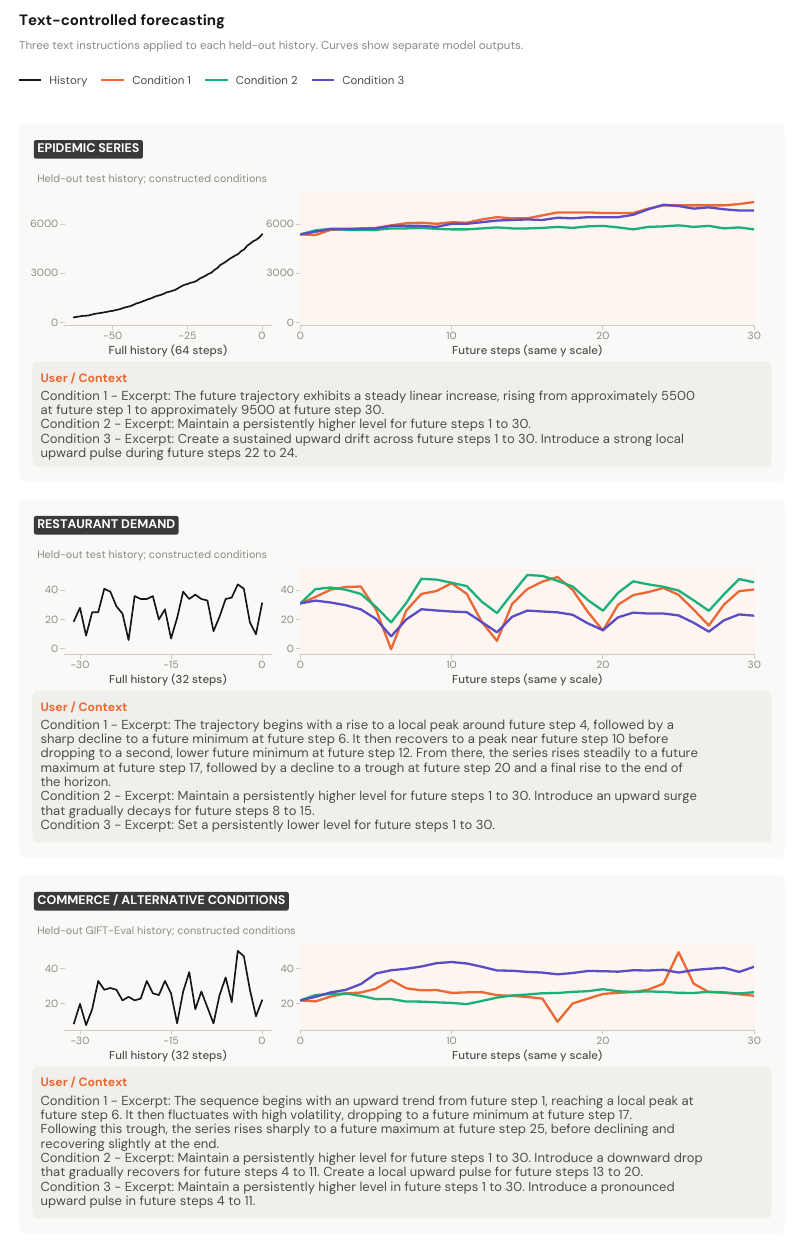}
\caption{\small Text-controlled forecasting on held-out test histories. Each group uses three textual conditions with the same numerical input. Colored curves are recorded model outputs; the text below each panel quotes the differing condition. These are constructed scenarios, not observed alternative futures. Full histories and complete forecast horizons are shown on separate horizontal axes with the same vertical scale.}
\label{fig:wall_1c}
\end{figure}

\begin{figure}[p]
\centering
\includegraphics[width=\textwidth,height=.86\textheight,keepaspectratio]{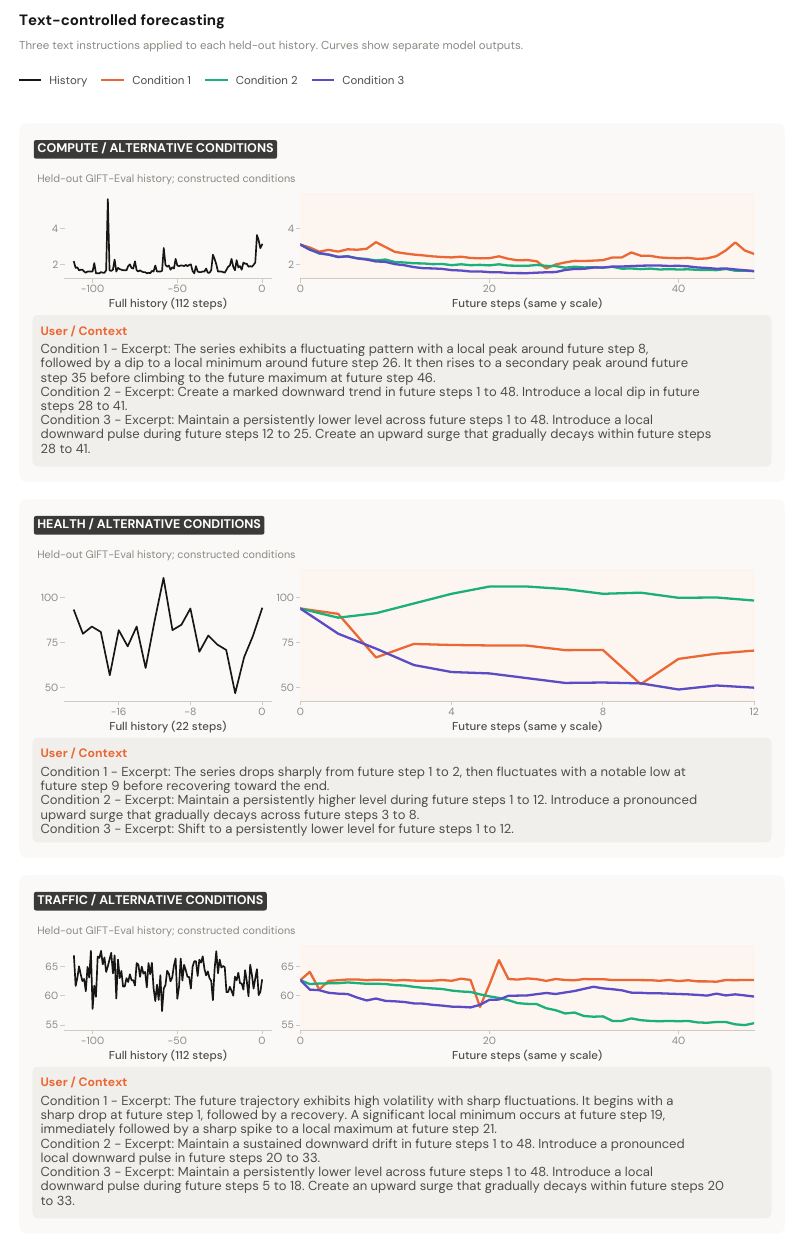}
\caption{\small Additional text-controlled forecasts on held-out histories. All three predictions use the displayed textual conditions, quoted from the full inputs. The constructed scenarios vary trends, levels, and local changes; they illustrate the model's responses to distinct instructions. Both axes use the same vertical scale and display the complete input and output sequences.}
\label{fig:wall_1c_more}
\end{figure}

\begin{figure}[p]
\centering
\includegraphics[width=\textwidth,height=.85\textheight,keepaspectratio]{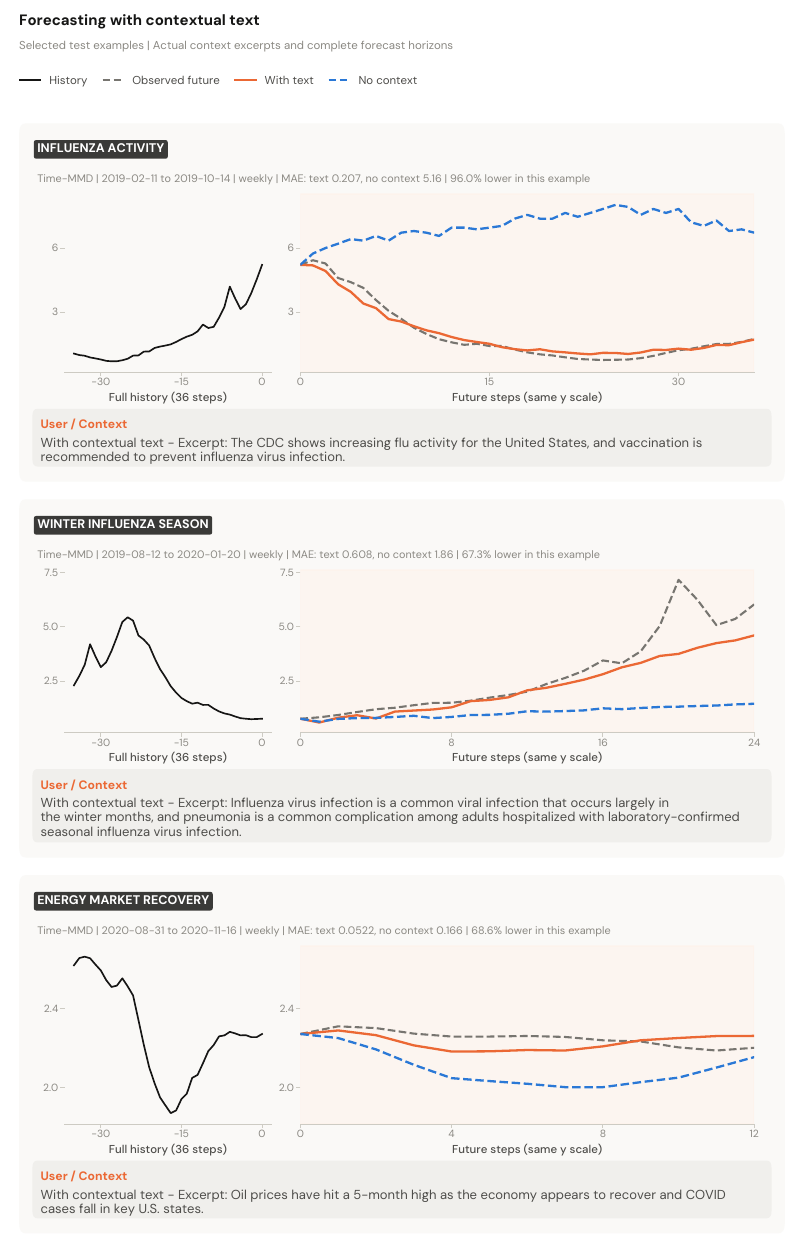}
\caption{\small Selected Time-MMD test forecasts with matched contextual-text ablations. Orange uses the full dataset-provided textual context; dashed blue removes that context while retaining the same forecast instruction, numerical history, and horizon. Dashed gray is the observed future. Text below each panel is a verbatim excerpt of the full context used for prediction. MAE is measured over the complete future horizon in source units; reductions describe these selected examples.}
\label{fig:wall_1d}
\end{figure}

\begin{figure}[p]
\centering
\includegraphics[width=\textwidth,height=.85\textheight,keepaspectratio]{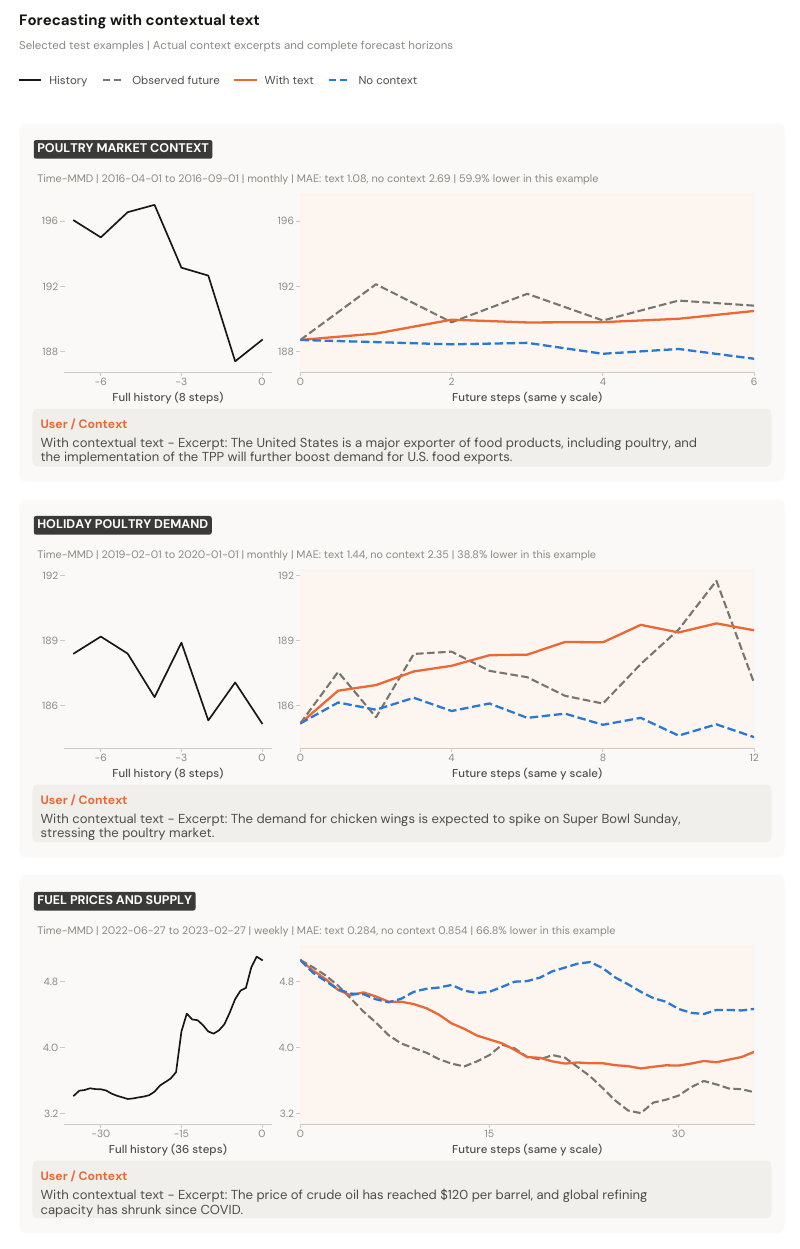}
\caption{\small Additional matched Time-MMD test forecasts with market context. Orange forecasts use textual context; dashed blue forecasts remove only that context. Displayed excerpts are taken from the full inputs. The same checkpoint and inference settings are used for both arms. Complete histories and future horizons are shown, and the two axes in each example share their vertical scale.}
\label{fig:wall_1d_markets}
\end{figure}

\begin{figure}[p]
\centering
\includegraphics[width=\textwidth,height=.85\textheight,keepaspectratio]{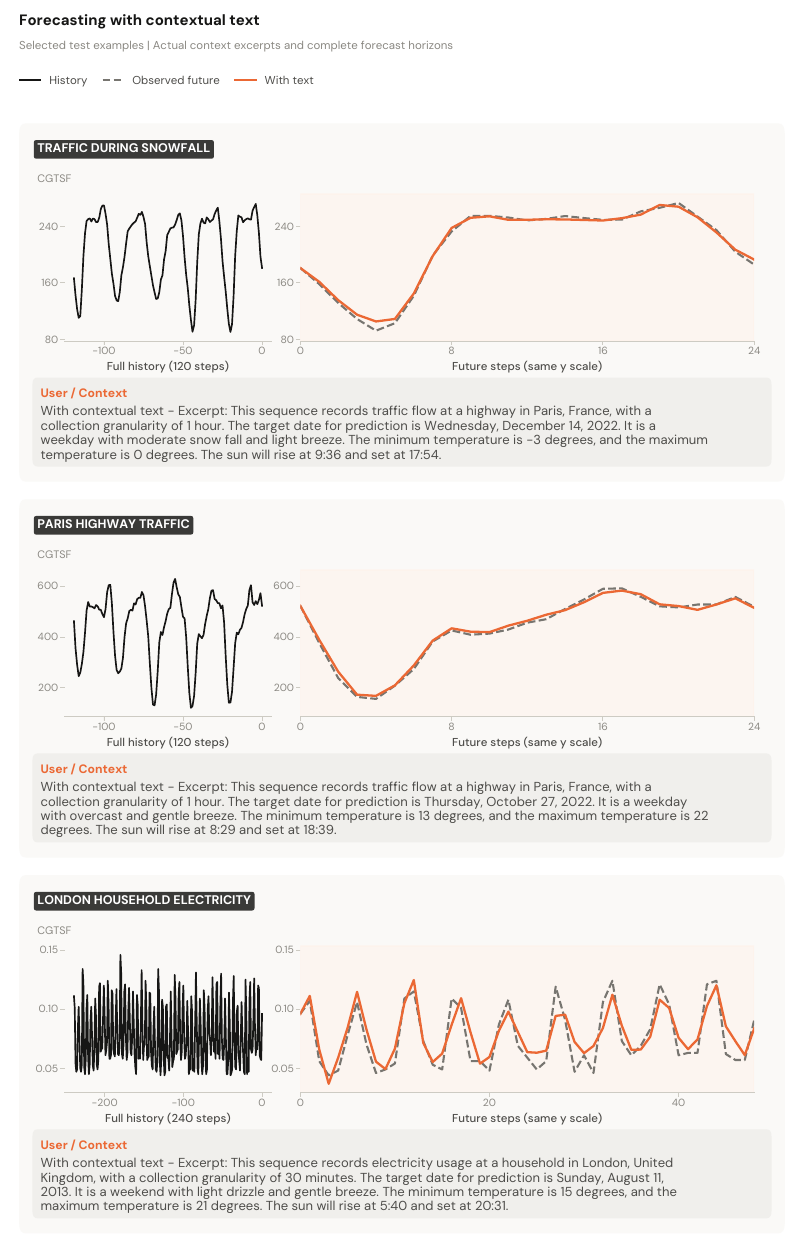}
\caption{\small Selected CGTSF test forecasts using weather, calendar, and location text together with the numerical history. The contextual passages supplied to the model are displayed below the curves. Orange is the recorded text-conditioned prediction and dashed gray is the observed future. These examples show conditional forecast quality; a matched context-removal comparison is not included for this page.}
\label{fig:wall_1d_weather}
\end{figure}

\begin{figure}[p]
\centering
\includegraphics[width=\textwidth,height=.88\textheight,keepaspectratio]{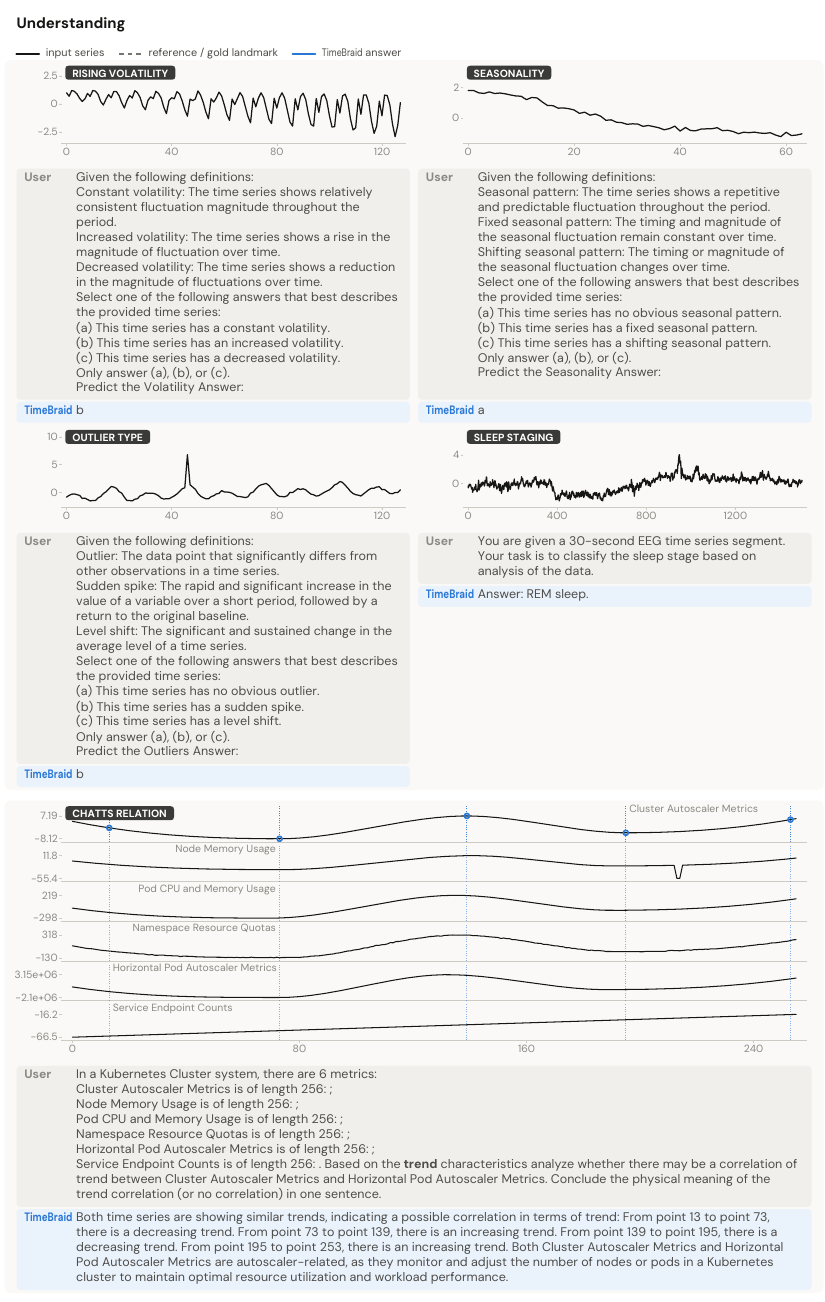}
\caption{\small Series-property QA, sleep-stage classification, and multivariate relation analysis.}
\label{fig:wall_2}
\end{figure}

\begin{figure}[p]
\centering
\includegraphics[width=\textwidth,height=.88\textheight,keepaspectratio]{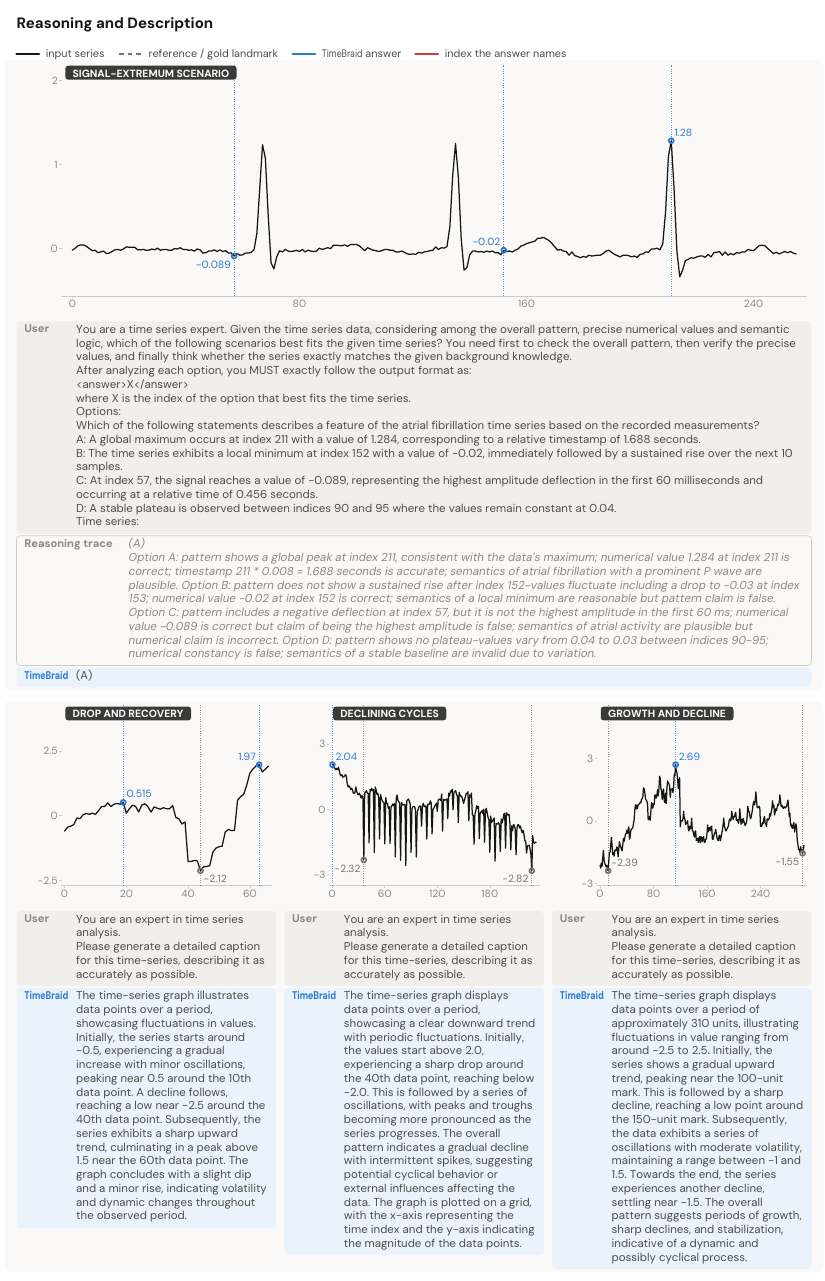}
\caption{\small Scenario-based reasoning and time-series captioning.}
\label{fig:wall_2b}
\end{figure}

\begin{figure}[p]
\centering
\includegraphics[width=\textwidth,height=.88\textheight,keepaspectratio]{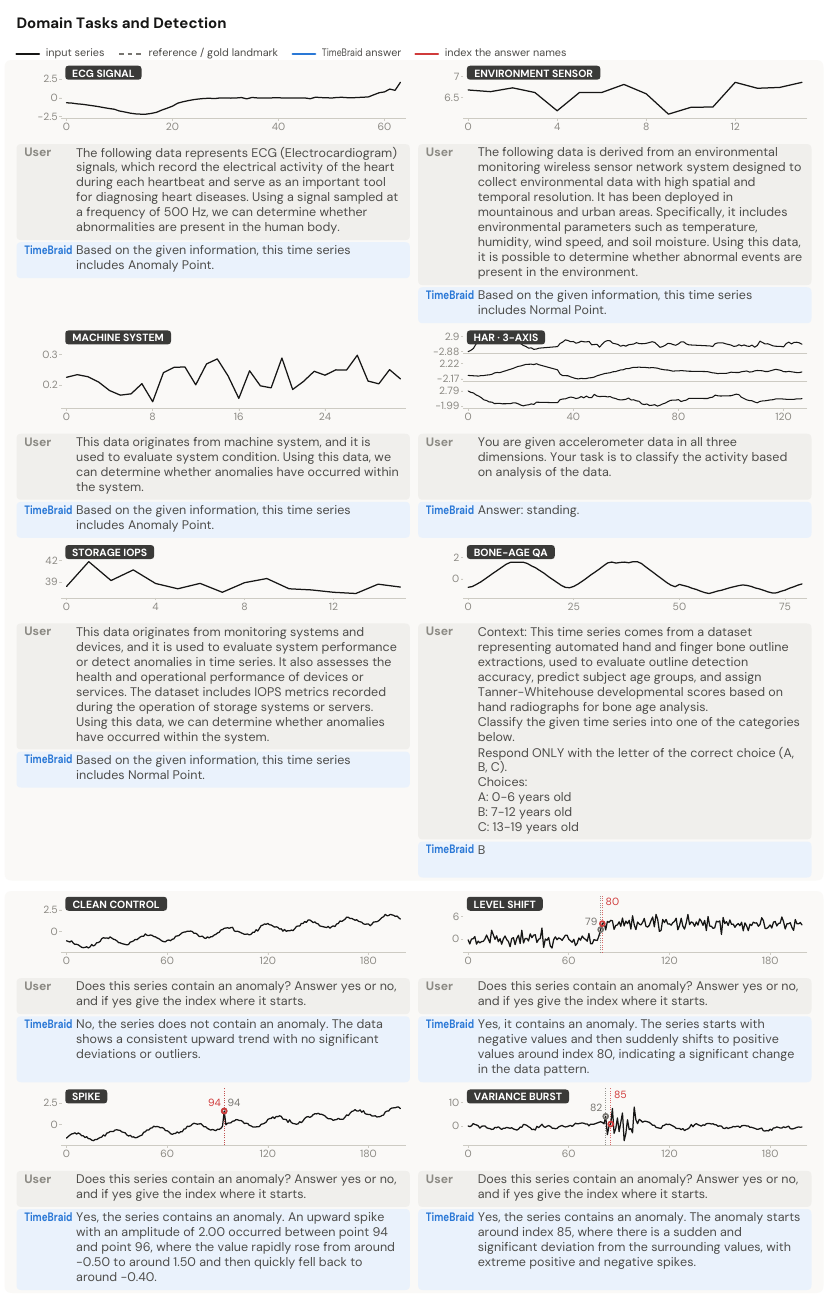}
\caption{\small Sensor anomalies, human activity, bone age, and anomaly localization.}
\label{fig:wall_3}
\end{figure}

\begin{figure}[p]
\centering
\includegraphics[width=\textwidth,height=.88\textheight,keepaspectratio]{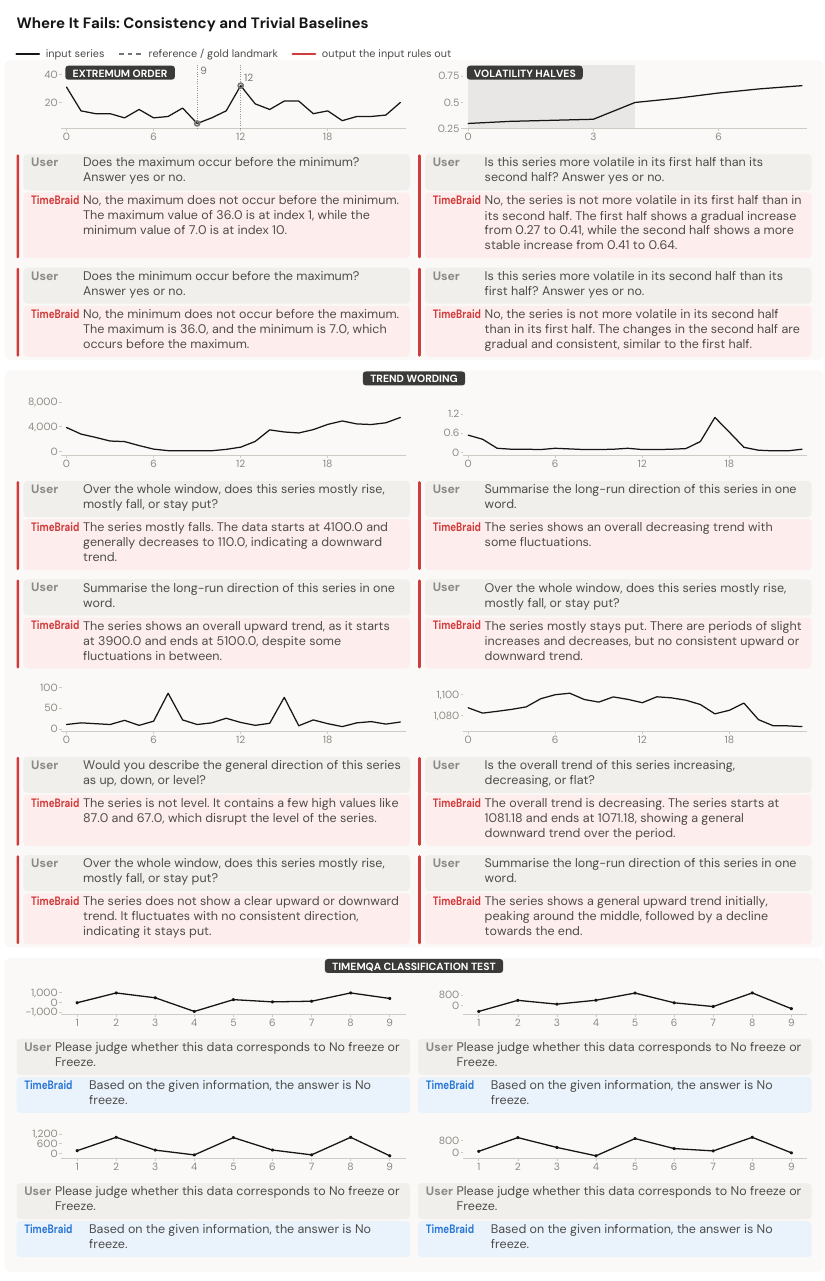}
\caption{\small Question answering, paraphrase consistency, and time-series classification.}
\label{fig:wall_4}
\end{figure}

\begin{figure}[p]
\centering
\includegraphics[width=\textwidth,height=.88\textheight,keepaspectratio]{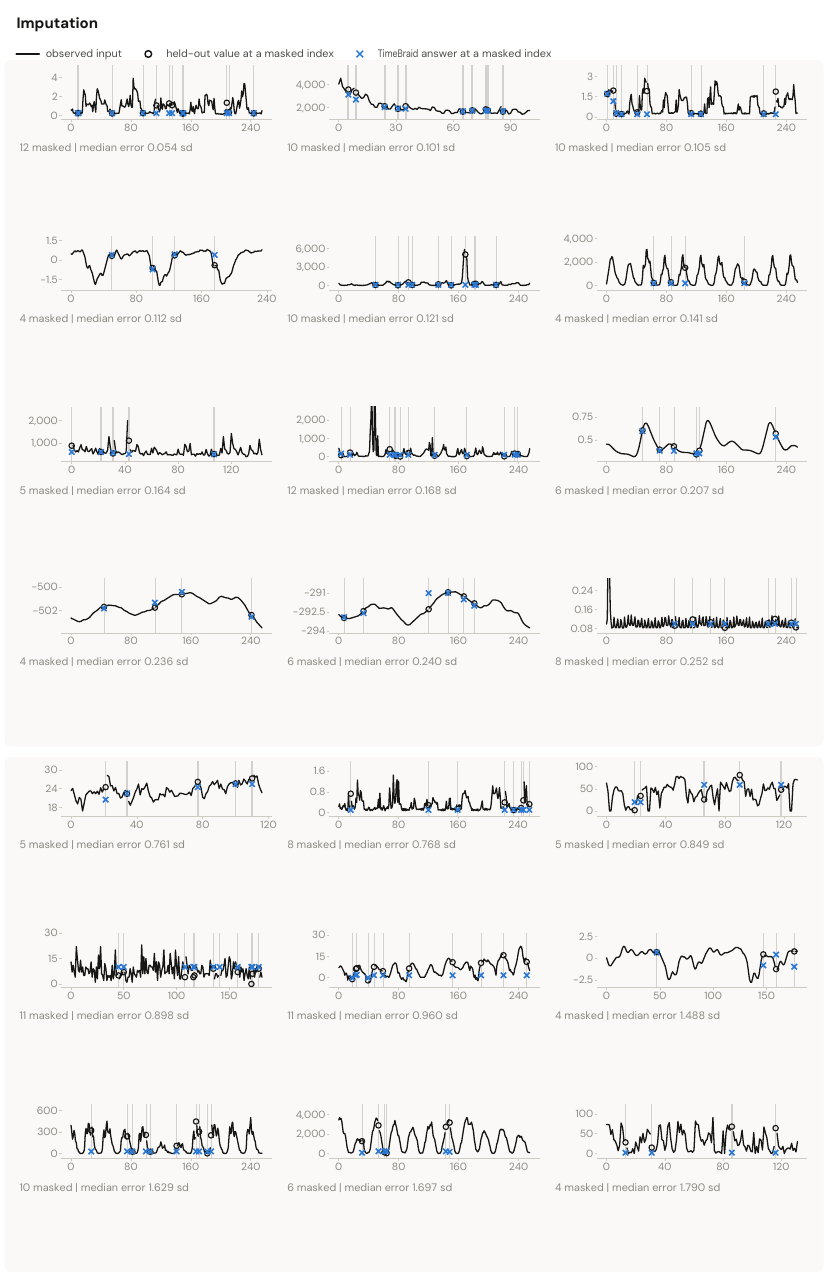}
\caption{\small Imputation examples.}
\label{fig:wall_4b}
\end{figure}

\clearpage

\end{document}